\documentclass[10pt]{article}

\usepackage[scaled]{helvet}

\usepackage{sectsty}

\sectionfont{\sffamily}
\subsectionfont{\sffamily}
\subsubsectionfont{\sffamily}

\usepackage{fullpage}
\usepackage{hyperref}
\hypersetup{colorlinks,allcolors=black}

\usepackage{graphicx}
\graphicspath{{./pics/python_pictures/}, {./pics/diagrams/}}
\usepackage{subfigure}
\usepackage{graphicx}

\usepackage{amsmath}
\usepackage{amsthm} 
\usepackage[toc]{glossaries} 
\usepackage{blkarray} 
\usepackage{csquotes}
\usepackage{multirow}
\usepackage{biblatex}
\newcommand{\Sec}[1]{Section~{#1}}
\newcommand{\Fig}[1]{Figure~{#1}}

\newcommand{\dontshow}[1]{}
\newcommand{\acrfulls}[1]{\acrlong{#1}s (\acrshort{#1}s)}
\newcommand{\acrshorts}[1]{\acrshort{#1}s}
\theoremstyle{definition}
\newtheorem{exmp}{Example}[section]

\newtheorem{rmk}{Remark}[section]
\newcommand{\remark}[1]{ \begin{rmk} #1 \end{rmk}}
\newtheorem{obj}{Objective}[section]
\newcommand{\objective}[1]{\begin{obj} The main objective of this section is #1 \end{obj}}
\newtheorem{mydef}{Definition}[section]

\newcommand{\quotes}[1]{``#1''}
\newcommand{\isofusa}{{ISO~26262:2018}}
\newcommand{\isosotif}{{ISO~21448:2022}}
\newcommand{\isoai}{{ISO/PAS~8800}}

\usepackage{todonotes}
\makeglossaries

\begin{document}
\title{\sffamily Safety Integrity Framework for Automated Driving}
%
%

\author{\sffamily
    Moritz Werling
    \thanks{BMW Group} 
    \and
    \sffamily
    Rainer Faller 
    \thanks{Exida} 
    \and
    \sffamily
    Wolfgang Betz
    \thanks{Eracons GmbH} 
    \and
    \sffamily
    Daniel Straub
    \thanks{Technical University of Munich}
}

\date{\sffamily March 2025}

%
%
%

\maketitle              

\renewcommand{\abstractname}{\sffamily Abstract}
\begin{abstract}
This paper describes the comprehensive safety framework that underpinned the development, release process, and regulatory approval of BMW's first SAE Level 3 Automated Driving System. The framework combines established qualitative and quantitative methods from the fields of Systems Engineering, Engineering Risk Analysis, Bayesian Data Analysis, Design of Experiments, and Statistical Learning in a novel manner. The approach systematically minimizes the risks associated with hardware and software faults, performance limitations, and insufficient specifications to an acceptable level that achieves a Positive Risk Balance.
At the core of the framework is the systematic identification and quantification of uncertainties associated with hazard scenarios and the redundantly designed system based on designed experiments, field data, and expert knowledge. The residual risk of the system is then estimated through Stochastic Simulation and evaluated by Sensitivity Analysis.
By integrating these advanced analytical techniques into the V-Model, the framework fulfills, unifies, and complements existing automotive safety standards. It therefore provides a comprehensive, rigorous, and transparent safety assurance process for the development and deployment of Automated Driving Systems.
\end{abstract}
\newpage
\tableofcontents
\newpage
\newacronym{ads}{ADS}{Automated Driving System}
\newacronym{alarp}{ALARP}{As low as reasonably practicable}
\newacronym{alks}{ALKS}{Automated Lane Keeping Systems}
\newacronym{asil}{ASIL}{Automotive Safety Integrity Level}
\newacronym{bn}{BN}{Bayesian Network}
\newacronym{ci}{CI}{Confidence Interval}
\newacronym{cdf}{CDF}{Cumulative Distribution Function}
\newacronym{dag}{DAG}{Directed Acyclic Graph}
\newacronym{doe}{DOE}{Design of Experiments}
\newacronym{dpa}{DPA}{Dependent Performance Analysis}
\newacronym{et}{ET}{Event Tree}
\newacronym{ffoa}{FFoA}{Functional Field of Application}
\newacronym{fn}{FN}{False Negative}
\newacronym{ft}{FT}{Fault Tree}
\newacronym{fta}{FTA}{Fault Tree Analysis}
\newacronym{eta}{ETA}{Event Tree Analysis}
\newacronym{fp}{FP}{False Positive}
\newacronym{hara}{HARA}{Hazard Analysis and Risk Assessment}
\newacronym{hazop}{HAZOP}{Hazard and Operability}
\newacronym{hira}{HIRA}{Hazard Identification and Risk Assessment}
\newacronym{hs}{HS}{Hazard Scenario}
\newacronym{hw}{HW}{Hardware}
\newacronym{ifv}{IFV}{Influence Factor Variable}
\newacronym{kpi}{KPI}{Key Performance Indicator}
\newacronym{l2}{L2}{SAE Level 2}
\newacronym{l3}{L3}{SAE Level 3}
\newacronym{mbse}{MBSE}{Model-based Systems Engineering}
\newacronym{mil}{MiL}{Model in the Loop}
\newacronym{ml}{ML}{Machine Learning}
\newacronym{mle}{MLE}{Maximum Likelihood Estimation}
\newacronym{mrm}{MRM}{Minimal Risk Maneuver}
\newacronym{mcs}{MCS}{Monte Carlo Simulation}
\newacronym{mtbf}{MTBF}{Mean Time Between Failures}
\newacronym{odd}{ODD}{Operational Design Domain}
\newacronym{ofat}{OFAT}{One-factor-at-a-time}
\newacronym{pfd}{PFD}{Probability of Failure on Demand}
\newacronym{pfh}{PFH}{Probability of Failure per Hour}
\newacronym{prb}{PRB}{Positive Risk Balance}
\newacronym{pdf}{PDF}{Probability Density Function}
\newacronym{rac}{RAC}{Risk Acceptance Criterion}
\newacronym{rra}{RRA}{Residual Risk Analysis}
\newacronym{sa}{SA}{Sensitivity Analysis}
\newacronym{sifad}{SIFAD}{Safety Integrity Framework for Automated Driving}
\newacronym{sil}{SiL}{Software in the Loop}
\newacronym{safetyil}{SIL}{Safety Integrity Level}
\newacronym{spv}{SPV}{Safety Performance Variable}
\newacronym{sw}{SW}{Software}
\newacronym{sysml}{SysML}{Systems Modeling Language}
\newacronym{tja}{TJ-ADS}{Traffic Jam Automated Driving System}
\newacronym{tlsr}{TLSR}{Top-Level Safety Requirement}
\newacronym{too3}{2oo3}{2-out-of-3}

\section*{Acknowledgments}
The authors would like to thank Alexander Prehn, Arne Haas, Daniel Matlok, Felix Fahrenkrog, Felix Modes, Georg Tanzmeister, Ivanova Maltiza, Lailong Song, Ludwig Drees, Mehdi Farid, Moritz Schneider, Robert Ziener, and Stanislav Braun for their thorough reviews of the manuscript and the fruitful discussions.

\section*{Disclaimer}
This paper describes\footnote{The text has been optimized using ChatGPT.} the safety framework that has been applied in the development and release process of BMW's first SAE Level 3 system.\footnote{cf.\ \cite{SAEJ3016_2018} for the different automation levels.} The BMW Personal Pilot L3 is an \quotes{eyes-off} system operating in traffic jams on motorways with up to 60\,kph.
The safety framework builds largely on our previous publications \cite{werling2023statistical, werling2021quantitative} and incorporates the lessons learned from the Personal Pilot L3 project. The presented examples are chosen for their ability to clearly explain the framework, even if some sacrifices in realism, particularly regarding numerical values, must be made. Readers should exercise caution and understand that the described methods may undergo further refinement or modification based on future research. After all, this publication aims to foster discussion and contribute to the development of a standardized approach.

\section{Introduction}\label{sec:intro}
\label{sec:problem_statement}
One of the most challenging questions regarding an \acrlong{ads} (\acrshort{ads}) concerns its safety:
\begin{displayquote}
    \textit{How can one build a safe \acrshort{ads} and provide sufficient evidence for it?}
\end{displayquote}
The term ``safe'' is often defined as the ``absence of unreasonable risk'', cf.\ \cite{rausand2021system}. From a probabilistic safety analysis point of view ``risk'' is often defined as ``expected harm''.
To effectively estimate and reduce risk, it is crucial to identify and \textit{quantify uncertainties} in the system and the environment (incl.\ the driver). For further discussion it helps to differentiate two types of uncertainty \cite{hullermeier2021aleatoric}:
\subsubsection*{Aleatoric Uncertainty}
Aleatoric uncertainty refers to the inherent variability or randomness in a system or process. It is often associated with natural phenomena and cannot be reduced through the acquisition of more knowledge or data. 
Regarding an \acrshort{ads}, aleatoric uncertainties arise from the driving scenarios including the behavior of other traffic participants, weather, and road conditions. 
Furthermore, some uncertainties associated with parts of the \acrshort{ads}, such as the perception system,  can also be interpreted as being of an aleatoric nature.

\begin{exmp}[Pedestrian in lane and brake system failure]
    The uncertainty associated with the unpredictable appearance of a pedestrian in the lane is aleatoric in nature. Similarly, the uncertainty about the failure of the braking system is also aleatoric. Simply collecting more data will not make the occurrence of a pedestrian or a brake failure less likely. These events can be seen as inherently unpredictable, random events that introduce aleatoric uncertainty.
\end{exmp}

\subsubsection*{Epistemic Uncertainty}
Epistemic uncertainty arises from a lack of knowledge over a system or process. Unlike aleatoric uncertainty, epistemic uncertainty can be reduced through the acquisition of more knowledge. In the context of automated driving, this involves in general gathering more data (\quotes{coverage}) about the system's performance and the driving environment. \\

\begin{exmp}[Wheel diameters]
Precise wheel diameter estimates are required for a reliable velocity measurement. Immediately after starting the vehicle, the epistemic uncertainty about the diameters is large, but it can be reduced more and more through measurements.
\end{exmp}

Distinguishing between epistemic and aleatoric uncertainty can be challenging in some situations, but is also not crucial. Both uncertainties can be quantified and propagated to a risk estimate based on probability theory. 
Nevertheless, the distinction helps in  understanding and addressing uncertainties of \textit{complex Automated Driving Systems} in the context of \textit{limited data}.

\subsection{Quantification and Propagation of Uncertainties}
The quantification of the uncertainties directly based on data at \textit{vehicle level} leads to the ``Approval Trap'', cf.\ \cite{wachenfeld2016release, kalra2016driving}. It refers to the fact that the straightforward approach, where the system is build first (``black-box'') and then the (uncertain) failure rate of the vehicle is quantified by field testing only (``driving to safety''), would require a substantial fleet and many years of testing to demonstrate a reasonably safe system. As industries such as the nuclear power and aviation sector show, there are alternative approaches that fall under the disciplines of \textit{Engineering Risk Analysis} \cite{rausand2021system, fenton2018risk}. This needs to be combined with \textit{Systems Engineering} \cite{incose2023handbook, friedenthal2014practical} in order to coordinate numerous software and hardware engineering teams. If these teams devise a smart system design with \textit{redundancies}, the amount of data needed to quantify the uncertainty on the \textit{component level} is significantly decreased, see Ex.~\ref{ex:2oo3_aleatory_and_epistemic_uncertainty}. However, in order to identify the uncertainties in the components as well as accurately propagate them to the vehicle-level for a risk estimation, particular aspects in both the system and the environment have to be well understood (\quotes{gray-box}).

\begin{exmp}[Aleatoric and Epistemic Uncertainty]
\label{ex:2oo3_aleatory_and_epistemic_uncertainty}
Let us consider a typical combination of aleatoric and epistemic uncertainties, and
how they propagate through a system. In this case, we have a redundant system of three sensors that trigger a braking maneuver if at least two sensors detect a pedestrian, a so-called \acrfull{too3} voting. This logic is often a suitable compromise because a single false detection can neither lead to a \acrfull{fn}\footnote{E.g., the vehicle does not react to an actual object.} nor a \acrfull{fp}\footnote{E.g., the vehicle reacts to an imaginary object.} reaction. In this example we focus on the \acrshort{fn} reaction.\\
The sensors $S_i$ with $i \in \{1, 2, 3\}$ fail to detect a pedestrian (early enough to come to full stop) with probabilities $p_i$. Note that we consider the failure events as aleatoric, whereas the $p_i$ as epistemic.
Other uncertainties in the system are neglected. This is illustrated in Fig.~\ref{fig:twooothree}. \\
We collect a sample of 1000 representative pedestrian encounters and evaluate that the sensors 1, 2, and 3 have misdetected the pedestrian 0, 1, and 2 times, respectively. Based on this evidence we can quantify the epistemic uncertainty in the parameters as shown in Fig.~\ref{fig:posterior}. As $S_1$ has never failed, the probability density is the highest at zero and quickly decays towards higher failure probabilities. The other sensors have shown to fail, so that the graphs start at the origin, increase their densities and then also decay towards zero. For mathematical details, we refer to Sec.~\ref{sec:nononformativeprior}. \\
As the sensors are based on different modalities (camera, lidar, radar) we assume statistical independence of the failures in this first analysis. With this assumption, the \acrshort{too3} voting mechanism can be represented by the Bayesian Network shown in Fig.~\ref{fig:twooothree}. The quantified uncertainties of the single channels are propagated through the network to the failure of the voter. Mathematically, this is a multidimensional integration which can be performed in closed form\footnote{The epistemic uncertainty of each sensor can be integrated (marginalized) into its failure probability and we get $p_1 = 1/1002, p_2 = 2/1002, p_3 = 3/1002$, respectively, see Sec.~\ref{sec:failure_prob_discrete}. The probability of failure of the \acrshort{too3} voter is 
$
    p =  p_1 p_2 + p_1 p_3 + p_2 p_3 - 2 p_1 p_2 p_3 
$, which can be derived by basic principles of probability \cite{rausand2021system}.} in this simple case and leads to a predicted failure probability of  $p=1.1 \times 10^{-5}$ at the voter output. Note that we would have to collect about $10^{5}$ successful vehicle stops without any fails as evidence to get to the same estimate, cf.\ Sec.~\ref{sec:failure_prob_discrete}, if the combined system is evaluated as a black box. Based on the system design and the assumptions we made, we can reduce the required amount of data by a factor of 100.
%
%
\begin{figure}[htbp]
    \centering
    \subfigure[Parameter uncertainties]{\label{fig:posterior}
        \includegraphics[width=0.45\textwidth]{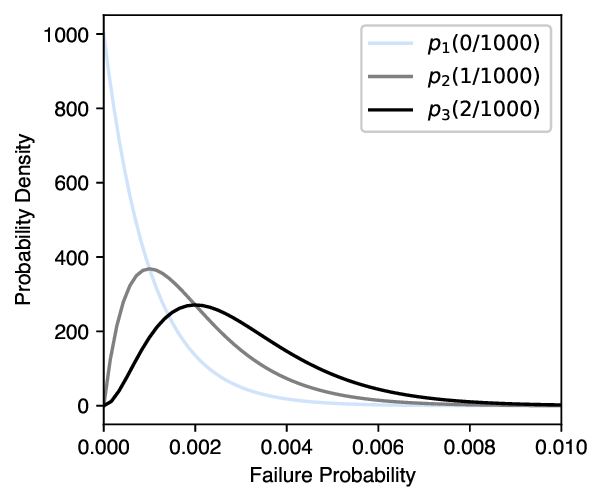}
    }
    \subfigure[Bayesian Network of a \acrshort{too3} voting system with independent sensors]{\label{fig:twooothree}
        \includegraphics[width=0.45\textwidth]{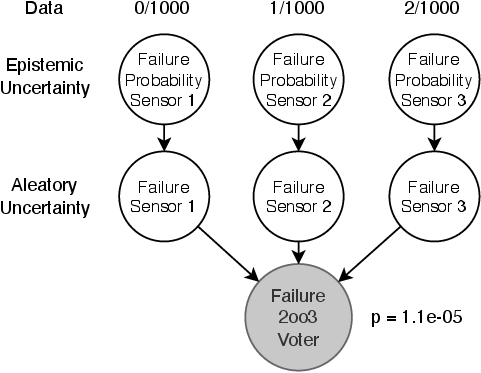}
    }
    \caption{Quantification and propagation of epistemic and aleatoric uncertainty for a \acrshort{too3} voter and independent sensor performances}
    \label{fig:2oo3_example}
\end{figure}
\end{exmp}

The preceding example shows in a simplified way how system uncertainties propagate to a vehicle's response. To determine the overall risk for the scenario, there are additional uncertainties to consider regarding the occurrence of the scenario, parameters within the scenario, and the severity in case of a collision -- as well as their interdependencies.

\begin{obj}
    The goal of the presented \acrfull{sifad} is not only to tailor statistical methods to the safety validation of the \acrshort{ads} but also support the development of a safe, verifiable, and validatable \acrshort{ads}. Furthermore, it serves as the foundation for building its safety case.
\end{obj}

\subsection{Related Standards}\label{sec:relatedwork}
Existing standards alone do not provide the means to validate the safety of \acrshorts{ads} with \acrshort{l3} and above.
Nonetheless, there is value in leveraging established standards that already offer guidance in certain areas. Additionally, we observe some overlap between standards, resulting in redundant work and, in certain cases, the potential for responsibility gaps in transitional areas. Therefore, a key objective of the presented framework is to establish a unified terminology and generate work products that align with and complement the objectives of the relevant existing standards.

The subsequent section offers an overview of the key standards that are most relevant to the safety integrity of \acrshorts{ads}. It outlines their applicability and briefly discusses how they are addressed within the proposed framework.

\subsubsection{\isofusa ~-- Functional Safety}\label{sec:iso26262}
The ISO\,26262 series \cite{ISO26262} provides guidelines and requirements for the systematic identification of hazards, the derivation and implementation of safety requirements as well as a verification and validation process to minimize the risk associated with incorrect requirements and faults of the hardware and software. It strongly builds on the concept of \acrfull{asil}, a risk classification system defined by the standard itself. The standard is comprehensive, covering the whole safety life cycle, which also includes organizational and tooling aspects. \\

At the time the standard was developed, safety aspects at the highest abstraction level (``item definition'', ``intended functionality'') were not considered. The standards therefore assumes a safe functional specification of the \acrshort{ads} as the starting point and only addresses deviations from it (``malfunctioning behavior'') without assessing the safety of the intended behavior itself.

\begin{exmp}[Potholes not addressed in item definition]
The standard only addresses issues related to potholes and other bad road surface conditions if the item definition includes a requirement for the \acrshort{ads}, e.g., to adapt its speed accordingly. Without this requirement, no malfunctioning behavior related to bad road surface conditions will be assessed and therefore neither a safety requirement will be derived nor a safety mechanism will be implemented to address potholes.  
\end{exmp}

However, the most significant constraint in implementing the standard for \acrshort{ads} lies in the paradigm of resolving any identified error. This approach is especially not suitable for the perception system of an \acrshort{ads} due to its inherent variability, which results in infrequent but non-negligible failures. In other words, the standard struggles to deal with \acrshort{ads} components that have some non-negligible uncertainty associated with their functionality (with the exception of random hardware failures).

\begin{exmp}[Uncertainty in pedestrian detection]
    Let us assume, that in order to avoid collisions with vulnerable road users, the \acrshort{ads} shall detect a pedestrian within 50\,m with a maximum longitudinal error of 1\,m. The current state of the art of perception systems cannot implement this requirement with \quotes{probability 1}. If the system is intensively tested, there will always be cases where the requirement is violated. We can improve the system but we can never be certain that the requirement will be met under any new combination of environmental factors.
\end{exmp}

\subsubsection{\isosotif ~-- Safety of the Intended Functionality}
ISO\,21448 addresses hazardous behavior of \acrshort{ads} caused by its perception and decision making modules (\quotes{functional insufficiencies}) \cite{ISO21448}. It also emphasizes on a safe intended functionality at the vehicle level. Furthermore, the standard focuses significantly more on scenarios (\quotes{situations} in ISO\,26262) as the environment has a large impact on the perception system. And lastly, the standard centers on an iterative function development process based on feedback loops, even after the system release in the operation phase based on field monitoring. \\
While doing so, the standard opposes treating \quotes{functional insufficiencies} as equivalent to \quotes{faults} of ISO\,26262. It justifies\footnote{Introduction of \isosotif} the use of a separate framework with new terminology based on the need for different measures than those suggested by ISO\,26262. This comes at the cost of additional effort to keep the activities and work products of the two frameworks in sync.\footnote{\isosotif, A.2} Another drawback of the two separate frameworks is the unaddressed limitation when uncertainties are encountered within the ISO\,26262 framework, as mentioned above. For instance, the authors are unable to determine how the validation of the ``Functional'' and ``Technical Safety Concept'' of ISO\,26262 can be accomplished in the presence of sensor noise with the measures defined in ISO\,26262. \\
Lastly and foremost, the ISO\,21448 standard suggests for \acrshort{l3} and above the break down of a quantitative risk acceptance criterion on the vehicle-level to a verifiable performance on a component level, but stays vague on how to accomplish that. Essential statistical methods like Stochastic Simulations and Sensitivity Analysis are only mentioned.

\subsubsection{SAE Level 0-2 vs. 3-5}
For systems of Level 0-2, the above issues of ISO\,21448 and ISO\,26262 do not arise thanks to the presence of the driver serving as a fallback. More concretely, limiting automatic steering and braking/acceleration actions serves as an effective safety mechanism to address failure modes that could potentially overwhelm the driver.
These simple safety mechanisms can be derived, implemented, verified, and validated within the ISO\,26262 framework. The remaining failure modes can most likely be handled by the driver, leading to relaxed requirements falling outside of ISO\,26262 (quality measures, QM, or even no measures). At the same time, the high level of controllability for the remaining failure modes leads to probabilistic safety requirements that can be validated on a vehicle level by means provided by ISO\,21448.

\begin{exmp}[Keep lane system with steering limiter]
    For a lane-keeping system, the hazard of leaving the lane with extreme steering torque is uncontrollable, even with hands firmly on the steering wheel, requiring an \acrshort{asil} D safety mechanism. This can be implemented by limiting the steering dynamics/force to a certain level, which can be realized without any complex perception components. Leaving the lane with a moderate steering movement within the limits is not addressed by this simple mechanism but is likely to be managed by the driver instead (e.g., C1 controllability class). This usually leads to a QM assignment
    and will therefore not be addressed by ISO\,26262. Based on ISO\,21448 and a suitable risk acceptance criterion, one can derive an acceptable vehicle-level performance rate, e.g., 1 keep-lane failure every 1000\,h, which can be demonstrated based on field tests.    
\end{exmp}

This separation does not work for \acrshort{l3} systems and above because of the limited driver fallback. It leads to a high safety load on all components involved in detecting a dangerous traffic scenario and reacting to it. Especially for the perception but also for the controls components this translates into high requirements in terms of \acrfull{hw}/\acrfull{sw} faults \emph{and} performance. While it is true that the measures addressing faults and performance issues differ, large parts of the development activities are identical or can be at least unified. This leads to a leaner, less complex and therefore safer framework.

\remark{We have found the common usage of the terms ``Functional Safety'' (FuSa) and ``Safety of the Intended Functionality'' (SOTIF) within the automotive industry to be misleading in the context of \acrshort{l3} and above. These terms are often used as synonyms for the standards ISO\,26262 and ISO\,21448, respectively. However, outside the automotive industry, the definition of a safe Intended Functionality is an important aspect of Functional Safety. Additionally, a sensor insufficiency is considered part of Functional Safety -- but is distinct from the Intended Functionality.\footnote{The intended behavior of the system is certainly not to misdetect a pedestrian.}\\
To promote clear understanding, the presented framework therefore avoids the terms ``Functional Safety'' and ``Safety of the Intended Functionality''. Furthermore, it uses, wherever possible, a neutral terminology that is accessible to all experts in the field of ISO\,26262 and ISO\,21448.}


\subsubsection{\isoai ~-- Safety and Artificial Intelligence}
The increasing integration of Artificial Intelligence (AI) in safety-critical components of vehicles has led to the development of ISO/PAS\,8800. This industry-specific guidance aims to tailor and extend the approaches defined in the existing standards, ISO\,26262 and ISO\,21448, to the needs of \acrfull{ml}. In general, it can be observed that the safety focus shifts from the implementation to the data and tooling.
The \acrshort{sifad} also builds on this paradigm. \\
However, the previously mentioned limitations of the existing standards are not addressed in ISO/PAS\,8800 either. Specifically, the ISO/PAS\,8800 framework assumes that the vehicle-level risk acceptance criterion is already broken down into performance requirements amenable for verification (target metrics, Key Performance Indicators (\acrshort{kpi}), error rates and probabilities) and assigned to the AI component as per the guidelines set forth in ISO\,26262 and ISO\,21448. As explained earlier, this is not given for \acrshort{l3} systems. \\
While building on ISO/PAS\,8800, our framework only differentiates between \acrshort{ml} and non-\acrshort{ml} components where necessary. That is, we found the statistical methods proposed by our framework to be largely applicable independent of the underlying technology. However, additional care needs to be taken when dealing with \acrshort{ml} technologies, especially regarding issues like over-fitting, see Remark~\ref{rmk:overfitting} below.
\subsection{Foundational Approach and Overview of the SIFAD}  
The harmonization and supplementation of the aforementioned standards is subject of the presented work.
The supplementation builds on a variety of publications, which are referenced at the appropriate places throughout the paper. \\

In a nutshell, the framework comprises
\begin{itemize}
    \item the systematic \textbf{identification of undesired vehicle behavior in \acrfulls{hs}},
    \item the design of a \textbf{redundant system} with \textbf{identifiable uncertainties} aiming at a safe vehicle behavior in each \acrshort{hs},
    \item the \textbf{quantification of uncertainties} in the \textbf{system and the environment} per \acrshort{hs} through data collection,
    \item the validation of the system design through \textbf{Stochastic Simulation} of the \acrshort{ads} in the \acrshorts{hs}
\end{itemize}
combining top-down Systems Engineering\footnote{More precisely, \acrfull{mbse}. The key distinction between traditional document-based and model-based systems engineering lies in the approach of managing system information. Traditional document-based systems engineering relies primarily on textual, static, and disconnected documents, while \acrshort{mbse} utilizes semi-formal models to represent system structure, behavior, and requirements. \acrshort{mbse} enables increased automation, integration, and consistency throughout the engineering process, with the model serving as the central repository of system knowledge \cite{friedenthal2014practical}.} and bottom-up Empirical Modeling.

\begin{figure}[htbp]
    \centering
    \small
    \includegraphics[width=1.0\textwidth]{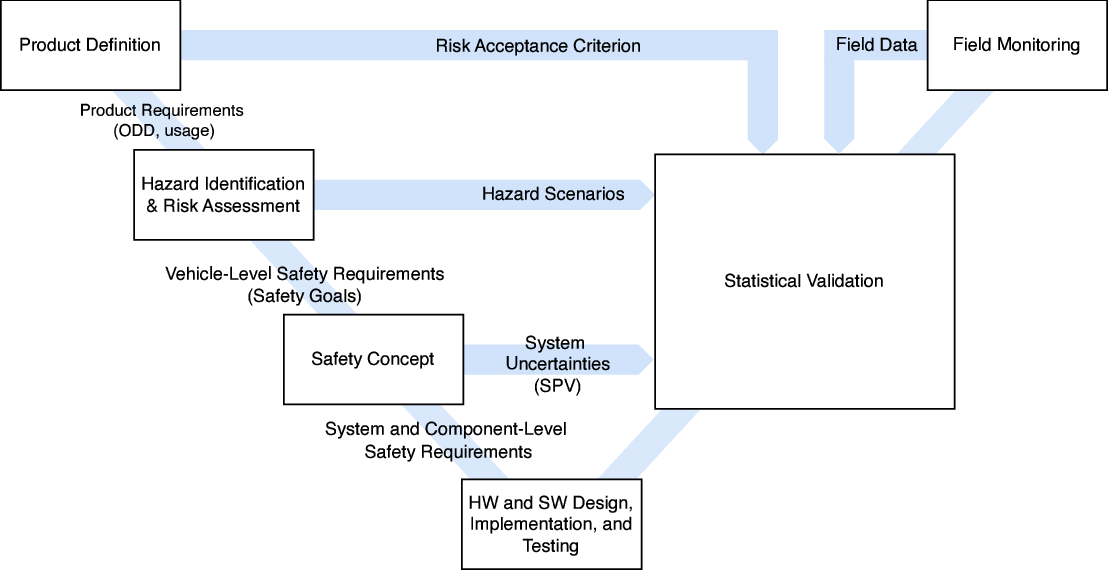}
    \caption{Simplified overview of \acrshort{sifad}}
\label{fig:VV_Overview}
\end{figure}

As can be seen in Fig.~\ref{fig:VV_Overview}, these concepts are embedded into the classical V-model with the following steps:

\subsubsection*{Product Definition (\Sec{\ref{sec:product_definition}})}
The development commences with the definition of the product context including the Operational Design Domain (\acrshort{odd}), the general product behavior with its usage as well the definition of a Risk Acceptance Criterion (\acrshort{rac}). The subsequent safety steps build on these product requirements as they highly influence the system design.
    
\subsubsection*{\acrlong{hira} (\Sec{\ref{sec:safety_goals_and_HS}})}
Based on the Product Definition, a \acrfull{hira} is conducted, aimed at identifying all undesired vehicle behaviors in the various scenarios of the \acrshort{odd}. In general, each undesired behavior has a variety of potential causes,\footnote{E.g., a null pointer dereference in the \acrshort{sw}, a bit flip in the \acrshort{hw}, a classification error of an \acrshort{ml} component, or the residual of a Kalman filter can all lead to a false braking.} but there is no need to differentiate them at this stage.
The identified hazardous behaviors in the respective \acrshorts{hs} are quantitatively assessed and, in case of an unacceptable risk, measures are taken. This leads either to a modification of the Product Definition or \acrfull{tlsr} with risk acceptance targets and integrity levels.

\subsubsection*{Safety Concept (\Sec{\ref{sec:safety_concept}})} 
    The \acrshorts{tlsr} are further broken down into an architectural system design and safety requirements for all elements within. This is a non-trivial, iterative task leading to a comprehensible documentation of the rationale behind the design decisions, hence \quotes{concept}. \\
    A subsequent systematic analysis identifies expected deviations from certain safety requirements related to the algorithms used in the implementation (\quotes{performance insufficiencies}), usually, but not limited to, of the perception system. As part of this process, so-called \acrfulls{spv} are defined to quantitatively capture the uncertainties of these deviations. Notice that \acrshort{hw} and \acrshort{sw} failures (\quotes{bugs}) are not covered as they are addressed by the measures defined in ISO\,26262 in the next step.

\subsubsection*{\acrshort{hw} and \acrshort{sw} Design, Implementation and Testing (\Sec{\ref{sec:physical_level}})}
    Eventually, the \acrshort{hw} and \acrshort{sw} are designed, implemented, integrated, and tested, realizing the architectural design in fulfillment of the safety requirements of the Safety Concept. This is done in accordance with \isofusa,\footnote{Part 4 (6.4.9 Verification, 7.4 System and Item Integration and Testing), Part 5, and Part 6} guided by the previously derived integrity levels.
    For \acrshort{ml} components the development is guided by ISO/PAS\,8800.\footnote{Clause 9 through 12} In the presented framework, independent of the technology, inevitable deviations from the requirements (performance insufficiencies) are accepted as long as they are covered by the associated \acrshorts{spv} in the Safety Concept, as they will be subjected to an in-depth quantitative analysis within the Statistical Validation step.

\subsubsection*{Statistical Validation (\Sec{\ref{sec:validation_concept}})}
    The verification activities in the previous step confirm that the requirements of the Safety Concept have been correctly implemented. However, to assure safety, evidence also needs to be provided that the Safety Concept fulfills the \acrshorts{tlsr}, a process referred to as Safety Validation. \\
    The Safety Concept incorporates in general various architectural measures to robustify the overall system against faults and performance insufficiencies, such as through redundant sensor channels. Therefore, the suitability of the Safety Concept is demonstrated by verifying that the uncertainties of these insufficiencies (\acrshorts{spv}) only propagate to the vehicle level (in the form of TLSR violations) up to the \acrshort{rac}. This process is analogous to the \acrfull{fta} described in ISO\,26262 Part 5 Clause 9, which quantifies the residual risk associated with the uncertainties of random hardware failures. However, unlike random hardware failures, the \acrshorts{spv} are often continuous and dependent on various factors in the environment, such as  weather conditions and the motion of other traffic participants. To address this, the presented framework provides a multi-step approach that combines inductive and deductive methods to systematically derive a combined model of the environment and the system in the \acrshort{hs}, using various data sources. The residual risk of the model is then evaluated by a Stochastic Simulation. During development it is combined with a \acrfull{sa}, which helps identify areas where the system needs improvement (reduction of aleatoric uncertainty) or where more data needs to be collected (reduction of epistemic uncertainty). If the total risk over all \acrshorts{hs} meets the \acrshort{rac}, the Safety Concept is validated, and the \acrshort{ads} can be released. Details are described in \Sec{\ref{sec:validation_concept}}.

\subsubsection*{Field Monitoring (\Sec{\ref{sec:field_monitoring_and_road_clearance}})}
    During operation the system is closely monitored to detect deviations from the modeled system and environmental behavior. If necessary, concrete actions, such as temporary deactivation in certain road sections, can be taken to maintain an acceptable residual risk.

\section{Product Definition} \label{sec:product_definition}\label{sec:risk_acceptance_criterion}
\objective{to establish a clear and agreed-upon understanding of the targeted product behavior from the user's perspective.}

The Product Definition for an \acrshort{ads} serves as an agreement between the Product Management and the Product Development teams. It contains the key stakeholder needs and describes the targeted product behavior, including the top-level degradation functionality, the \acrshort{odd}, and the interaction of the system with the end-user and the environment.
The Product Definition also incorporates the relevant regulatory and compliance needs, such as the \acrfull{alks} standard \cite{UNECE2021Reg157}, that the \acrshort{ads} must adhere to.

\begin{exmp}[Product Definition of a \acrfull{tja}]\label{ex:product_definition}
Let us assume that for an \acrshort{l3} \acrshort{tja} a requirement elicitation process with all stakeholders has brought forward the following, preliminary list of required system behaviors:
\begin{itemize}
    \item Operate in traffic jam situations on German highways
    \item Keep the vehicle in the lane while following the lead vehicle up to 60 kph
\end{itemize}
Moreover, the system shall operate under the following conditions:
\begin{itemize}
    \item During day and night times, and in tunnels
    \item Handle diverging and merging lanes up to a defined maximum slope and minimum width
    \item Require that all vehicle doors are closed and all passengers' seat belts are fastened
    \item (\ldots)
\end{itemize}
The system shall not activate / shall require the driver to take over under the following conditions:
\begin{itemize}
    \item In rain
    \item If the traffic jam resolves
    \item In / before construction sites
    \item Before exiting the highway
    \item (\ldots)
\end{itemize}
If the driver does not take over control within a predefined time after a system request, the automated driving system shall initiate a safe and controlled stop maneuver within the current lane of travel. The same behavior shall be triggered when an abnormal system condition is detected that prevents the system from safely continuing its operation. \\
From a model-based systems engineering's perspective, the teams are at this stage in the problem space (not in the solution space) and they want to express the dynamic, both discrete and continuous \acrshort{ads} behaviors above in a semi-formal way that can be systematically analyzed (see Sec.~\ref{sec:safety_goals_and_HS}). For this kind of task, we suggest using \acrshort{sysml} Activity Diagrams combined with text specifications, cf.\ \cite{delligatti2014sysml}.
\end{exmp}

While the Product Definition does not claim to be comprehensive with respect to safety, it needs some form of a \acrshort{rac}. 
A common choice is the \acrfull{prb}, cf.\ \cite{di2017ethic}. In simple terms, it requires that the activation of the \acrshort{ads} leads on average to a risk reduction, cf.\ \cite{kauffmann2022positive}. \\

The \acrshort{sifad} quantify risk in terms of the average frequency of injuries and fatalities. 
For the definition of these adverse outcomes we refer to the Abbreviated Injury Scale (AIS), which is used worldwide by automotive injury research. It classifies an individual injury by body region according to its severity on a 6 point scale with 1 being minor and 6 being maximal \cite{gennarelli2006ais}.
In this work we simplify this to the scale shown in Table~\ref{tab:il}.
\begin{table}[ht]
\begin{center}
    \begin{tabular}{c c c}
      Injury Level & Injury & AIS \\
    \hline
    \noalign{\smallskip}
    I1+ & light or higher & $\geq$ 1 \\
    I2+ & severe or higher & $\geq$ 3 \\
    I3  & fatal & $\geq$ 5 \\
    \end{tabular}
\caption{Definition of Injury Levels}\label{tab:il}
\end{center}
\end{table}

Notice that the \quotes{+} means \quotes{or higher}, which is for mathematical convenience\footnote{E.g., the probability of an I1+ incident (light injury or higher) monotonically increases with higher impact velocities, whereas the probability of light injuries itself starts to decrease at some point as a result of the higher probabilities of severe and fatal injuries.} within the injury risk models, see \Sec{\ref{sec:injury_risk}}. 

Regarding the \emph{average frequency}, we define for each Injury Level an incident rate\footnote{Instead of failure rates, the \acrfull{mtbf} is often used as the inverse metric \cite{rausand2021system}.}
\begin{equation}\label{equ:lambda}
\lambda_x, \quad x \in \{I_{1+},I_{2+},I_3\}
\end{equation}
as the average number of events that occur per unit of time. Notice that
we focus on self-inflicted crashes.
Collisions caused by the misconduct of other road users that the automated system cannot prevent are excluded from evaluations of the system's performance.
If the activated system results for each Injury Level in lower rates compared to human drivers, cf.\ \cite{otte2003scientific}, the system leads to a \acrshort{prb}. In order to be on the safe side, an additional reduction factor $k_s>1$ is introduced:
\begin{align}\label{equ:prb}
    k_s \cdot \lambda_{x, \text{system}} \overset{!}{<} \lambda_{x, \text{human}}, \quad x  \in \{I_{1+},I_{2+},I_3\}
\end{align}

\begin{exmp}[\acrshort{rac} for slight injuries and higher]\label{ex:rac}
    Based on the targeted product behavior of the \acrshort{tja} above, public accident statistics are evaluated and reported, for the sake of this example, a total of 
    750 slight injuries, 
    140 severe injuries, and 10 fatalities in estimated $10^9$ hours of manual driving. For the Injury Level 2+ (slight injuries and higher), for example, this leads to
\begin{align*}
\lambda_{I_{2+}, \text{human}} = (140+10)/(10^9 h) = 1.5 \times 10^{-7}\,/ h.
\end{align*}
\end{exmp}

\section{From Hazard Identification to Top-Level Safety Requirements}\label{sec:safety_goals_and_HS}
\objective{to systematically analyze the Product Definition for \acrshorts{hs} and to specify for all identified \acrshorts{hs} a safe vehicle-level behavior.
}

A starting point of any risk analysis is the analysis of hazard scenarios \cite{stewart1997probabilistic}, which we refer to in this work as \acrfull{hira}.\footnote{Cf., ISO\,26262-3:2018: hazard analysis and risk assessment (HARA) and ISO\,21448:2022 Clause~6, Identification and evaluation of hazards}

It aims at 
\begin{enumerate}
    \item identifying all combinations of traffic scenarios and vehicle behaviors potentially leading to harm,
    \item assessing their risks, and
    \item deriving mitigation measures.
\end{enumerate}
Before we go through theses steps in more detail, we want to point out some specifics of the proposed process:
\remark{The term \textit{hazard scenario} in a general risk assessment already encompasses system behavior. However, in the \acrshort{ads} context, it is beneficial to exclude the system behavior from the scenario. In the \acrshort{hira}, scenarios and system behaviors are then assessed in pairs, which are also known as \textit{hazardous events}.\footnote{Cf.\ ISO\,26262-1:2018, 3.77 and ISO\,21448:2022, 4.2.1.}}

\remark{The \acrshort{hira} remains at the top-level, the vehicle-level for the \acrshort{ads}, and focuses exclusively on hazardous behaviors irrespective of the underlying causes. 
Whether the hazardous behavior comes from a missing top-level or sub-level requirement, a performance insufficiency, or a hardware/software fault does not need to be differentiated at this stage. This lack of differentiation is intentional because, for most hazardous behaviors of an \acrshort{l3} system, no potential cause can be categorically ruled out. By adopting this approach, the analysis is streamlined and yields consistent results, enhancing the efficiency of the risk assessment process.

\remark{At this stage, we only consider scenario parameters required to assess the consequences of the hazardous behavior --- and not the parameters that could provoke the behavior. For example, while rain can decrease perception and brake performance and therefore increase the likelihood of the hazardous behavior of an  insufficient brake reaction, it does \textit{not} affect the consequences of the hazardous behavior, the collision. This reduces \textit{significantly} the scenario space and makes the \acrshort{hira} tractable, see also next remark.
}

\remark{Throughout the \acrshort{sifad}, the level of abstraction of the scenarios decreases:
The \acrshort{hira} begins with \textit{functional scenarios}, i.e.\ those that are described in natural language supplemented by drawings. The quantitative assessment within the \acrshort{hira} considers the frequency/probability of the scenarios, which in turn require the definition of parameter intervals. This leads to the creation of so-called \textit{logical scenarios}. Later, within Statistical Validation in Section \ref{sec:validation_concept}, more parameters will be systematically identified that influence the system's safety performance. The (joint) distribution of all parameters, required to evaluate the residual risk of the designed system in Sec.\,\ref{sec:validation_concept}, will then replace the parameter intervals.\footnote{The support of distributions should be the same as the intervals for consistency between the development steps.}
Finally, also in Sec.\,\ref{sec:validation_concept}, within the stochastic simulation samples are drawn from these distributions, leading to the generation of so-called \textit{concrete scenarios}. For more details on the different scenario types, please refer to \cite{ISO34501, menzel2018scenarios, ulbrich2015defining}.
}

\subsection{Hazard Identification} \label{sec:hira}
Due to the open context an \acrshort{ads} operates in, even for the comprehensible Product Definition of a \acrshort{tja}, hazard identification is a challenging task. No approach known to us can claim completeness of the screening process. However, systematic elicitation methods and the evaluation of field data during development can be combined to minimize the likelihood of missing a \acrshort{hs}. The key is to reduce the scenario space to the  parameters essential to the assessment of hazardous behaviors -- not their causes. \\
A helpful tool for spanning the scenario space is for example the PEGASUS 6-layer-model \cite{bode2019pegasus} (bottom-up). It can be combined with the modeling of the desired product behavior in \acrshort{sysml}  \cite{friedenthal2014practical, delligatti2014sysml} (top-down).

\subsection{Risk Assessment}\label{sec:risk_assessment}
A commonly used method in a risk assessment is \acrfulls{bn} \cite{fenton2018risk}, which are discussed in more detail in Sec.~\ref{sec:validation_concept}. Figure\,\ref{fig:bn_risk_assessment} shows the graph of a \acrshort{bn} for the simplified assessment of a hazardous behavior.
\begin{figure}[htbp]
    \centering
    \includegraphics[width=0.7\textwidth]{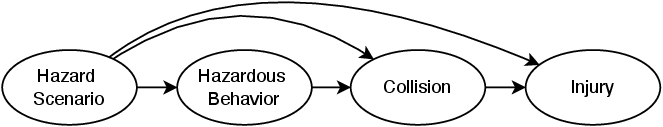}
    \caption{Generic \acrlong{bn} of the risk assessment}
\label{fig:bn_risk_assessment}
\end{figure}
}
The following parameters refer to the nodes of the graph:
\begin{description}
  \item[$\lambda_s$ / $p_s :$ ] rate / probability of the scenario
  \item[$p_b$ / $\lambda_b :$] probability / rate of the behavior given the scenario
  \item[$p_c :$] probability of a collision given the behavior and the scenario
  \item[$p_i :$] probability of an injury given a collision and the scenario
  \item[$\lambda_{i,s}:$] rate of an injury associated with the scenario
\end{description}
In general we distinguish in the framework between two operational modes that can lead to risk:
\begin{enumerate}
    \item In \textit{discrete mode}\footnote{Cf.\ \quotes{exposure frequency} in ISO\,26262,  \quotes{low-demand mode} in IEC\,61508, and \quotes{discrete use} in \cite{fenton2018risk}} the system is stressed due to short-lived scenarios, as illustrated in Fig.~\ref{fig:discrete_mode}. One example is a slower vehicle abruptly entering the lane directly in front of the host vehicle. The Hazardous Behavior of the system would be a delayed reaction or a complete lack of response.
    For the risk quantification, the rate of the scenario $\lambda_s$ and the probability of the behavior $p_b$ given the scenario is required. The associated risk is then given by
    \begin{align}\label{equ:discrete_mode}
        \lambda_{i,s} = (\lambda_s \cdot p_b) \cdot p_c \cdot p_i
    \end{align}
    \begin{figure}[ht]
        \centering
        \includegraphics[width=0.9\textwidth]{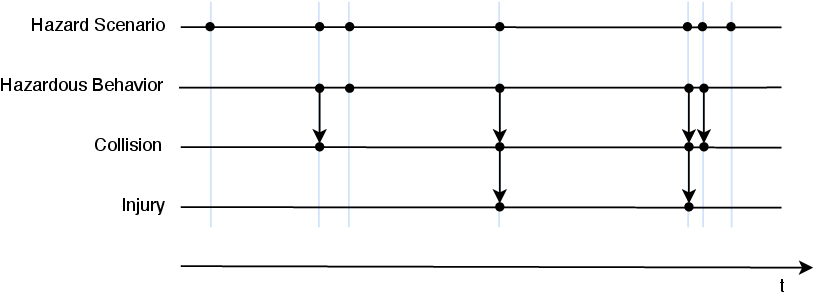}
        \caption{In discrete mode, Hazard Scenarios occur randomly over time, as indicated by the dots on the first horizontal line. These Hazard Scenarios demand a system reaction, which can potentially lead to a Hazardous Behavior, represented by the dots on the second line. This Hazardous Behavior is then reflected with a certain probability in a collision, shown by the dots on the third line. Lastly, this collision can ultimately result in an injury, as indicated by the dots on the fourth line.
        }
        \label{fig:discrete_mode}
    \end{figure}

    \item  In \textit{continuous mode}\footnote{Cf.\ \quotes{exposure duration} in ISO\,26262, \quotes{continuous mode} in IEC\,61508, and \quotes{continuous use} in \cite{fenton2018risk}. Notice that IEC\,61508 refers to operational conditions more frequent than once a year as \quotes{high demand mode} and treats them similar to \quotes{continuous mode}.} the system continuously handles the situation for a certain time and a failure of the system directly leads to a hazard as depicted in Fig.~\ref{fig:continuous_mode}. An example of this is where the \acrshort{ads} follows a truck and the Hazardous Behavior would be inadequate braking or even accelerating. The risk quantification is based on the proportion in the scenario w.r.t. the total driving time $p_s$, and the rate of the behavior $\lambda_b$ given the scenario.\footnote{Cf.\ ISO\,21448:2022, C.2.1} The associated risk can be computed by
    \begin{align}\label{equ:continuous_mode}
        \lambda_{i,s} = (p_s \cdot \lambda_b) \cdot p_c \cdot p_i
    \end{align}
    \begin{figure}[ht]
        \centering
        \includegraphics[width=0.9\textwidth]{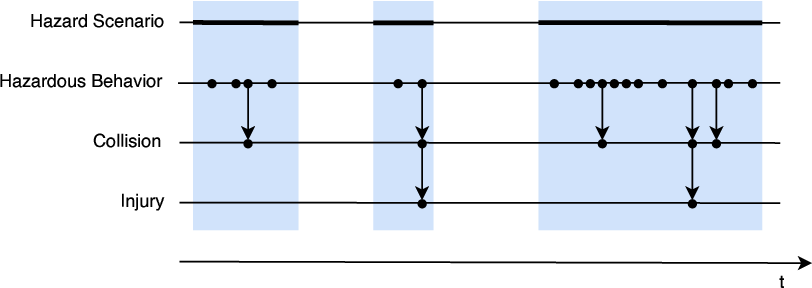}
        \caption{In continuous mode, the Hazard Scenario lasts over an extended period of time, as indicated by the thick sections on the first horizontal line. During this hazard scenario period, a Hazardous Behavior can occur, represented by the dots on the second line. This Hazardous Behavior can then lead to a collision, shown by the dots on the third line. Lastly, this collision can ultimately result in an injury, as indicated by the dots on the fourth line.}
        \label{fig:continuous_mode}
    \end{figure}
\end{enumerate}
At this stage our focus is on deriving the maximum allowed probability $p_b$ or rate $\lambda_b$ for the hazardous behavior given an injury rate budget $\lambda_{i,s}$.
To estimate the required parameters, different data sources need to be drawn upon. The rate or probability of the scenario, $p_s$ and $\lambda_s$, can be estimated based on an evaluation, for example, of customer fleet data, for which the parameter intervals must be precisely specified. For the probability of a collision, $p_c$, closed-form approximations or simulations are required, whereby the knowledge of the development of systems of \acrshort{l2} and below can often be built upon. For the injury probability, $p_i$, accident statistics, complemented by crash simulations, are utilized as they are employed for passive safety, see also Sec.~\ref{sec:empirical_factor_modeling}. Notice that the parameters estimates are required for a first, preliminary analysis, mainly to sort out harmless situations, which do not require any specially vehicle behavior. High risk scenarios, ultimately leading to safety requirements, demand more modeling rigor, see Sec.~\ref{sec:validation_concept}, possibly leading to an update of the above parameters, see Sec.~\ref{sec:iterations}.

The formulas above in combination with the estimated parameters now enable us to predict for a given system performance $p_b$ or $\lambda_b$ the associated risk $\lambda_{i,s}$. Based on this we can make well-founded decisions regarding further development, as outlined in Sec.~\ref{sec:safety_measures}. \\

\begin{rmk}\label{rem:simplified_model}
    The \acrshort{bn} depicted in Fig.~\ref{fig:bn_risk_assessment} often represents a simplified model of the accident. It suggests that the probability of hazardous behavior, subsequent collision, and resulting injury is primarily influenced by the occurrence of a scenario without taking into account additional scenario factors. This simplification may lead to inaccuracies in risk assessment, particularly if the probabilities of multiple nodes depend on a common variable (cf.\ Sec.~\ref{sec:dependent_performance_anaylsis}). These scenarios either require conservative parameter estimates or a breakdown into sub-scenarios. Alternatively, employing a more detailed model could provide a more accurate representation, but this may compromise the ability to directly map the E, C, S of ISO\,26262, cf.\ Remark~\ref{rem:compute_asil} and Sec.~\ref{sec:summary_and_outlook}.
\end{rmk}

\subsection{Safety Measures and Integrity Levels}\label{sec:safety_measures}
\begin{figure}[htbp]
    \centering
    \small
    \includegraphics[width=0.8\textwidth]{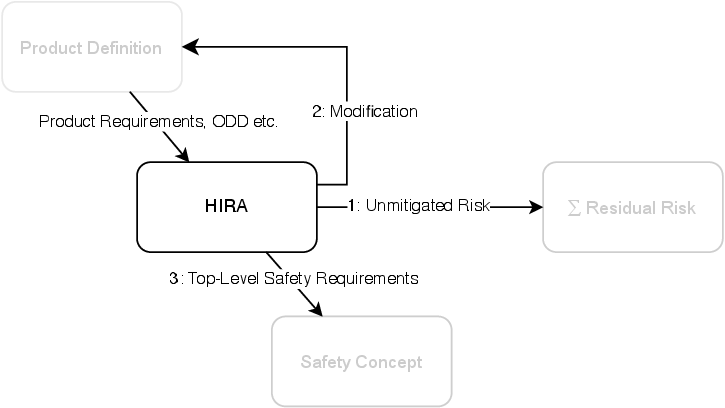}
    \caption{Possible outcomes for hazards in the \acrshort{hira}}
\label{fig:hira_outcome}
\end{figure}
In general, one has three different options to address the risk of the identified hazardous behavior, see Fig.\,\ref{fig:hira_outcome}:
\begin{enumerate}
    \item Accept the unmitigated risk associated with the hazardous event
    
    In case of the discrete mode, we simply set $p_b=1.0$, i.e., the hazardous behavior happens every time the situation occurs, and compute $\lambda_{i,s}$ based on \eqref{equ:discrete_mode}. If it amounts to a small fraction of the \acrshort{rac}, it can be accepted.
    Conversely, continuous mode scenarios must always be addressed by requirements, as they represent a significant portion of operational time.
    
    \item Modify the Product Definition to make the hazardous scenario less likely or the potential consequence less severe

    This measure usually has the largest leverage on risk. On the other hand, it might result in a more stringent \acrshort{rac} derived from a \acrshort{prb}, if human drivers are less involved in accidents in a reduced \acrshort{odd}. After the modification, it is essential to reassess the hazardous events -- including any new or altered ones -- and decide whether to proceed with Option 1 or Option 3.
    
    \item Define \acrlong{tlsr} to be implemented to make the hazardous behavior itself unlikely

    If Option 1 or 2 are not feasible, then a system behavior must be formulated as a vehicle-level requirement\footnote{ISO\,26262:2018 refers to TLSR as Safety Goals.} with the appropriate integrity level so that the hazardous behavior is avoided in the scenario. A hazardous false-negative behavior usually leads to \quotes{The System shall avoid a collision in scenario X. [\acrshort{asil} Y]}. A hazardous false-positive behavior is usually addressed by \quotes{The System shall avoid unnecessary braking above comfort level in scenario Y. [\acrshort{asil} Z]}. The required integrity can be anything between QM and \acrshort{asil} D.
\end{enumerate}

\begin{rmk}\label{rem:compute_asil}
    Quantitative analyses, even if only approximate, often lead to more consistent decisions compared to purely qualitative analyses. Furthermore, in case of an unacceptable unmitigated risk, the assessment can directly provide quantitative requirements for the mitigation measures (Option 3). These quantitative requirements propagate through the system design to its implementation. However, there are currently no broadly accepted methods to quantitatively (statistically) validate software reliability \cite{daniels2022software}, as discussed in Sec.~\ref{sec:physical_level}. For this reason, software safety relies on processes and methods empirically validated by past projects and consolidated in standards, which can vary significantly in their rigor and efforts. In order to chose the right set of measures, the admissible violation probabilities and rates, $p_b$ and $\lambda_b$, are often quantized in powers of 10, and assigned to so-called safety integrity levels (\acrshorts{safetyil}). An example is IEC\,61508:
    The lower the accepted violation rates and probabilities, the higher the integrity level\footnote{Cf.\ IEC\,61508-5, Annex D, Determination of safety integrity levels – A quantitative method} (Table~\ref{table:sil_mapping_iec}) and the greater the effort of the proposed safety methods (Table~\ref{tab:asil_methods}). \\
    In contrast, ISO\,26262 automatically carries out this process through the application of predefined tables\footnote{Cf.\ ISO\,26262-3:2018, Annex B} without any explicit computations of $p_b$ and $\lambda_b$. For compliance, we therefore cannot directly assign the \acrshort{asil} based on $p_b$ and $\lambda_b$. Instead, we need to map $\lambda_s$ or $p_s$, $p_c$, as well as $p_i$ to classes of exposure (E), controllability (C), and severity (S), respectively, which then defines the \acrshort{asil}. Also notice that in the standard $p_c$ may only take the reaction of the driver or other traffic participants into account and not other scenario factors that could avoid a collision. If this leads to an overly conservative estimate, these additional factors can be accounted for in the scenario definition possibly lowering the exposure E. 
    
    \begin{table}[ht]
    \begin{center}
    \begin{tabular}{c c c}
      \acrshort{pfd}$_\text{avg}$ & \acrshort{pfh} &\acrshort{asil} \\
    \hline
    \noalign{\smallskip}
    $10^{-5}$ to $10^{-4}$ & $10^{-9}$ to $10^{-8}$ & 4 \\
    $10^{-4}$ to $10^{-3}$ & $10^{-8}$ to $10^{-7}$ & 3 \\
    $10^{-3}$ to $10^{-2}$ & $10^{-7}$ to $10^{-6}$ & 2 \\
    $10^{-2}$ to $10^{-1}$ & $10^{-6}$ to $10^{-5}$ & 1 \\
    \end{tabular}
    \caption{Quantitative determination of integrity levels of IEC\,61508 (\acrshort{pfd}$_\text{avg}$ $\cong p_s$ and \acrshort{pfh} $\cong \lambda_s$)}
    \label{table:sil_mapping_iec}
    \end{center}
    \end{table}
    
    \begin{table}[h]
    \centering
    \begin{tabular}{c c c c c}
     \acrshort{safetyil} 1 & \acrshort{safetyil} 2 & \acrshort{safetyil} 3 & \acrshort{safetyil} 4 & 
     Recommendation \\
     \hline
     \noalign{\smallskip}
      x & x &   &   & Method 1 \\
        & x & x &   & Method 2 \\
        &   & x & x & Method 3 \\
    \end{tabular}
    \caption{Example qualitative methods recommendations (x) based on integrity levels}
    \label{tab:asil_methods}
    \end{table}
\end{rmk}

\begin{figure} [htbp]
    \centering
    \includegraphics[width=0.5\textwidth]{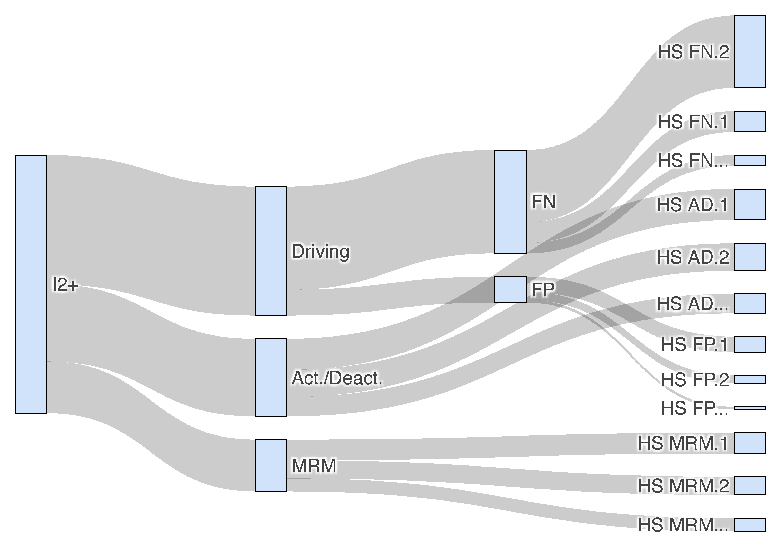}
    \caption{Example I2+ risk distribution}
    \label{fig:sankey_diagram}
\end{figure}

\remark{
    At the start of the project the \acrshort{hira} is preliminary conducted in order to guide the development. In case of Option 2 and 3, a provisional budget must be reserved for the considered hazard as a fraction of the \acrshort{rac} as shown in Fig~\ref{fig:sankey_diagram}. However, in the course of the development, it may turn out that this share has to be shifted between the \acrshorts{hs} depending on the actual safety performance of the implemented system. Also, especially for high-risk scenarios, the parameters in the residual risk analysis must be supported by data, which will be extensively collected during the activities described in Sec.~\ref{sec:validation_concept}. This can, as an example, lead to the need to update an initial integrity classification.
}

\begin{figure}[htbp]
    \centering
    \small
    \includegraphics[width=0.9\textwidth]{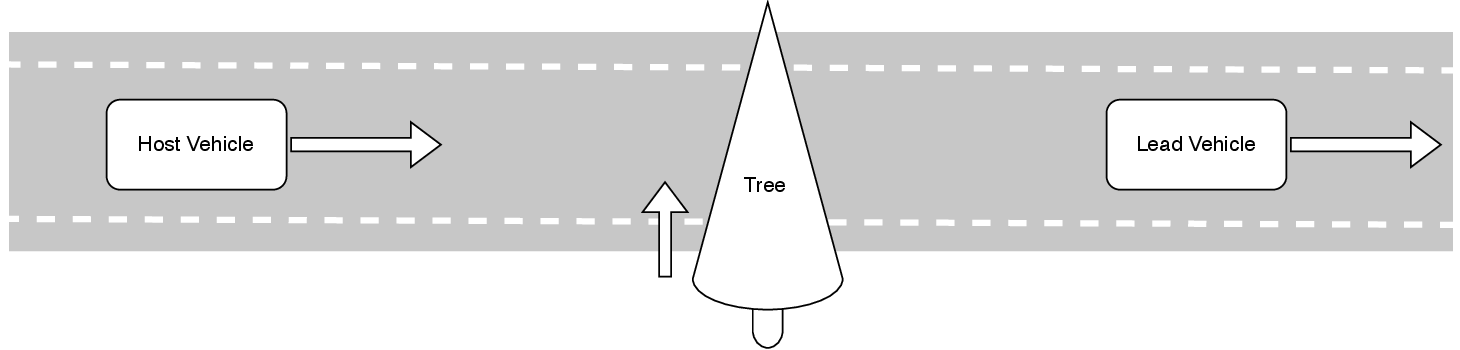}
    \caption{Tree fall scenario leading to an acceptable unmitigated risk}
    \label{fig:hs_tree_topview}
\end{figure}

\begin{exmp}[Accept the unmitigated risk]
Let us assume that the analysis of the Product Definition has identified the \acrshort{hs} of a falling tree as illustrated in Fig.~\ref{fig:hs_tree_topview}. Furthermore, we assume there has been one collapsing tree incident reported on the German Highway within the last 10 years. As a conservative estimate, we assume 1 tree every 5 years  (cf.\ Sec.~\ref{sec:nononformativeprior}). Since the system always follows reliably a lead vehicle with a safe distance, the tree can only do harm to the \acrshort{ads}, when it falls between the lead and the host vehicle. Considering the total number of highway lanes in Germany, the time headway of the system and the average speed in traffic jams, the frequency of hazardous scenario is estimated to be  $\lambda_s \approx 6 \times 10^{-9} / h$. With $p_b = 1.0$ (no system reaction), $p_c = 1.0$ (lane fully blocked by tree, no driver intervention) and $p_i = 0.1$ (for severe injuries), with \eqref{equ:discrete_mode} we get $\lambda_{i,s} = 6 \times 10^{-10} / h$. As we can accept this risk without any measures as part of the overall residual risk, no TLSR is formulated and an \acrshort{asil} ranking is superfluous.\footnote{Due to the low scenario exposure leading to E0, ISO\,26262:2018 (Part 3, 6.4.3.7) does not require a classification at all.}
\end{exmp}


\begin{exmp}[Modification of the Product Definition]
We now consider an \acrshort{l3} system capable of executing lane changes. Based on an estimation of the proportion of scenarios $p_s$ where an inappropriate lane change can lead to an accident (continuous mode\footnote{The operational mode is lane-keeping, which is continuous, whereas the faulty lane change is discrete.}), see Fig.~\ref{fig:hs_lane_change_topview}, and the parameters $p_c$ and $p_i$, as well as a given maximum risk budget $\lambda_{i,s}$ as a fraction of the \acrshort{rac}, we can solve \eqref{equ:continuous_mode} for the required maximum failure rate $\lambda_b$. Let us assume, that it seems unrealistically low given the targeted sensors. We therefore advise against implementing the automatic lane change functionality. In agreement with the product stakeholders we modify the Product Definition accordingly and reiterate.
\begin{figure}[htbp]
    \centering
    \small
    \includegraphics[width=0.9\textwidth]{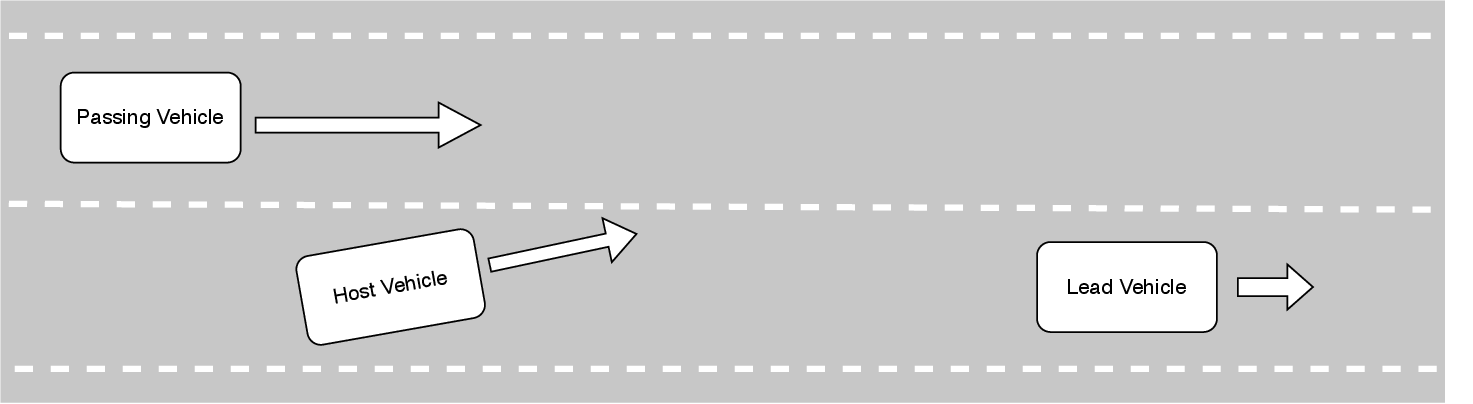}
    \caption{Lane change scenario resulting in a modification of the Product Definition}
    \label{fig:hs_lane_change_topview}
\end{figure}
\end{exmp}

\begin{exmp}[Formulation of a \acrshort{tlsr}]\label{ex:hira_partially_blocked_lane}
Let us assume that the hazardous scenario \quotes{Partially blocked lane} (\acrshort{hs}\,1) was identified. In this scenario the lead vehicle navigates around another vehicle that has come to a stop between the lanes, see \quotes{Intruder} in Fig.~\ref{fig:hs_partially_blocked_topview}. If the host vehicle does not react to the intruding vehicle but continuous driving in the lane center, which is the hazardous behavior to be analyzed, it collides with the blocking vehicle. Due to the short duration of the scenario, we model the discrete mode. 

\begin{figure} [htp]
    \centering
    \small
    \includegraphics[width=0.9\textwidth]{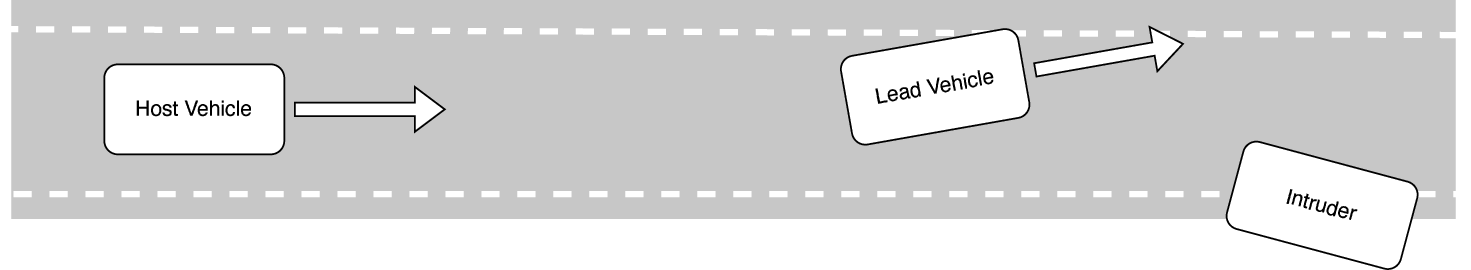}
    \caption{Partially-blocked lane scenario covered by a Top-Level Safety Requirement}
    \label{fig:hs_partially_blocked_topview}
\end{figure}

Let us also assume that the analysis of vehicle fleet data leads to one scenario every 50 operating hours on average, and thus $\lambda_s \approx 2.0 \times 10^{-2} / h$. This requires a definition of additional parameters, such as the maximum velocity and the intrusion depth of the blocking vehicle, and their intervals, so that the scenario can be clearly distinguished from others such as a cut-out with a moving and/or fully blocking vehicle (which is in general easier to detect). As the intruder will not react to the host vehicle to avoid a collision and the scenario is chosen such that without a system reaction the host will always lead to a collision, we get $p_c = 1.0$. Based on the evaluation of accident statistics for small overlap crashes with similar parameters as in the \acrshort{hs}, we estimate $p_i = 0.1$ for $i=\text{I3}$. (The other Injury Levels are handled equivalently.) To determine if the unmitigated risk is acceptable, we set $p_b = 1.0$, that is the hazardous behavior happens every time we encounter the scenario. Eq.~\eqref{equ:discrete_mode} then reveals that the unmitigated risk amounts to
 $\lambda_{i;s} = 2.0 \times 10^{-3} / h$, which is unacceptably high. Assuming that we chose the maximum allowed risk for this scenario to be $1.0 \times 10^{-9} / h$ for I3, we can solve \eqref{equ:discrete_mode} for the allowed failure probability and get $p_b = 5.0 \times 10^{-7}$. This demanding number already indicates that the system design must rely on redundancies, see Ex.~\ref{ex:safety_concept} below. As mentioned above, the number cannot be directly mapped to an \acrshort{asil}. Instead, we have to map the estimated factors to levels of exposure, controllability and severity, respectively, namely E2, C3, and S3, which then lead to an \acrshort{asil} B classification for the hazardous behavior. To avoid it, we require: \\

\qquad TLSR-01: \emph{The system shall avoid a collision with a partially blocking vehicle.} [\acrshort{asil} B]
\end{exmp}

\section{Safety Concept} \label{sec:safety_concept} 
\begin{obj} The main objectives of this section are
    \begin{enumerate}
    \item to ensure that the \acrshort{ads} architectural design and the behavior of its elements is specified in accordance with the \acrshorts{tlsr}, including safety mechanisms as redundancies or degradation mechanisms, and \acrshort{asil} assignment and
    \item to identify and account for the safety-relevant uncertainties within the design.
\end{enumerate}
\end{obj}

In doing so, we not only specify the system behavior to be physically implemented in \acrshort{hw} and \acrshort{sw}, see \Sec{\ref{sec:physical_level}}, but also
support safety-oriented analyses, e.g., in Statistical Validation, see \Sec{\ref{sec:validation_concept}}.

\begin{figure} [htbp]
    \centering
    \small
    \includegraphics[width=1.0\textwidth]{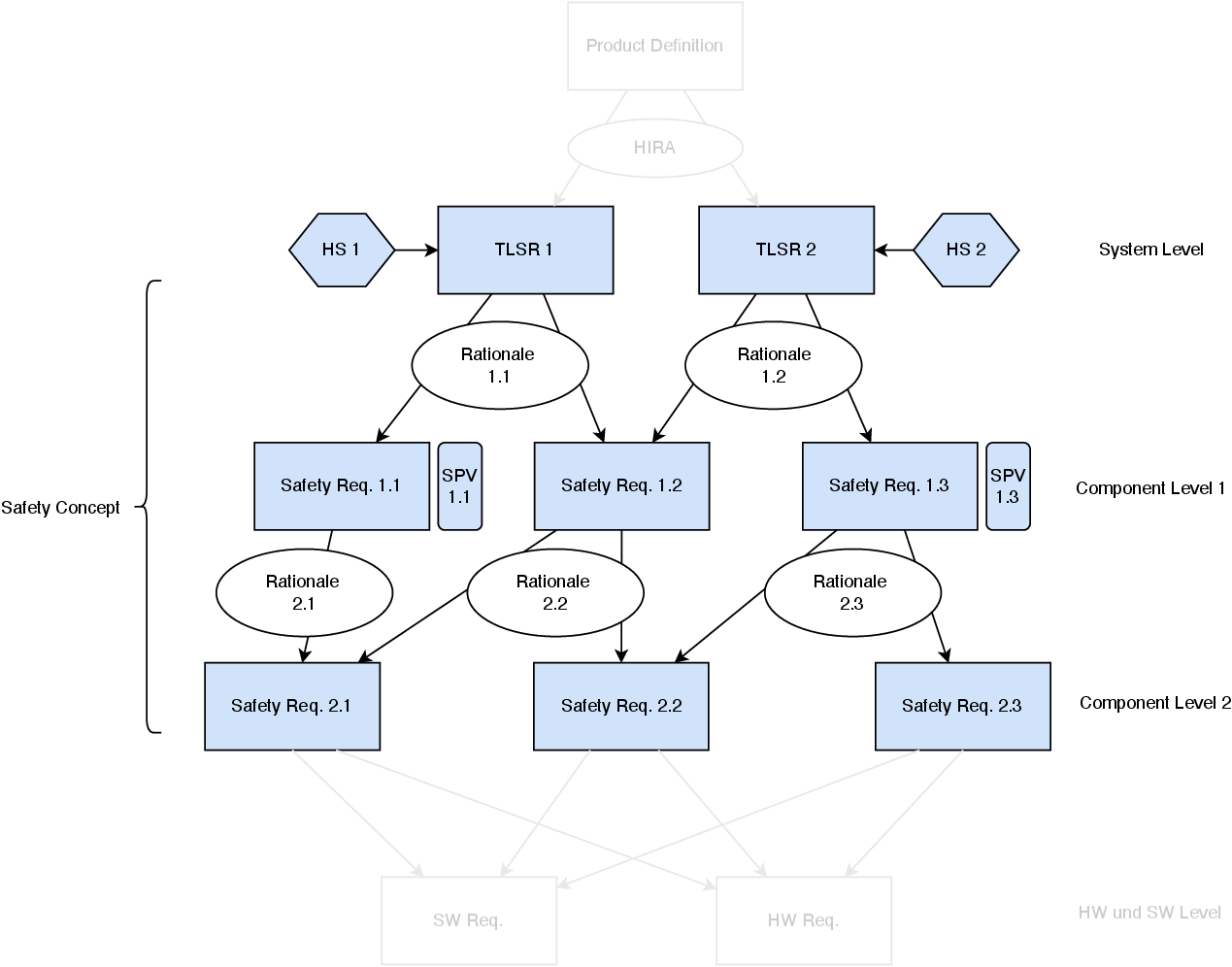}
    \caption{Hierarchical, modular requirement breakdown from the Product Definition over component-level requirements to \acrshort{sw} and \acrshort{hw} requirements}
\label{fig:req_breakdown}
\end{figure}

\subsection{Architectural Design and Requirements} \label{sec:safety_requirements}
The previous section~\ref{sec:safety_goals_and_HS} deals with the problem space by identifying, understanding, and defining the problem that needs to be solved from a safety's perspective resulting in \acrshorts{tlsr}. We now enter the solution space, which focuses on developing technical solutions to realize these safety requirements. The solutions are described in what we refer to as the Safety Concept, which comprises the documentation (incl.\ models, e.g., in \acrshort{sysml}) of the mental representation of the system shared among the development teams. It also includes rationales for the chosen designs essential for an efficient iteration process. The Safety Concept serves as a bridge between the intelligible \acrshorts{tlsr} and the detailed \acrshort{sw} and \acrshort{hw} requirements ensuring that the implemented system with its various components interacting with the environment can be analyzed, verified and validated against the \acrshorts{tlsr}. 

\remark{Generally speaking, a significant amount of \acrshort{sw} faults are not attributed to implementation defects (\quotes{bugs}) but poor or incorrectly specifications of the required system behavior.}

The breakdown of \acrshorts{tlsr} into system requirements is a nontrivial, iterative task due to the many degrees of freedom involved. This process also needs to consider non-safety related requirements such as system availability and comfort. In general, safety properties must be designed into the system from the outset, as they cannot be easily added afterward (\quotes{safe-by-design}).
The Safety Concept of a \acrshort{l3} system typically features safety mechanisms common to lower automation levels, such as
\begin{itemize}
    \item \textbf{Redundancies}: For instance, using a combination of camera, lidar and radar to detect pedestrians under various environmental conditions
    \item \textbf{Layers of protection}: For example, avoiding collisions with the road infrastructure by keeping the lane (layer 1) and avoiding collision with obstacles (layer 2)
    \item \textbf{Fail-safe / fail-operational designs}: For example, detecting heavy rain reduces the vehicle velocity, triggers a driver take-over request or even leads to the execution of a \acrfull{mrm}
\end{itemize}

As illustrated by the examples, these classical mechanisms enhance system robustness not only against faults but also against performance insufficiencies.\footnote{Cf., e.g., ISO\,21448:2022, C.6.3.3} Notably, the components of these mechanisms are allowed to exhibit reduced performance (higher uncertainty) compared to the overall mechanism's performance, which can be estimated with less data, cf.\ Sec.~\ref{sec:safety_performance_variables}, Ex.~\ref{ex:2oo3_aleatory_and_epistemic_uncertainty}, and Sec.~\ref{sec:validation_concept}. \\

Generally speaking, it is beneficial to structure the Safety Concepts by \acrshorts{hs}, which also facilitates the validation of these concepts in Sec.~\ref{sec:validation_concept}. Since many \acrshorts{hs} lead to identical or similar requirements not specific to the treatment of a particular scenario (control system, degradation due to environmental conditions, etc.), a modular structure, as shown in Fig.~\ref{fig:req_breakdown}, avoids the duplication of requirements and identifies potential contradictions in the design early on. Furthermore, introducing additional layers of abstractions (denoted as Component Level 1 and 2) ensures that the hierarchical sub-concepts can be more easily understood and analyzed.\footnote{ISO\,26262:2018 requires two levels of hierarchy for the Safety Concept, the so-called Functional Safety Concept and the Technical Safety Concept.} \\
Since the ultimate goal is to develop a single system that addresses both faults and performance insufficiencies that could lead to a \acrshort{tlsr} violation, and the software components generally contain aspects related to both, a separation into two distinct concepts -- one according to ISO\,21448 and one according to ISO\,26262 -- is not conducive for a \acrshort{l3} system.

\begin{exmp}[Redundancies in the Safety Concept]\label{ex:safety_concept}
In the following, we outline how the system requirements for TLSR-01: \textit{The system shall avoid a collision with a partially blocking vehicle} are derived. The \acrshort{hira}, see Ex.~\ref{ex:hira_partially_blocked_lane}, provides the preliminary maximum failure probability of $p_b = 5.0 \times 10^{-7}$ for the entire system given the scenario. Since this requirement is contingent on the vehicle's perception system detecting the intruder, the perception system's failure probability must be at least as low as $p_b$. The estimation of such a low probability in turn would require about $2.0 \times 10^{6}$ successful trials without fails, cf.\, Sec.~\ref{sec:nononformativeprior}. However, due to the rarity of the scenario, it is infeasible to collect the required amount by field testing. Also, reenacting this scenario on a test track $2.0 \times 10^{6}$ times exceeds any reasonable time frame, not to mention the associated costs. \\
As a solution, the system architecture depicted in Fig.~\ref{fig:2oo3_architecture} is proposed. It employs a \acrshort{too3} voting system where three (mostly) independent channels share the safety-critical load. These channels are based on different sensor principles, radar, camera, and lidar, leading to a heterogeneous redundancy. Each channel monitors the position and speed of objects (Tracking), reads the host vehicle's speed (not shown), and ensures that a speed-dependent safety distance \cite{althoff2016can, shalev2017formal} is maintained (Safety Check) including safety margins for perception and control errors, as well as system delays. If the intruder is detected within the host's lane (Lane Estimation) breaching the safety distance, an emergency braking request is triggered in the individual channels. If at least two of the three channels request emergency braking, the central braking system is activated to perform the emergency brake. \\
As mentioned in Ex.~\ref{ex:2oo3_aleatory_and_epistemic_uncertainty}, this majority voting system enhances safety by reducing not only the risk of missed detections (\acrshorts{fn}) but also erroneous detections of non-existent objects (\acrshorts{fp}) that could trigger an unwarranted emergency stop, which would otherwise compromise another \acrshort{tlsr} related to false-positive braking. \\
The proposed safety mechanism is intentionally kept simple and does not account for comfort. However, that poses no issue because it only comes into action through the Arbiter if the otherwise active Comfort Channel does not respond sufficiently, thereby ensuring that no safety-critical load is placed on the Comfort Channel. 
In addition to the requirements described above, the system's runtime behavior is also specified. \\
Due to space constraints, this example will not delve into additional safety mechanisms, but for the purposes of further discussion, it is assumed that extreme weather conditions and component failures are monitored and lead to a driver takeover or the execution of an \acrshort{mrm}, which renders their occurrence highly improbable in conjunction with the scenario. \\
Lastly, during implementation of the described safety mechanism the set of safety measures defined by ISO\,26262 must be fulfilled so that the \acrshort{asil}-B of the TLSR-01 is assigned to all requirements of the above concept.

\remark{The architecture in example Ex.~\ref{ex:safety_concept} is simplified to highlight the key aspects relevant to the application of the \acrshort{sifad}. The BMW Personal Pilot L3 is based on a slightly more complicated architecture, which further enhances the system availability and user comfort. It incorporates a Comfort Channel that also considers safety constraints based on a fusion of all sensors. If the current vehicle state violates two or more safety constraints, each computed based on one of the individual sensor channels, the system will switch to a safe fallback trajectory, cf.\ \cite{althoff2014online}.
This architecture bears a strong resemblance to the recently introduced Doer/Checker/Fallback \cite{theautonomous2024} and Primary/Guardian/Fallback \cite{shalevsafety} architectures.
}

\remark{Note that the redundant design also effectively prevents \acrshort{sw} and \acrshort{hw} errors in any individual channel from impacting the vehicle's reaction. Consequently, the integrity of the channels could be reduced. However, according to ISO\,26262-9:2018 5.4.4, it is required, that each channel needs to comply with the initial safety requirement by itself to reduce the integrity level (decomposition). Due to the uncertainties present within the channels (\acrshorts{spv}), this is not given. Therefore, even with redundancy, the individual channels must be implemented maintaining their original level of integrity. For further details of this limitation of the ISO, refer to the comment in Sec.~\ref{sec:summary_and_outlook}.
}
    \begin{figure} [htbp]
        \centering
    \includegraphics[width=1.0\textwidth]{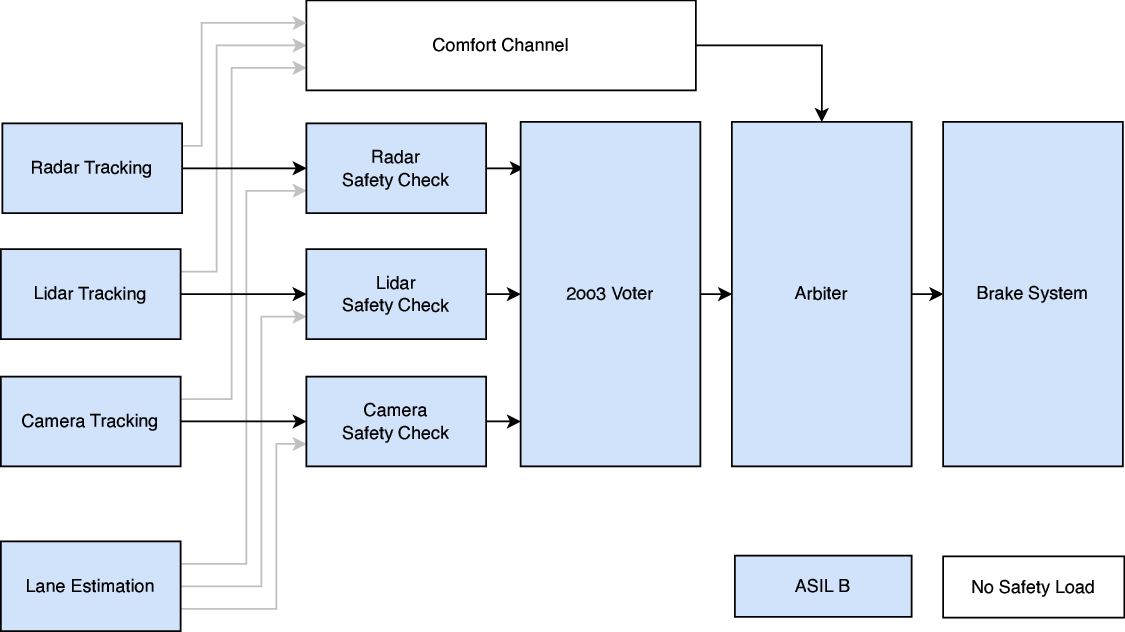}
        \caption{Simplified architectural design to avoid collisions with partially blocking vehicles}
    \label{fig:2oo3_architecture}
    \end{figure}
\end{exmp}

\subsection{Uncertainties in the System}\label{sec:safety_performance_variables}
As previously discussed, the Safety Concept outlines the full effect chain for each \acrshort{tlsr}/\acrshort{hs}, just as the deterministic approach demanded by ISO\,26262 does. This chain extends from sensors to actuators. However, the perception and control modules within the system are influenced by complex interactions with the environment. It is essential to quantify the uncertainties associated with their emergent behavior. For example, the requirements may state that the perception system should detect an object at a certain distance with a specified level of accuracy. During implementation, it may become evident that achieving this requirement with absolute certainty (a probability of 1) is not possible. This leads to the question of whether or not the residual risk is acceptable. In the case of deviations due to random hardware errors, ISO\,26262:2018 Part 5 offers guidance on handling random hardware errors, suggesting, for example, a \acrfull{fta}. However, modeling the error patterns of performance issues and their impact is different and often more involved. It requires a detailed multi-stage process, which we will elaborate on in the Sec.~\ref{sec:validation_concept}. This process requires that all significant uncertainties have to be identified and mathematically defined within the Safety Concept. \\
As is standard in Engineering Risk Analysis, we model system uncertainties using random variables. We refer to them as \acrfulls{spv}, capturing stochastic deviations from the deterministic behavior specified by the requirement, see also \Fig{\ref{fig:req_breakdown}}. \acrshorts{spv} can be either discrete or continuous and are ideally defined at the highest level of the concept and as far back as possible in the chain of dependencies to facilitate the analysis. At this stage, specifying a distribution for the random variables is not necessary.
\remark{The \acrshorts{spv} only encompass performance insufficiencies within the system. 
Deviations from the System Requirements due to \acrshort{sw} and \acrshort{hw} faults are addressed by (mainly qualitative) measures defined by ISO\,26262 for the derived integrity level of the respective TLSR, cf.\ \Sec{\ref{sec:physical_level}.}
}
\begin{exmp}[Uncertainties in the Safety Concept]\label{ex:spv}
Upon reviewing the system architecture shown in Fig.~\ref{fig:2oo3_architecture} and the associated requirements, a team of domain experts has identified that the primary uncertainties within the system reside in the tracking modules of the individual sensors. The risk of a collision arises if an obstructing object (target) is detected too late or not at all by more than one channel.
Another risk presents itself if the target's estimated velocity and position across and along the lane are inaccurate. To circumvent the need to model multiple random variables, a single SPV per channel is defined, which encapsulates all of the aforementioned uncertainties, namely the minimum detection distance at which the target's critical corner is estimated within defined lateral and longitudinal ranges and velocities (During later validation, the maximum conservative error within the ranges is always assumed). The SPV is deliberately kept continuous, as binarization in pass/fail based on a distance threshold would lead to an overly conservative risk estimate. The reason is that a threshold violation would have to be considered as a total failure of the sensor channel, even though in reality, a delayed detection could have led to a significant reduction in collision speed or completely prevented a collision at low initial speeds. \\
Furthermore, the SPV for the Brake System is defined as a pure brake delay computed as a conservative approximation of the maximum position error during braking to standstill. \\
For simplification, we neglect the uncertainties in the Lane Estimation (a redundant system itself based on camera and map/GPS-information) and the host vehicle speed, as well as the jitter in the channels' runtime.
The Safety Checks, the Brake Voter, and the Arbiter are assumed to work deterministically as specified, due to their simplicity.
\end{exmp}

To confirm the thoroughness of the architectural design and requirements, and to ensure that all safety-relevant performance issues are captured by the \acrshorts{spv}, we combine deductive (top-down) and inductive analysis (bottom-up) methods as mandated by ISO\,26262:2018.\footnote{Part 4, Clause 6 and Part 9, Annex A.2} This approach completes the activities of the Safety Concept and we can move on to the implementation. \\
\section{Hardware and Software Design, Implementation and Testing} \label{sec:physical_level}\label{sec:verification_concept}
\begin{obj} The main objectives of this section are
\begin{itemize}
    \item to design and implement the Safety Concept in hardware and software and
    \item to ensure that the concept has been realized with sufficient integrity.
\end{itemize}
\end{obj}
We have reached the bottom of the V-model, where we implement the previously developed Safety Concept in both \acrshort{hw} and \acrshort{sw}. This step involves integration, followed by thorough analysis and testing of the implementation. The development of an \acrshort{ads} is generally data-driven, inevitable when using Machine Learning, so the underlying data sets also carries a safety load. 
The different steps involve significant efforts but we can keep this section short, as the \acrshort{sifad} follows here closely the related recommendations of ISO\,26262\footnote{Part 5 (Development at the hardware level), Part 6 (Development at the software level), Part 8 (Clause 11, Confidence in the use of software tools)} as well as ISO~8800.\footnote{Clause~9 (Selection of AI Technologies, AI Measures and design-related considerations), Clause~10 (Data-related considerations) and Clause~11 (Verification and validation of AI systems)} \\
It is crucial to reiterate at this juncture that the probability of encountering software faults is minimized to an acceptable level by adhering to qualitative (meaning non-statistical) measures that are dependent on the integrity level. \\
For hardware components, a tailored set of qualitative methods is also employed, which varies according to the \acrshort{asil} classification. However, to estimate the average rate of top-level safety requirements being breached due to random hardware failures, also a quantitative Fault Tree or Markov Analysis is conducted. This analysis takes into account the failure rates of components as indicated by established failure rate databases and empirical evidence, as well as the intricacies of the hardware architecture. The analysis results need to confirm that the hardware failure rates comply with the allowed limits set by the respective \acrshort{asil}. \\
In contrast, the current version of the ISO~8800 standard does not provide a distinction in the methods that are pertinent for implementation based on the integrity level.

In addition to the standard measures above, it is imperative in our framework that during the translation of requirements from the Safety Concept into \acrshort{sw} (including Machine Learning), any uncertainty in the system's behavior is covered by the \acrshorts{spv} defined in the Safety Concept.

\begin{exmp}[Training of Sensor Channels]\label{ex:independen_ml_training}
The Safety Concept outlined in Ex.~\ref{ex:safety_concept} requires redundant radar, lidar, and camera tracking channels, which are implemented based on state-of-the-art neural networks. To minimize the likelihood of simultaneous failures, these channels are trained on independent datasets collected in the field. Furthermore, the utilized tooling for training and evaluation is thoroughly analyzed for potential errors, and additional measures are implemented to ensure that any tooling errors do not lead to safety violations in the vehicle. All of this is carried out in accordance with the requirements as specified in ISO/PAS\,8800.
\end{exmp}

\begin{exmp}[Implementation of \acrshort{too3} voter]
    As part of the Safety Concept from Ex.~\ref{ex:safety_concept}, the implementation of the \acrshort{too3} voter must comply with the set of \acrshort{asil}-B measures mandated by ISO\,26262. As an example, this includes performing a static code analysis\footnote{Static code analysis, ISO\,26262-6:2018, Table 7} as well as ensuring full line and branch coverage of the implemented software.\footnote{Statement and branch coverage of ISO\,26262-6:2018, Table 9} \\
    A quantitative evaluation of random \acrshort{hw} failures is not necessary -- according to the standard -- as it only pertains to higher integrity levels.\footnote{Evaluation of probabilistic metric for random hardware failures, ISO\,26262-6:2018, 9.4.2}
\end{exmp}

The mostly qualitative measures referenced in this section are to ensure that both \acrshort{sw} and \acrshort{hw} errors are sufficiently excluded. From experience, this exclusion is up to a probability that is acceptable for the associated \acrshort{asil}. Consequently, the Safety Concept is implemented as specified, and the process of evaluating this is referred to as verification. However, evidence that the concept meets the \acrshorts{tlsr} within the bounds of a tolerable residual risk, especially in the presence of uncertainties (\acrshorts{spv}), is still pending. This step, known as validation, will be addressed in the subsequent chapter.

\section{Validation} \label{sec:validation_concept}
\objective{to confirm that the Safety Concept leads to a sufficiently low residual risk.}

\begin{figure} [htb]
    \centering
\includegraphics[width=1.0\textwidth]{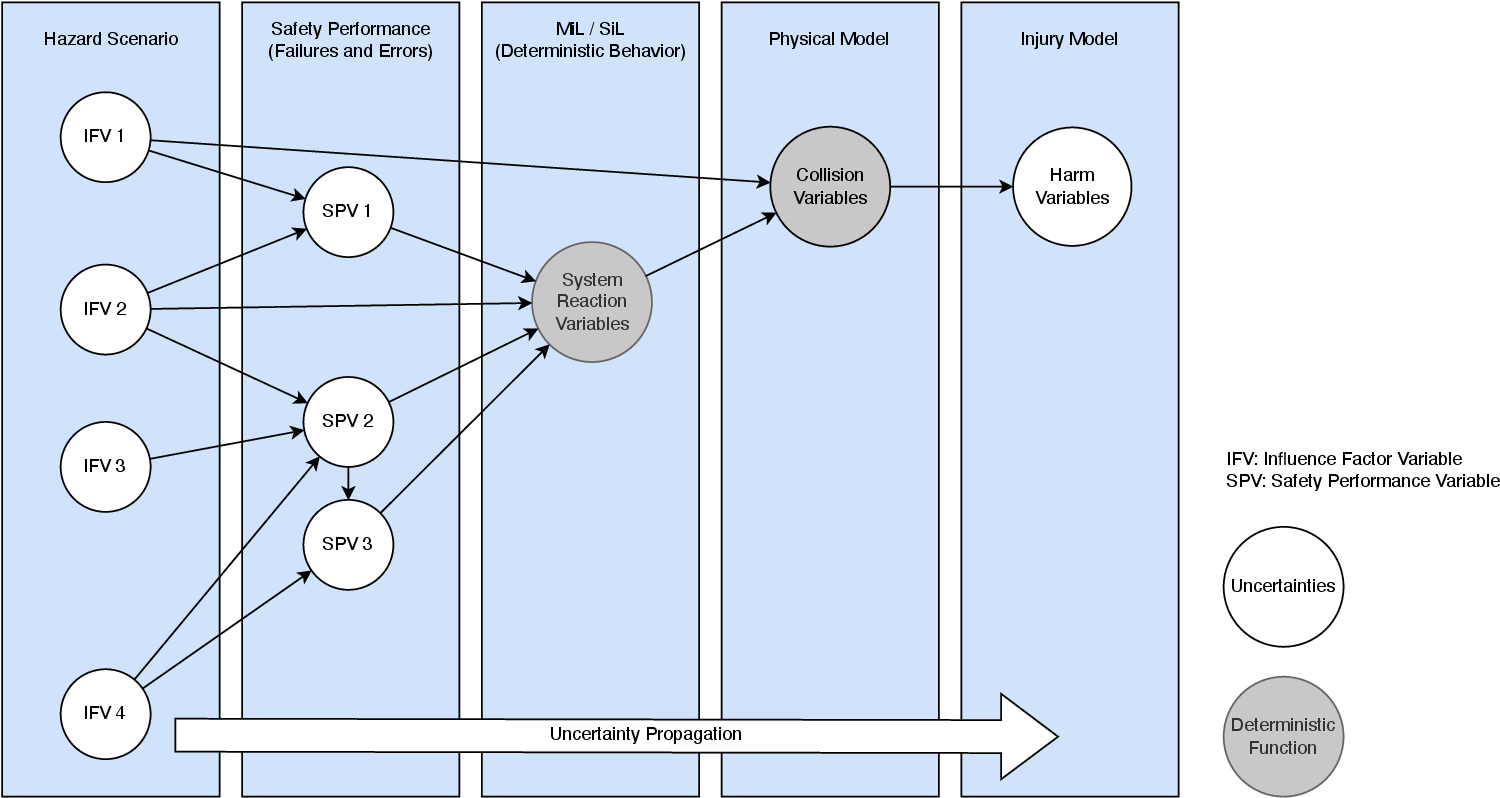}
    \caption{Generic Bayesian Network of a Hazard Scenario}
\label{fig:probabilistic_modelling}
\end{figure}

The validation of the Safety Concept in the \acrshort{sifad} is carried out by analyzing how the uncertainties in the system propagate under the influence of the uncertainties in the \acrshort{hs} to the vehicle's response, possibly leading to collisions and injuries. A residual risk estimate below the tolerated risk then validates the design. \\
As mentioned before, three strategies stand out in the Safety Concept of an \acrshort{ads}: redundancy, layers of protection, and fail-safe / fail-operational designs. These patterns significantly enhance system safety, especially when the associated uncertainties are statistically independent. \\
However, when components operating in parallel share certain influence factors -- known as common causes or common factors -- it can reduce the effectiveness of redundancy, resulting in a high risk:

\begin{exmp}[Dependent redundancies]\label{ex:dependencies_stationary_object}
Consider the example of stationary objects being detected by the radar and lidar system. The algorithms sometimes struggle to differentiate static objects from the background, resulting in a higher probability that both radar and lidar may simultaneously fail to recognize the stationary object.
\end{exmp}
\begin{exmp}[Dependent layers of protection]
Due to their length, object tracking of trucks can be problematic under certain conditions. Additionally, the effectiveness of passive safety measures may differ when colliding with a truck versus a passenger vehicle. Therefore, vehicle tracking performance and the injury resulting from a collision are statistically dependent, leading to an increased overall risk.
\end{exmp}

\begin{exmp}[Dependent fail-safe design]
Consider a system that is intended to detect a failure in the lateral control and then have the longitudinal control system bring the vehicle to a safe stop. Due to the statistical dependency between the lateral and longitudinal systems caused by slippery conditions, the longitudinal control is also impaired and cannot decelerate the vehicle quickly enough, which might lead to a collision despite the fail-safe mechanism.
\end{exmp}

From the perspective of \acrshort{sw} and \acrshort{hw} faults, common causes can often be mitigated through technical measures, an approach referred to as \quotes{freedom from interference}.\footnote{ISO\,26262-6:2018, Annex~D} However, when it comes to performance insufficiencies, the influence of common factors can often be minimized, e.g., by employing different sensor principles, but not entirely eliminated. As such, modeling statistical dependencies is imperative to adequately incorporate them into risk assessments \cite{kluppelberg2014risk}. \\

In statistical terms, for a set of random variables $ X_1, X_2, ..., X_n $ representing the uncertainties in the scenario, the system, and the potential collision, we are required to model the joint distribution $f(X_1, X_2, ..., X_n)$. 
Since a large number of variables are typically involved, it is often not feasible to directly derive a reliable model for their joint distribution with the amount and kind of available data.
However, we can make assumptions, based on expert knowledge and additional experiments, about which variables are directly dependent on each other and which are not. These dependencies can be modeled by a so-called \acrfull{dag}, a graph where the edges have a specific direction, see Fig.~\ref{fig:probabilistic_modelling}. The nodes of the graph represent conditional (and marginal) distributions, which can be estimated based on smaller amounts of data. The combination of a \acrshort{dag} with a conditional distribution for each node therein is referred to as a \acrfull{bn}. With this, the joint distribution can be factorized into a product of conditional distributions,  
\begin{align}
    f(X_1, X_2, ..., X_n) = \prod_{i=1}^{n} f(X_i | \text{pa}(X_i)),
\end{align}
where $ \text{pa}(X_i) $ denotes the parents of the variable  $X_i$ in the \acrshort{dag} \cite{fenton2018risk}.

\begin{exmp}[Joint distribution of a simple BN]\label{ex:simple_joint_dist}
For the continuous random variables $X_1$ to $X_4$, the graph in Fig.~\ref{fig:example_joint_dist} is given. With this, the joint distribution can be factorized as follows:
\begin{align*}
     f(x_1, x_2, x_3, x_4) = f(x_1) \, f(x_2)  \,  f(x_3|x_1, x_2)  \,  f(x_4|x_2, x_3)
\end{align*}

\begin{figure} [htbp]
        \centering
        \small
        \includegraphics[width=.4\textwidth]{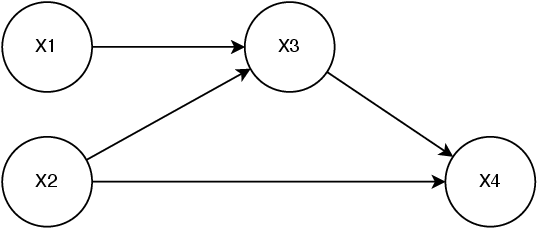}
        \caption{\acrshort{dag} of a simple \acrshort{bn}}
    \label{fig:example_joint_dist}
\end{figure}

The individual conditional and marginal distributions can now be estimated separately. For example, only a data set for $X_1$ is required for the marginal distribution $f(x_1)$, whereas for the conditional distribution $f(x_4|x_2, x_3)$, a joint data set for $X_2, X_3, X_4$ is needed -- without $X_1$.
\end{exmp}

\acrshorts{bn} are widely used in assessing safety-critical applications \cite{fenton2018risk, rausand2021system} and have been effectively applied, e.g., to the modeling of \acrshort{ads} perception \cite{berk2019safety}, environmental influences on the \acrshort{ads} perception \cite{adee2021systematic}, and traffic scenes \cite{jesenski2019generation}. Furthermore, \acrshorts{bn} have proven to be flexible in integrating various data sources and serve as an effective visualization tool simplifying communication among \acrshort{sw} developers, safety analysts, and simulation experts during the development process.

\remark{
    During the early stages of system development, \acrfulls{ft} and \acrfulls{et} \cite{rausand2021system} can be combined to model the causes and the consequences of a \acrshort{tlsr} violation, and therefore provide valuable initial indications of risk. However, these tools often prove insufficient for capturing a detailed system behavior in the \acrshorts{hs}. The key limitation is their reliance on binary variables. Furthermore, they may have difficulty accounting for common causes or coupling effects that are identified during the development process. The transfer of \acrshorts{ft} and \acrshorts{et} to the more flexible \acrshorts{bn} is straight-forward \cite{fenton2018risk}.
}

\begin{rmk}\label{rem:hs_as_cc}
    Experiments show that the performances of sensors strongly depend on the specific scenario under consideration; the scenario itself thus serves as a common cause/factor. Within the framework, the estimated performance models are conditioned on the scenario. This is already an important first step in modeling statistical dependencies. An example is \acrshort{hs}\,1 from Ex.~\ref{ex:hira_partially_blocked_lane}, which incorporates a stationary object that might be harder to detect by multiple sensors than a moving object, cf.\ Ex.~\ref{ex:dependencies_stationary_object}. The detection performance of each sensor channel is then estimated given the object is stationary.
\end{rmk}

\remark{
    The presented approach assumes the availability of extensive scenario data from a vehicle fleet, which sheds light on potential \acrshorts{hs} and their crucial parameters. Typically, the vehicle fleet consists of vehicles that are already operational, outfitted with an earlier generation of sensors. Consequently, while the data obtained from the existing fleet can aid in modeling the scenarios, it is insufficient for predicting the safety performance of the new system with the latest sensor technology. \\
    Therefore, the framework also necessitates a reasonably sized fleet equipped with the latest sensors technology to accumulate new field data. When it comes to measuring performance in connection with rare \acrshorts{hs}, the development fleet must also obtain data through carefully orchestrated test track experiments. These experiments are vital to collect information on how the components perform in rare but potentially high-risk situations, which are typically underrepresented in the field data gathered by the development fleet.
}

\begin{figure} [htbp]
    \centering
    \small
    \includegraphics[width=1.0\textwidth]{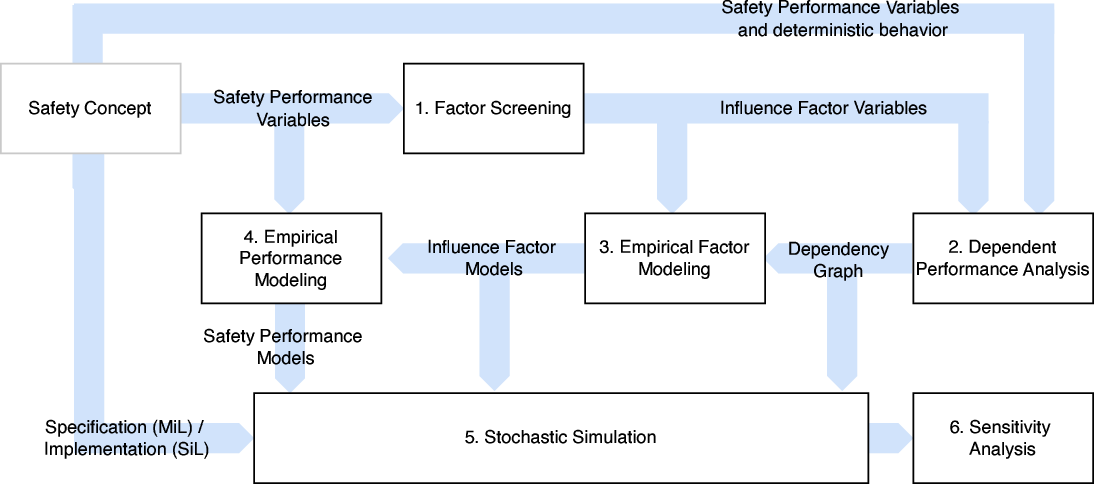}
    \caption{Validation steps for each \acrshort{hs} within the \acrshort{sifad}}
\label{fig:validation_overview}
\end{figure}

The creation of a detailed \acrshort{bn}, cf.\ Fig.~\ref{fig:probabilistic_modelling}, is a non-trivial task and therefore the \acrshort{sifad} divides for each \acrshort{hs} the derivation of this model  into a sequence of incremental steps (1 - 4) shown in Fig.~\ref{fig:validation_overview}. The model is then used in a Stochastic Simulation (5) and analyzed by a Sensitivity Analysis (6). All steps are introduced in the following sections.

\subsection{Factor Screening} \label{sec:factor_screeining}
\objective{to systematically identify the most influential factors of the \acrshort{hs} on the system uncertainties.}
To recap, \acrshorts{spv} represent random variables that quantify the uncertainties present in the system. These uncertainties are typically correlated with various factors in the \acrshort{hs}. For instance:
\begin{itemize}
    \item The color of lane markings can impact the effectiveness of lane detection algorithms.
    \item The size of an object may affect the distance at which it can be detected.
    \item The coefficient of road friction can alter the braking distance of a vehicle.
\end{itemize}
Predicting these factors with certainty is not feasible. Therefore, they are modeled as random variables referred to as \acrfulls{ifv}. \\

Before delving into detailed modeling of a plethora of influences, we make use of the sparsity of effects principle \cite{montgomery2013design}, a concept in statistics and machine learning that suggests that in many real-world problems, only a small number of input variables have a significant effect on the output variable. This principle is based on the observation that complex systems often exhibit a sparse structure, where a few key factors drive the majority of the observed outcomes. That is, most systems are dominated by some of the main effects and low-order interactions, and most high-order interactions are negligible \cite{montgomery2013design}, see below. The identification of the key factors is a fundamental aspect of \acrfull{doe} known as Factor Screening. Essentially, it is a methodological approach used to design and analyze experiments with the goal of pinpointing the factors that significantly affect a system's output. This technique is particularly beneficial in scenarios where there is a multitude of potentially interacting factors. By applying Factor Screening, one can streamline the experiments, concentrating on the factors that exert the most substantial influence.

\remark{
In regression analysis, an SPV is known as the dependent variable, response variable, or endogenous variable, whereas an \acrshort{ifv} is referred to as a factor, an independent variable, a regressor variable, or an exogenous variable. We do not restrict the modeling to binary variables, so we prefer the term Influence Factor Variable over the ISO\,21448 term \quotes{Triggering Conditions} and instead of \quotes{Identification of potential Triggering Conditions} we use Factor Screening as common in engineering risk analysis and experimental design literature  \cite{montgomery2013design}.
}

\begin{figure}[htbp]
    \centering
    \subfigure[Ishikawa diagram]{\label{fig:factor_screeing_fishbone}
        \includegraphics[width=0.45\textwidth]{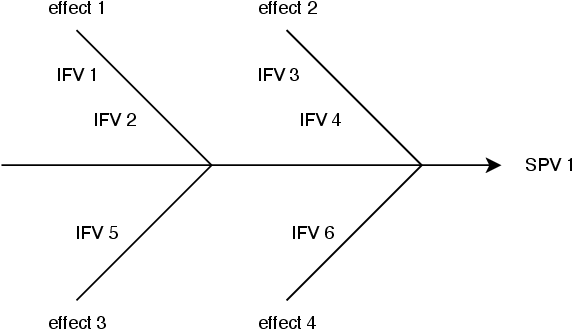}
    }
    \subfigure[\acrshort{ofat} vs. full factorial for three factors]{\label{fig:factor_screeing_factorial}
        \includegraphics[width=0.45\textwidth]{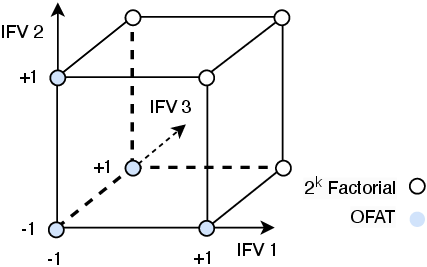}
    }
    \caption{Factor Screening}
    \label{fig:factor_screeing}
\end{figure}

Factor Screening within the framework typically comprises the following steps:
\begin{enumerate}
    \item \textbf{Identify Potential Factors and Ranges}: Starting point of the Factor Screening is an elicitation process\footnote{The PEGASUS 6-layer-model \cite{bode2019pegasus} already mentioned in the \acrshort{hira} in Sec.~\ref{sec:hira} supports a systematic screening similar to the guide words in a \acrfull{hazop} analysis.} that brings forward for each SPV a list of potential \acrshorts{ifv} based on domain expert knowledge and the \acrshort{sw} design. Visual support in the analysis can be given by a so-called Ishikawa diagram shown in Fig.\,\ref{fig:factor_screeing_fishbone}. An analysis of the \acrshort{odd} provides for each potential \acrshort{ifv} the typical range.
    \item \textbf{Design Experiment}: Next, an efficient experimental design is chosen based on the number of factors and their potential interactions such as a full factorial, fractional factorial, Plackett-Burman, or a definitive screening design \cite{montgomery2013design}. The design then needs to be checked and revised for feasibility of each experiment (e.g., rain in tunnel), setup times (optimizing the experimental order as opposed to a fully randomized order), and also for concerns associated to the safety of the personnel conducting the experiment (e.g., definition of point in time when the safety driver should intervene).
    \item \textbf{Conduct Experiments}: Execute the designed experiments on a test track with test drivers, dummy targets, road infrastructure, etc., and account for uncontrollable variables like weather conditions through timing of the experiment and the location of the test ground. During execution, we record all input data like sensor measurements as well as ground truth information that are required to reprocess the component's software and to evaluate the \acrshorts{spv} offline, so that the experiments do not have to be repeated with every \acrshort{sw} version. 
    \item \textbf{Reprocess \acrshort{sw} and compute \acrshorts{spv}}: The recorded experiments are then reprocessed with the \acrshort{sw}, i.e., the input data is fed into the \acrshort{sw} and the output is used to compute the \acrshorts{spv} according to their definitions. If certain ground truth data is not available, typically for technical reasons, a labeling process needs to precede this step.
    \item \textbf{Analyze Results and Select Factors}: In a last step, the factors are chosen, usually iteratively  (e.g., backward-elimination) and in general based on a linear regression analysis \cite{montgomery2013design}. During this process the domain experts do not rely solely on the numerical results, but are supported by accompanying plots providing valuable insights regarding the underlying model assumptions and to make a final decision on the \acrshorts{ifv} to be considered in the \acrshort{bn}. In case of doubt, the respective factors are included or more experiments are initiated.
\end{enumerate}
\remark{The general Factor Screening methodology does not differentiate between factors inside and outside the system. If an influence from an internal source is potentially relevant, such as another SPV, it will be analyzed as well. An example could be the influence of the velocity estimation error (\quotes{odometry}) on the obstacle detection distance. However, instead of controlling the influence during the experiment, the effect can often be injected during reprocessing.}
\remark{For frequent \acrshorts{hs} the development fleet data might already suffice to identify the \acrshorts{ifv}. This is known as an observational study and the evaluation is typically the same as the one of a designed experiment, see below. However, special attention is required, when the \acrshorts{ifv} are strongly correlated (\quotes{multicollinearity}) \cite{montgomery2013design}.}
\remark{If a factor is weak but influences multiple \acrshorts{spv} (\quotes{coupling factor}) within a redundant architecture, the factor should become an \acrshort{ifv}, because it might influence the overall risk, cf.\ Sec.\,\ref{sec:dependent_performance_anaylsis}.}

\newcommand{\pone}{\textcolor[rgb]{0.2,0.5,0.8}{ +1}}
\newcommand{\mone}{-1}
\begin{exmp}[\acrfull{ofat}]\label{ex:ofat}
    In order to determine the dominant factors influencing the SPV \quotes{detection distance} of a particular channel from Ex.~\ref{ex:spv}, it is presumed that the dedicated team of sensor specialists has created a list of potential factors in the \acrshort{hs}. As displayed in Table~\ref{tab:factors_and_ranges}, the team also derives for each factor the low (-1) and high (+1) levels for the \acrshort{hs}. \\
\begin{figure} [htbp]
    \centering
    \includegraphics[width=.7\textwidth]{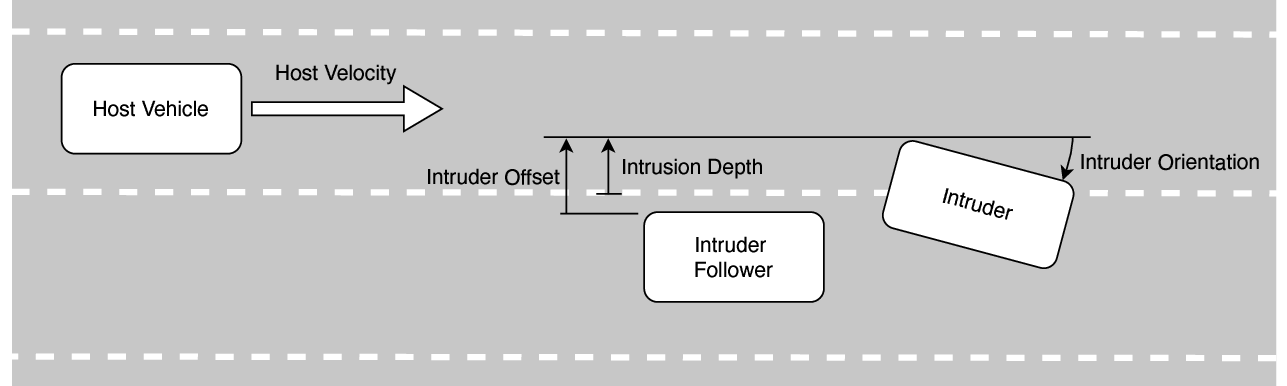}
    \caption{Definitions of sensor-related \acrshorts{ifv} in Table~\ref{tab:factors_and_ranges}. The \textit{Intrusion Depth} is another random variable in the \acrshort{bn} of \acrshort{hs}~1.}
\label{fig:hs_partially_blocked_variables}
\end{figure}

\begin{table}[ht]
\centering
\begin{tabular}{lccc}
Name & Variable &  Low (-1) & High (+1) \\
\hline
\noalign{\smallskip}
\textit{Intruder Offset} & $X_1$ & .5 m & 1.5 m \\
\textit{Intruder Orientation} & $X_2$ & 0.0 & 30.0° \\
\textit{Host Velocity} & $X_3$ & 8.0 m/s & 17.0 m/s \\
\textit{Precipitation} & $X_4$ & 0.0 mm/h &  3.0 mm/h \\
\end{tabular}
\caption{Factor names and levels for Ex.~\ref{ex:ofat} and \ref{ex:factorial}}
\label{tab:factors_and_ranges}
\end{table}

As the team is in the process of becoming acquainted with the \acrfull{doe} methodology and the required tooling, the \acrshort{ofat} method will be utilized as a starting point. In a first step, the following potential influences, see also Fig.~\ref{fig:hs_partially_blocked_variables}, are considered with different motivations:
\begin{itemize}
    \item \textit{Intruder Offset:} The larger the offset, the more of the intruder is visible.
    \item \textit{Intruder Orientation:} The stronger the twist, the smaller the reflected signal.
    \item \textit{Vehicle Speed:} The faster the vehicle, the further the host vehicle travels during the delay on the output, e.g., due to debouncing.\footnote{filtering out rapid changes in sensor readings to reduce \acrshorts{fp}}
    \item \textit{Precipitation:} The more water droplets in the air, the more damping of the sensor signal.
\end{itemize}
The response of the SPV is evaluated with all variables at their low level, and then measured again as one variable is changed to its high value while keeping all other variables at their low level, see blue points in Fig.\,\ref{fig:factor_screeing_factorial}. This can be coded in the design matrix
\begin{align*}
 \mathbf{D }= 
\begin{blockarray}{cccc}
x_1 & x_2 & x_3 & x_4 \\
\begin{block}{(cccc)}
  \mone & \mone & \mone & \mone\\
  \pone & \mone & \mone & \mone\\
  \mone & \pone & \mone & \mone\\
  \mone & \mone & \pone & \mone\\
  \mone & \mone & \mone & \pone\\
\end{block}
\end{blockarray} \, ,
\end{align*}
where each line represents an experiment. 
Due to the inherent randomness of the SPV, the influence of a factor cannot be estimated based on a single experiment. Therefore, the team replicates each combination 20 times leading to a total of 100 experiments. After conducting the experiments and reprocessing the \acrshort{sw} each experiment provides a value $y$ for SPV. Based on the response model 
\begin{align}\label{equ:response}
    y = \beta_0 + \beta_1 x_1 + \beta_2 x_2 + \beta_3 x_3 + \beta_4 x_4 + \epsilon = \beta_0 + \sum_{i=1}^{4} \beta_i x_i + \epsilon, \quad \epsilon \sim \mathcal{N}(0, \sigma^2),
\end{align}
i.e., a linear combination of the influences plus an unexplained, normally distributed error $\epsilon$ with zero mean, one can compute the maximum likelihood values for the regression coefficients $\beta_i$ as well as their 95\% \acrfull{ci}, see black lines in Fig.\,\ref{fig:regression_coefficients_ofat}. If the \acrshort{ci} for a specific regression coefficient does not include zero, one can reject the null hypothesis that the associated term has no significant influence, at a 95\% confidence level. 
We would therefore conclude that the influences $X_1$ and $X_3$ are significant, whereas $X_2$ and $X_4$ probably are not.
\begin{figure} [htbp]
    \centering
    \small
    \includegraphics[width=1.0\textwidth]{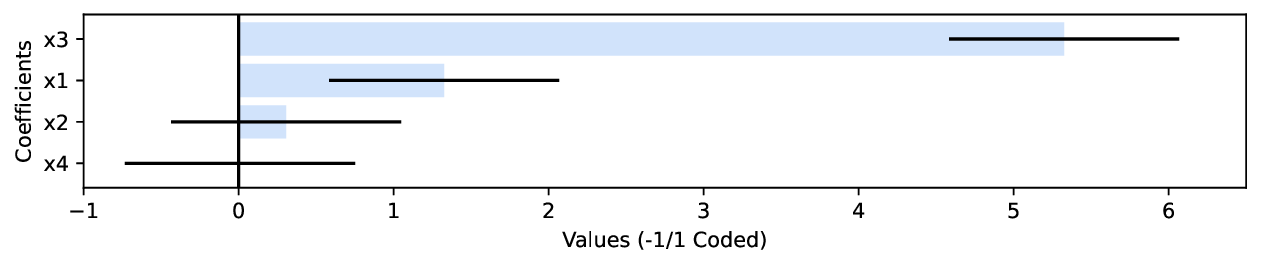}
    \caption{Pareto-chart of coefficients with their 95\% \acrshorts{ci} based on a \acrshort{ofat} design. Notice that it is common practice to not label the regression coefficients as $\beta_x$, but instead, to display the regressor expressions of the independent variables that the coefficient is multiplied with, i.e., $x_1$ instead of $\beta_1$.}
\label{fig:regression_coefficients_ofat}
\end{figure}
\end{exmp}

\begin{figure} [htbp]
    \centering
    \small
    \includegraphics[width=1.0\textwidth]{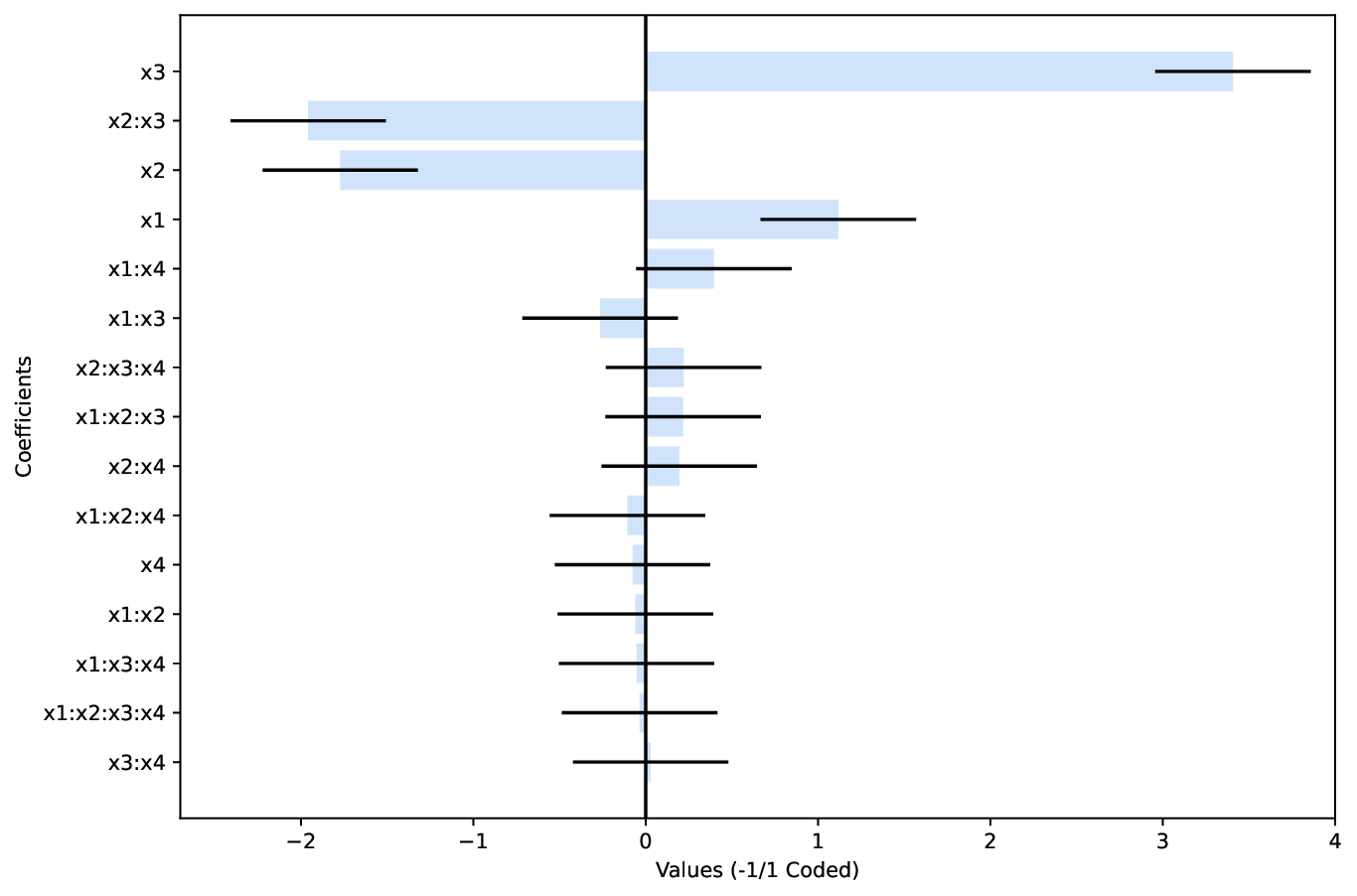}
    \caption{Pareto-chart of coefficients with their 95\% \acrshorts{ci} based on a $2^k$ factorial design. Notice that the colon sign (:) represents multiplication and therefore, e.g., x1:x2:x3 denotes the coefficient $\beta_{123}$ in \eqref{eq:interactions}.}
\label{fig:regression_coefficients_fullfaktorial}
\end{figure}

\begin{exmp}[$2^k$ factorial design]\label{ex:factorial}
    The confidence intervals for the \acrshort{ofat} design remain relatively large despite 100 experiments, see Fig.~\ref{fig:regression_coefficients_ofat}. Furthermore, the design cannot identify interactions between factors. Ultimately, it may be that a factor only leads to an effect in combination with another factor. Therefore, the team decides to try a another design for the same problem, a $2^k$ factorial design with $k$ being the number of factors. As implied by its designation, this design yields $2^4=16$ unique combinations. This remains manageable owing to the relatively limited number of factors involved. The design matrix is then structured in the following manner:
    \begin{align*}
        \mathbf{D}= 
    \begin{blockarray}{cccc}
    x_1 & x_2 & x_3 & x_4 \\
    \begin{block}{(cccc)}
      \mone & \mone & \mone & \mone \\
      \pone & \mone & \mone & \mone \\
      \mone & \pone & \mone & \mone \\
      \pone & \pone & \mone & \mone \\
      \mone & \mone & \pone & \mone \\
      \pone & \mone & \pone & \mone \\
      \mone & \pone & \pone & \mone \\
      \pone & \pone & \pone & \mone \\
      \mone & \mone & \mone & \pone \\
      \pone & \mone & \mone & \pone \\
      \mone & \pone & \mone & \pone \\
      \pone & \pone & \mone & \pone \\
      \mone & \mone & \pone & \pone \\
      \pone & \mone & \pone & \pone \\
      \mone & \pone & \pone & \pone \\
      \pone & \pone & \pone & \pone \\
    \end{block}
    \end{blockarray}
    \end{align*}
    To make the result comparable to the \acrshort{ofat} design with 100 experiments, the team chooses for each combination 6 replications leading to a total of 96 experiments. Since the design includes all combinations of factors, the model can now also incorporate interaction terms: 
    \begin{align}\label{eq:interactions}
        y = \beta_0 + \underbrace{\sum_{i=1}^{4} \beta_i x_i}_{\text{main effects}} + \underbrace{\beta_{12} x_1 x_2 + \beta_{13} x_1 x_3 + \ldots + \beta_{34}x_3 x_4}_{\text{two-factor interactions}} + \underbrace{\beta_{123} x_1 x_2 x_3 
     + (\ldots) }_{\text{higher order interactions}} + \,\epsilon , \quad \epsilon \sim \mathcal{N}(0, \sigma^2)
    \end{align}
    After completing the 96 experiments of the factorial design, the team analyzes the regression coefficients, now including also the interaction effects, as depicted in Fig.~\ref{fig:regression_coefficients_fullfaktorial}. The analysis leads to the conclusion that the main effects $X_1$ and $X_3$ significantly affect the response, which was also concluded from the \acrshort{ofat} design. However, also the the main effect $X_2$ and the interaction between $X_2$ and $X_3$ was found to be significant, indicating the necessity to include $x_2$ as an \acrshort{ifv} in the \acrshort{bn}, an influence not detectable in the \acrshort{ofat} design. Moreover, the \acrshorts{ci} have been narrowed by approximately $\sim$33\% using the $2^k$ factorial design, despite a slight reduction in the number of experiments. 
\end{exmp}

\remark{The above two examples are based on random samples drawn from 
\begin{equation*}
    y = 3 + x_1 - 2 x_2 + 3 x_3 - 2 x_2 x_3  + \,\epsilon , \quad \epsilon \sim \mathcal{N}(0, 1).
\end{equation*}
 The $2^k$ factorial design therefore leads to the right conclusion. However, since there is noise involved in each experiment and the number of experiments is finite, statistical methods can never confirm that a certain variable has exactly zero influence. We can only conclude, e.g., by looking at the confidence intervals of the -1/+1 coded variables as in the examples above, that a particular influence can be neglected in comparison to other influences. If the intervals are still too large to make a decision, more experiments need to be conducted.
}

The examples above emphasize the importance of an efficient design, which becomes increasingly relevant as the number of factors increases. In many cases, the full factorial design is then impractical. However, based on expert knowledge, the model structure (such as interaction terms, etc.) can often be constrained (see also the \quotes{sparsity of effects principle} mentioned earlier), which drastically reduces the testing effort. More specifically, there are so-called D-optimal algorithms that exploit the structure while defining the experiments to minimize the variance of the regression coefficients. In case of a first-order model with two-factor interactions, the $2^k$ factorial design above is already optimal. However, there are situations in which some combination of factors cannot be tested, in which these D-optimal algorithms iteratively modify the design accordingly. \\

Confidence intervals provide insights into the precision of parameter estimates for a given model such as \eqref{equ:response}. However, they do not reflect the overall suitability of the model. To assess the goodness of fit, a separate evaluation, such as a lack-of-fit test, should be conducted \cite{montgomery2013design}. We will elaborate on this topic in more detail in Sec.~\ref{sec:empirical_performance_modeling}. \\

The overview above and the examples are meant to provide a good intuition on the approach, challenges, and methods of Factor Screening. However, for a more comprehensive understanding, we refer to \cite{montgomery2013design}, which covers various aspects of experimental design with high relevance to \acrshort{ads} (randomization of experiments, split-plot designs, fitting of regression models, model diagnostics, categorical factors, transformations, etc.) in detail.

\subsection{Dependent Performance Analysis}
\label{sec:dependent_performance_anaylsis}
\objective{to model the dependencies of the uncertainties in the system, the scenario and the collision.}

Through Factor Screening in Sec.~\ref{sec:factor_screeining} we have determined which \acrshorts{ifv} need to be considered for each \acrshort{spv}. 
Additionally, accident statistics can be used, as common for \acrshorts{ads} of \acrshort{l2} and below, to identify the risk factors that predict the severity of injuries in a collision \cite{nishimoto2017serious}. These factors typically include impact speed, vehicle masses, angle of attack, and must be also included in the graph.
Once all the required random variables of the \acrshort{bn} are defined, the next step is to model their relationships. For this, the directed edges between the nodes need to be determined. \\

In some cases, \acrshorts{ifv} can be modeled independently of each other, leading to leaf nodes, i.e., nodes without parents, in the graph. In others cases, statistical independence cannot be assumed and neglecting the dependency can lead to both overly conservative or optimistic risk estimations \cite{kluppelberg2014risk}.

\begin{exmp}[Dependent Perception and Brake Performance]\label{ex:independen_ifvs}
Modeling rain intensity and the road's friction coefficient independently could underestimate the risk. This is because rain not only reduces detection distances of obstacles but also lowers the friction coefficient, which in turn affects the braking performance. Therefore, shorter detection distance occur more often in combination with larger stop distances increasing the probability of collision with obstacles.
\end{exmp}

\begin{figure}[htb]
\centering
\subfigure[Joint distribution]{\label{fig:bn_joint_distribution}\includegraphics[width=30mm]{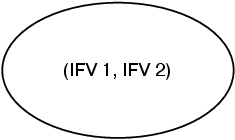}}
 \hspace{5mm}
\subfigure[Simple dependence]{\label{fig:bn_direct_dependence}\includegraphics[width=45mm]{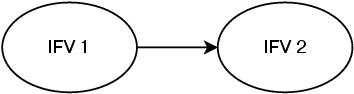}}
 \hspace{5mm}
\subfigure[Coupling factor]{\label{fig:bn_coupling_factor}\includegraphics[width=45mm]{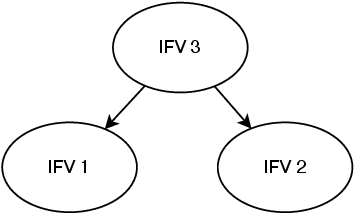}}
\caption{Node dependencies in a \acrshort{bn}}
\end{figure}

If sufficient data is available, we can model the joint distribution of the \acrshorts{ifv} (or a subset thereof), as shown in Fig.~\ref{fig:bn_joint_distribution}, using advanced methods, as described in Sec.~\ref{sec:empirical_factor_modeling}. When a joint dataset for the dependent variables is not available, or the data amount is insufficient, it is advisable to make the dependencies explicit through causal relations \cite{rausand2021system, neurohr2021criticality}, cf.\ Fig.~\ref{fig:bn_direct_dependence}. More concretely, for causal relations, the direction of the edges in the \acrshort{dag} should be from cause to effect \cite{fenton2018risk}. Introducing additional nodes for common causes / coupling factors might be also be an option in many cases, cf.\ IFV 3 in Fig.~\ref{fig:bn_coupling_factor}.

\begin{exmp}[Graphical structure for \acrshort{hs}\,1]
In the following we derive the \acrshort{dag} in Fig.~\ref{fig:dag_partially_blocked_lane}.
 According to Ex.~\ref{ex:spv}, the \acrshorts{spv} for the perception system of \acrshort{hs}\,1 are the \textit{Detection Distances} of the three sensor channels. Ex.~\ref{ex:factorial} provides the \acrshorts{ifv} \textit{Intruder Orientation, Intruder Offset}, and \textit{Host Velocity} for one of the channels, and we assume for simplicity that all three channels exhibit the same dependencies. \\
 Furthermore, the uncertainty in the control system, as described in Ex.~\ref{ex:spv}, is captured by the \textit{Longitudinal Control Error}. For simplicity, we assume that the control error only depends on the braking deceleration. Since we consider a constant brake reaction of $-7.0 m/s^2$, the error is modeled as a marginal distribution. \\
Additionally, we employ an injury risk model based on \cite{nishimoto2017serious}, which requires 
\textit{Collision Speed}, \textit{Intruder Mass}, and \textit{Intrusion Depth}, cf.\ Ex.~\ref{ex:combined_injury} for details. \\
The customer fleet provides data for all \acrshorts{ifv}, see Ex.~\ref{ex:joint_ifvs} below, with the exception of the \textit{Intruder Mass}. We therefore introduce \textit{Intruder Type}, a categorical random variable, that can take the values \textit{truck} and \textit{car}, a classification provided by the customer fleet data. If we assume that the \textit{Intruder Mass} is independent of all other \acrshorts{ifv} for a given \textit{Intruder Type}, we can model the \textit{Intruder Mass} conditional on the \textit{Intruder Type}, for which data sources are available \cite{otte2003scientific}. Furthermore, we simplistically assume that the host mass is constant and known, and therefore constitutes a parameter in the \textit{Injury} node. \\
Lastly, we know from the Safety Concept that the \textit{Collision Speed} depends on the system reaction, which in turn is influenced by all \acrshorts{spv}, the \textit{Host Velocity} as well as the Intrusion Depth, cf.\ Ex.~\ref{ex:functional_pyhsical} below.

\begin{figure}[htb]
    \centering
    \includegraphics[width=.9\textwidth]{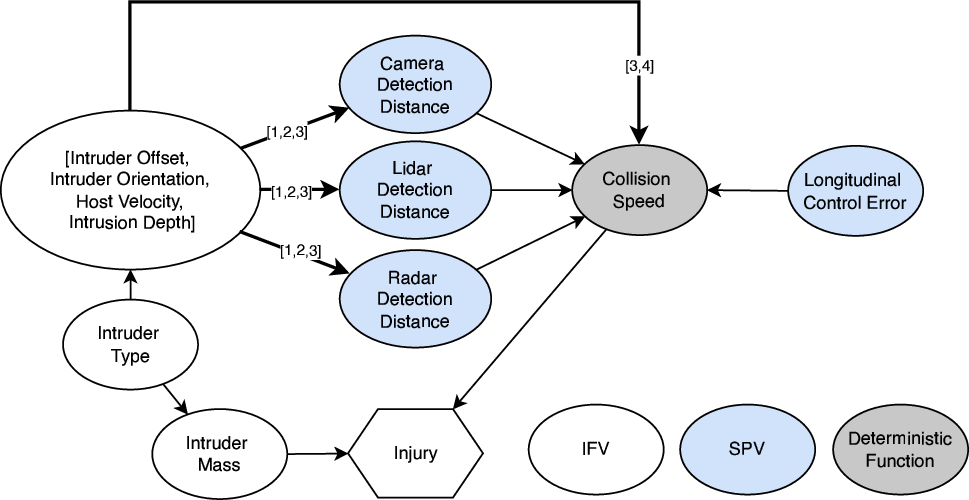}
    \caption{\acrshort{dag} of \acrshort{hs} Partially Blocked Lane. Notice that the child nodes of the random vector (largest node) depend only on the components that are specified in square brackets on the edge using 1-based indexing.} \label{fig:dag_partially_blocked_lane}
\end{figure}

\end{exmp}
In contrast to the simplified example above, in practice, \acrshorts{bn} become increasingly detailed over time. In such cases, Object-Oriented Bayesian Networks (OOBNs) are particularly suitable \cite{koller2009probabilistic}. They break down complex networks into smaller, manageable objects, making large networks easier to oversee. Furthermore, their modular nature enables the reuse of network components across different scenarios, streamlining the modeling process and ensuring consistency.

For more details on \acrshorts{bn}, we refer to \cite{fenton2018risk}, where additional patterns (idioms) for modeling dependency are discussed. Lastly, we would like to mention that it is often possible to represent parts of the graph by \acrshorts{ft}, which have established closed-form solutions \cite{rausand2021system}.

\subsection{Empirical Modeling of Influence Factors and Collisions} \label{sec:empirical_factor_modeling} \label{sec:injury_risk}
\objective{to estimate for each \acrshort{ifv} and the injury probability node in the \acrshort{dag} a statistical model.}

\begin{rmk}
    In the subsequent quantitative analyses, we will consistently use SI units and will not explicitly list them in the plots. More broadly, the general recommendation is to use SI units throughout the safety analyses as the standard, in order to avoid the error-prone conversions between different unit systems.
\end{rmk}

\subsubsection{Univariate Influence Factors}\label{sec:univariate_influence_factors}
We begin by modeling the \acrshorts{ifv} represented by the independent nodes of the \acrshort{bn}, referred to as leaves of the \acrshort{dag}, focusing initially on univariate distributions. \\
For the influencing factors, we generally have access to a large number of samples. This allows us to estimate distribution parameters using \acrfull{mle} disregarding the (epistemic) uncertainty associated with the parameters (hence \quotes{point estimate}). \\

Since parameter estimation (\quotes{fitting}) is typically handled by standard libraries, we do not delve into its details here but instead refer to foundational literature, such as \cite{james2023introduction}. Even though the process of sampling from these distributions is also managed by these libraries, we still provide a brief example of the Probability Integral Transform, which in general the sampling makes use of. This is because it serves as a foundation for understanding the estimation and sampling of correlated multivariate distributions in Sec.~\ref{sec:multivariate_factors}.\\

\begin{exmp}[Probability Integral Transform]\label{ex:inverse_transform}
\begin{figure}[htb]
\centering
\includegraphics[width=.5\textwidth]{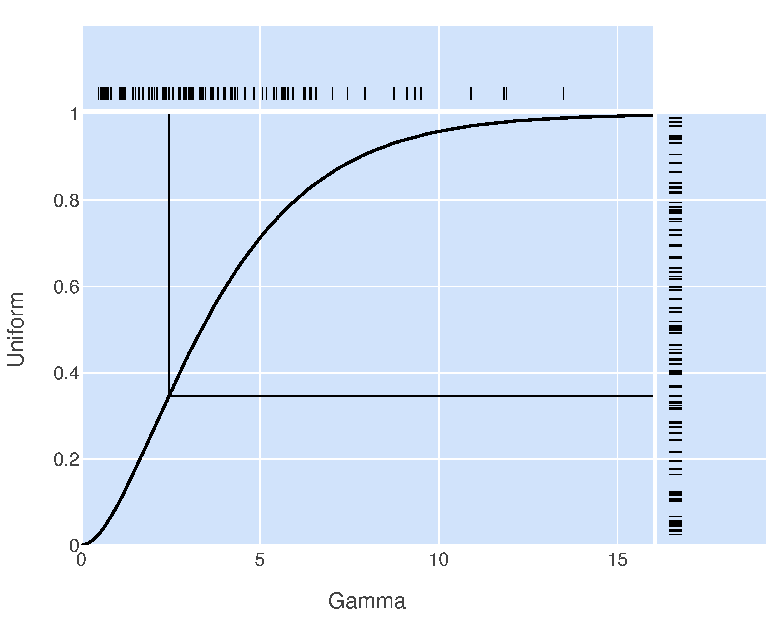}
\caption{Transformation of samples} \label{fig:inverse_transform}
\end{figure}
Let us assume we have collected samples $x_i$ of a certain influence factor shown as short vertical lines on the top of Fig.~\ref{fig:inverse_transform}, and apply MLE to compute the distribution
parameters, in this case of a gamma distribution. We can then transform each sample by the application of the  \acrfull{cdf} $F_X(x)$, see the black curve in the center of the plot. For example, for the original sample $2.451805$ we get the transformed sample 
\begin{equation*}
   0.346714 = F_X(2.451805). 
\end{equation*}
When this is done for all samples $x_i$, given the estimated distribution describes the data perfectly, the transformed samples, plotted on the right, follow a uniform distribution. \\
Inverse Transform Sampling works in the opposite direction by first drawing quasi-random samples from a uniform distribution, which are then transformed into the physical space using the inverse \acrshort{cdf} $F_x^{-1}$.
\end{exmp}

\subsubsection{Multi-Variate Influence Factors}\label{sec:multivariate_factors}
Some sources like vehicle fleets provide multivariate data sets from which the joint distributions of certain \acrshorts{ifv} can be directly estimated. Copulas, which are frequently used in safety analysis \cite{kluppelberg2014risk}, are well-suited for this. They are particularly appropriate for modeling in high dimension due to the small number of parameters and their computationally efficiency \cite{joe2014dependence}. The general approach can be explained using the following example.

\begin{exmp}[Estimation via Copulas]
In the first step, for each variable we separately estimate the marginal distribution. In the second step, as demonstrated in Sec.~\ref{sec:univariate_influence_factors}, all samples are transformed component-wise into the uniform space using the \acrshorts{cdf} of the marginals. Let us assume, that the best fit to a bivariate data set is given by a gamma (the same as in Ex.~\ref{ex:inverse_transform}) and a normal distribution. If the joint distribution is given by the \textit{independent} marginals, see left of Fig.~\ref{fig:independent_copula}, the transformed samples form a square, evenly distributed scatter plot with no trends, see right of Fig.~\ref{fig:independent_copula}. \\
If the two dimensions are strongly \textit{correlated}, as shown on the left of Fig.~\ref{fig:dependent_copula}, the dependency structure on the right  appears significantly different. It becomes evident that extreme values in one dimension tend to coincide more frequently with extreme values in the other, resulting in a spindle-like shape. This commonly observed spindle like distribution, known as a Gaussian copula, is characterized in the two-dimensional case by a single parameter, which can be learned, as the last step, via \acrshort{mle} \cite{kluppelberg2014risk}.

\newcommand{\imgwidth}{0.45\textwidth}\label{}
\begin{figure}[htb]
\centering
\subfigure{\label{fig:physical_space_0}\includegraphics[width=\imgwidth]{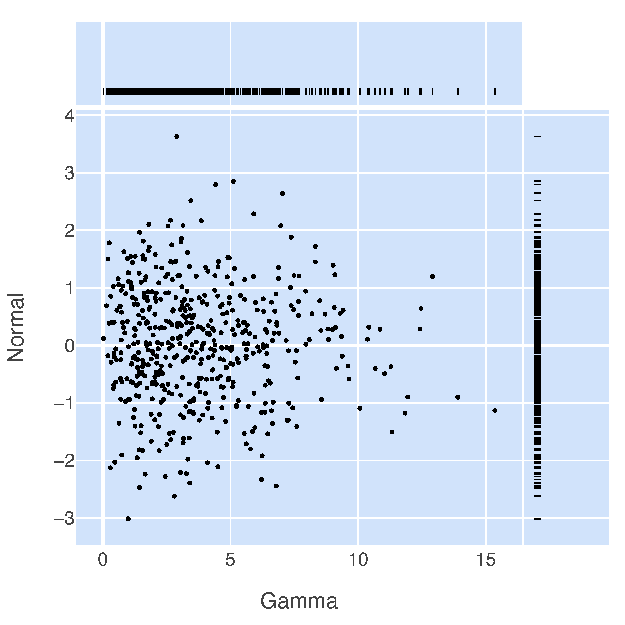}} 
\subfigure{\label{fig:uniform_space_0}\includegraphics[width=\imgwidth]{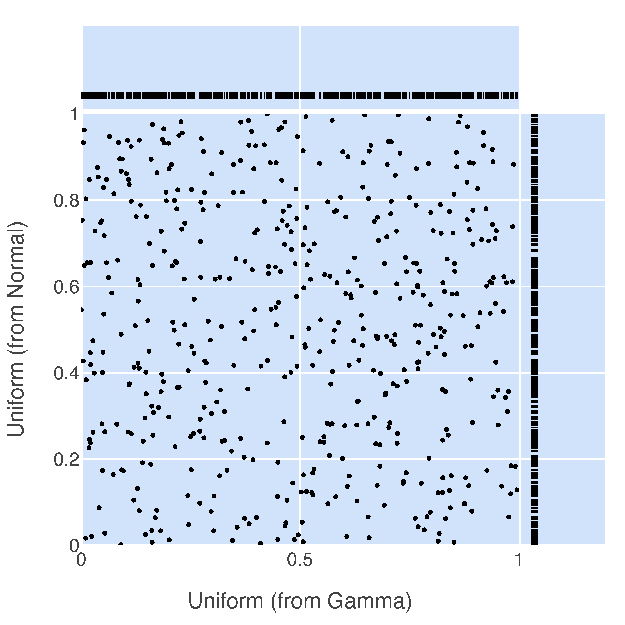}}
\caption{Original (left) and transformed (right) samples from independent distributions}
\label{fig:independent_copula}
\end{figure}

\begin{figure}[htb]
\centering
\subfigure
{\label{fig:physical_space_0.8}\includegraphics[width=\imgwidth]{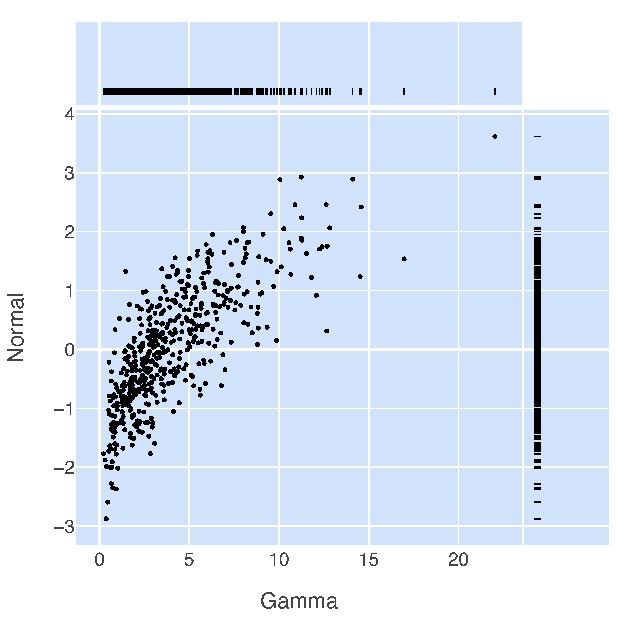}}
\subfigure
{\label{fig:uniform_space_0.8}\includegraphics[width=\imgwidth]{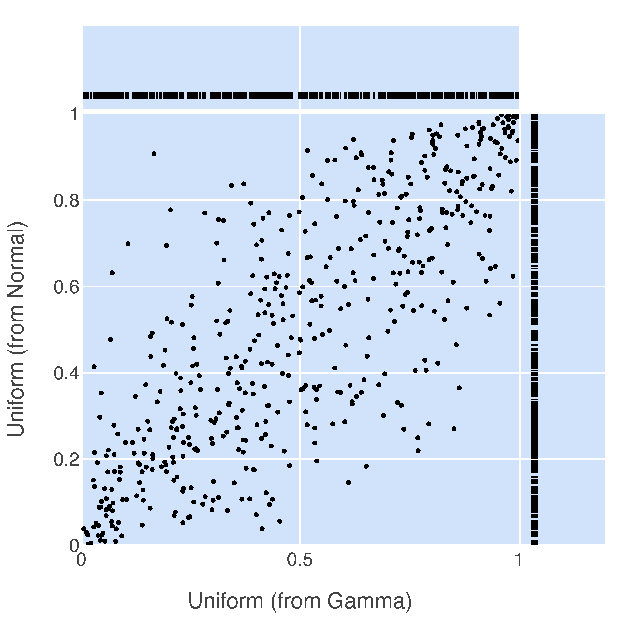}}
\caption{Original (left) and transformed (right) samples from correlated distributions}
\label{fig:dependent_copula}
\end{figure}

\end{exmp}

Generally speaking, a joint distribution can often be well represented by component-wise marginal distributions and a Copula. Drawing a sample from this model involves sampling from the Copula and transforming each dimension via the inverse \acrshort{cdf} of the respective marginal to the physical space.

\begin{exmp}[Joint distribution of \acrshorts{ifv} in \acrshort{hs}\,1 via Copula]\label{ex:joint_ifvs}
\newcommand{\orientation}{{\textit{Intruder Orientation}} }
\newcommand{\offset}{{} }
\newcommand{\velocity}{{\textit{Host Velocity}} }
\newcommand{\intrusion}{{\textit{Intrusion Depth}} }
When considering the fleet data points in Fig.~\ref{fig:joint_intrusion_offset_scatter}, it becomes apparent that there is a positive correlation between \textit{Intruder Offset} and \textit{Intrusion Depth}. This statistical dependency can be modeled, as part of a combined model together with the dimensions \textit{Host Velocity} and \textit{Intruder Orientation}, via a Copula. \\
\begin{figure}[htb]
\centering
\includegraphics[width=.6\textwidth]{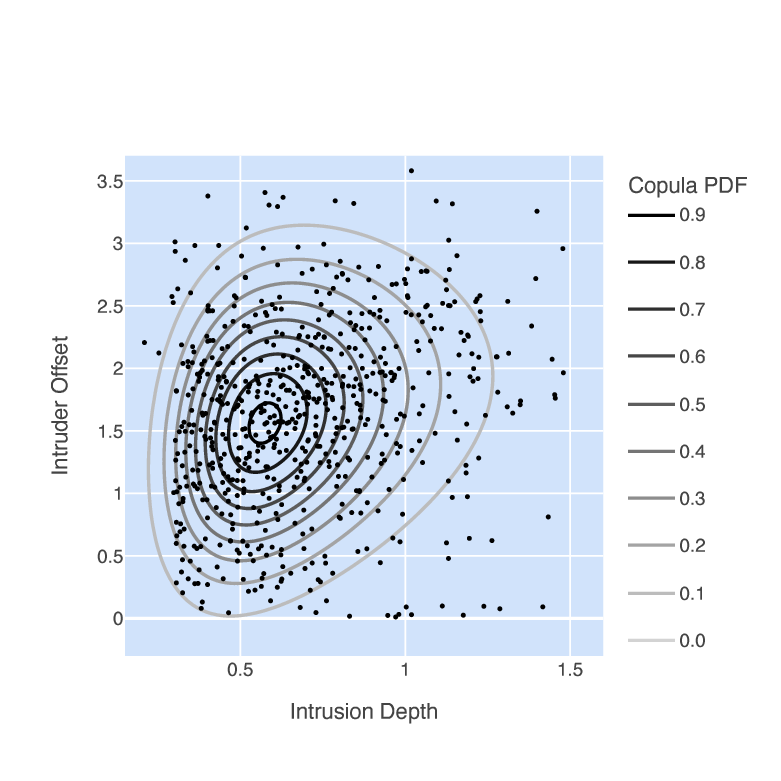}
\caption{\textit{Intrusion Depth} vs.\ \textit{Intruder Offset} data with estimated Gaussian Copula \acrshort{pdf}} \label{fig:joint_intrusion_offset_scatter}
\end{figure}

In a first step, the marginal distribution of each variable is separately estimated. By comparing different distributions, the choice falls on the parametric\footnote{An alternative choice is a kernel density function, which is considered non-parametric and essentially a smoothed version of a histogram \cite{james2023introduction}.} distributions given by Fig.~\ref{fig:marginals} in round brackets. The corresponding \acrshorts{cdf} show good agreement with the empirical \acrshorts{cdf}. \\
In the second step, the data points are transformed, and a Gaussian copula is estimated via maximum likelihood \cite{joe2014dependence}, which provides the correlation matrix\footnote{Note that these are not the correlations of the original data, but those of the transformed data.}
\begin{equation*}
\hat\Sigma = 
\begin{bmatrix}
1.00 & 0.24 & 0.05 & 0.11 \\
0.24 & 1.00 & -0.17 & 0.03 \\
0.05 & -0.17 & 1.00 & -0.01 \\
0.11 & 0.03 & -0.01 & 1.00
\end{bmatrix}.
\end{equation*}
Consequently, the correlation between \textit{Intrusion Depth} and \textit{Intrusion Offset} is 0.24, which is relatively strong in comparison to the correlation between \textit{Intruder Orientation} and \textit{Host Velocity} with is $-0.01$. \\
For visual verification, the resulting joint distribution density can be represented by the contour line as shown in Fig.~\ref{fig:joint_intrusion_offset_scatter}, and compared with the original data. Alternatively, samples can be drawn from the estimated distribution and compared with the original data. \\
Since we need the Copula model conditionally on the \textit{Intruder Type}, the evaluated data points above were previously filtered on \textit{car}. For this reason, the same procedure must also be carried out with \textit{truck}. The Stochastic Simulation in Sec.~\ref{sec:mcs} can then switch between the two models depending on the \textit{Intruder Type}. The probabilities for \textit{Intruder Type} can be directly estimated based on the same fleet data, concretely
\begin{equation*}
    P(\textit{car})= 0.72 \quad \text{and} \quad P(\textit{truck})=0.28.
\end{equation*}

\begin{figure}[htb]
\newcommand{\imgwidth}{0.49\textwidth}
\centering
\subfigure[Host Velocity (Generalize extreme value)]{\label{fig:marginal_v_host}\includegraphics[width=\imgwidth]{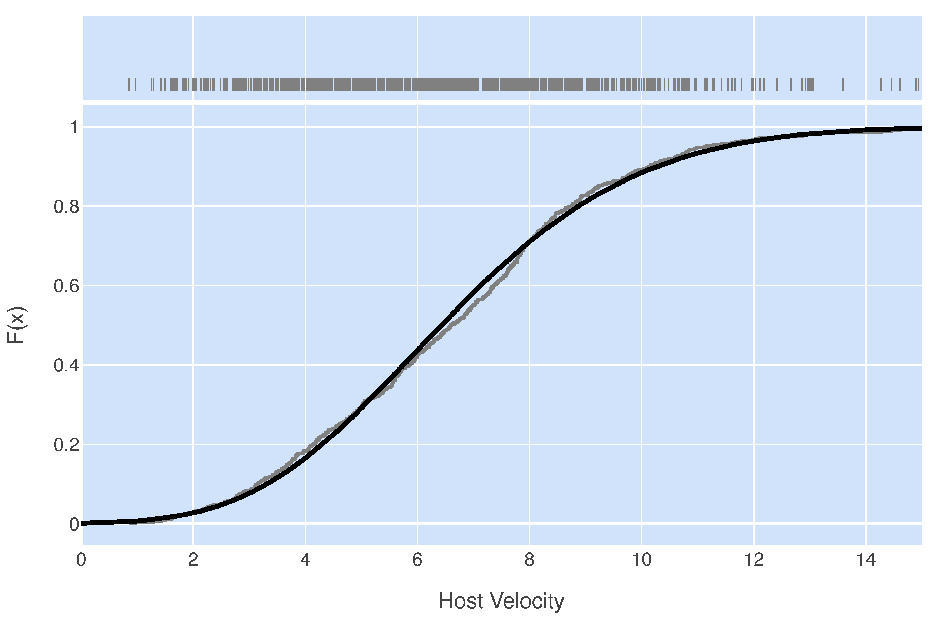}} 
\subfigure[Intruder Orientation (t)]{\label{fig:marginal_yaw_angle}\includegraphics[width=\imgwidth]{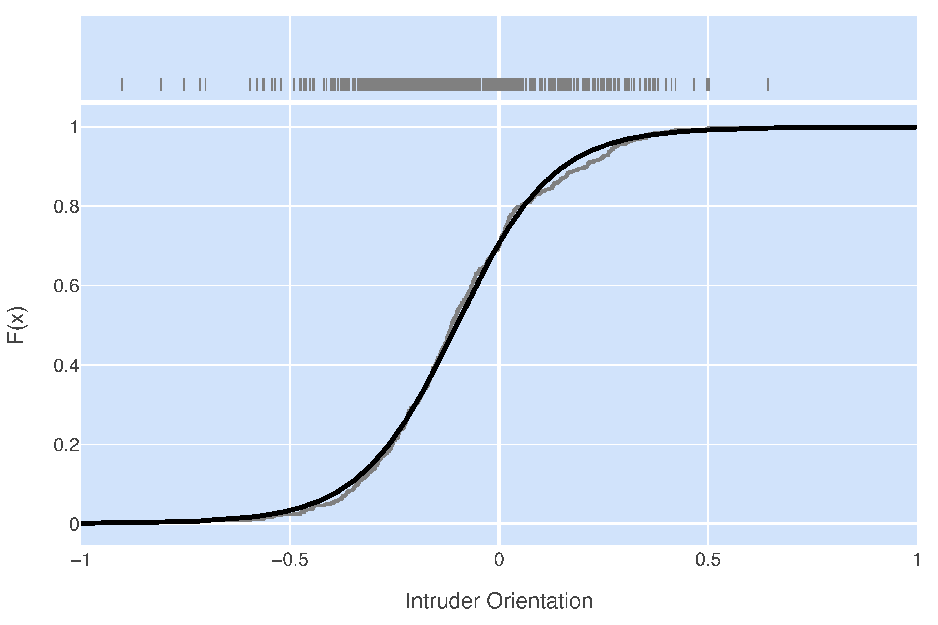}}

\subfigure[Intrusion Depth (Generalize extreme value)]{\label{fig:marginal_intrusion}\includegraphics[width=\imgwidth]{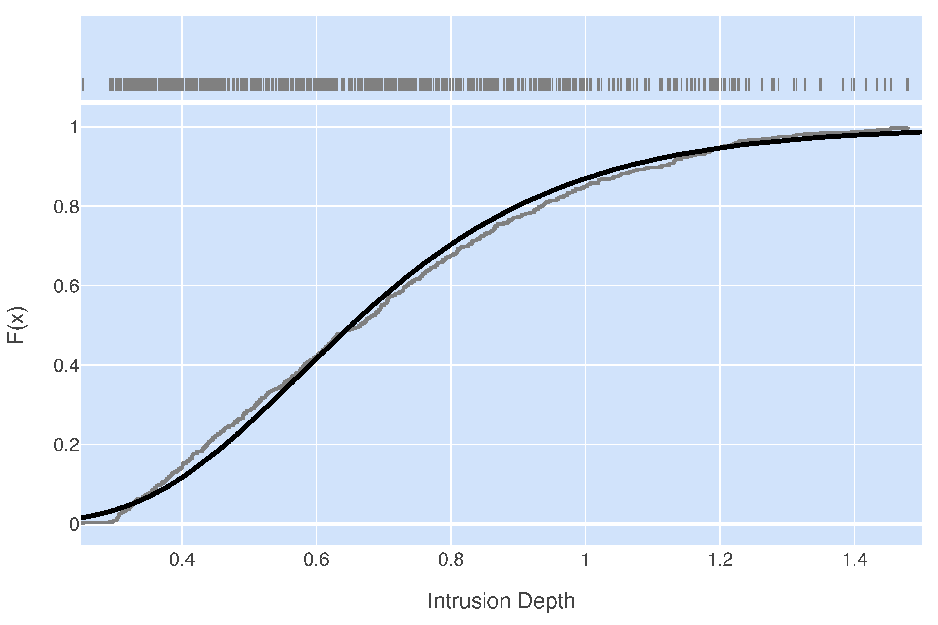}}
\subfigure[Intruder Offset (gamma)]{\label{fig:marginal_offset_lat}\includegraphics[width=\imgwidth]{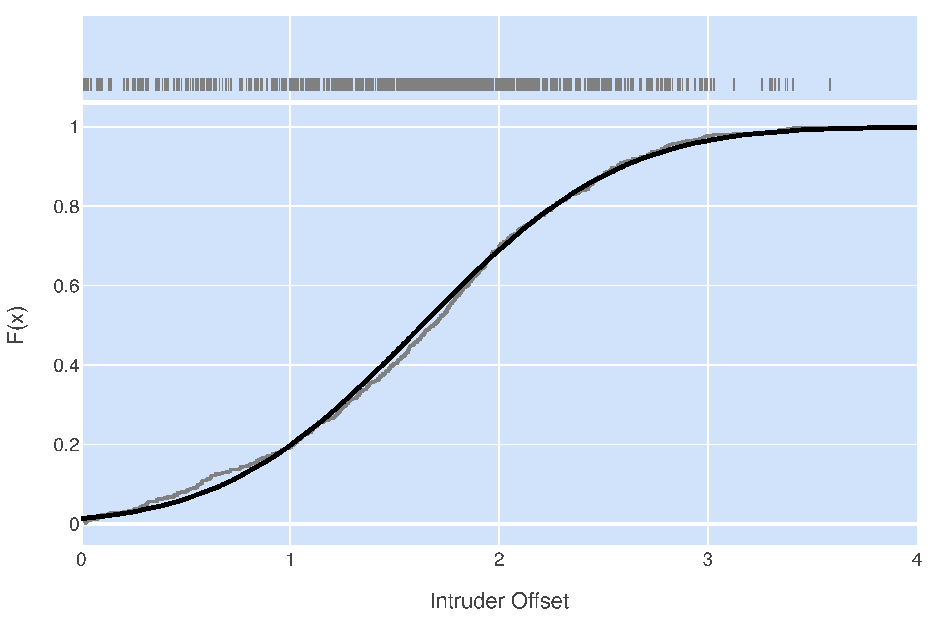}}
\caption{Empirical (grey) and fitted (black) \acrshorts{cdf} of marginals}
\label{fig:marginals}
\end{figure}
    
\end{exmp}


\subsubsection{Injury Risk Modeling}
The modeling of injury risk is generally performed through logistic regression, e.g.\ \cite{nishimoto2017serious}, where the parameters are learned from accident data. Often, the coefficients are directly provided in the literature, so they do not need to be re-estimated. However, the available models are limited to estimate the risk of a single accident participant, so they need to be combined, as the following example shows.

\begin{exmp}[Combined injury risk model]\label{ex:combined_injury}
\newcommand{\intr}{\text{target}}
\newcommand{\host}{\text{host}}
\newcommand{\deltavintr}{\Delta v_\text{target}}
\newcommand{\deltavhost}{\Delta v_\text{host}}

\begin{figure}[htb]
\centering
\includegraphics[width=.7\textwidth]{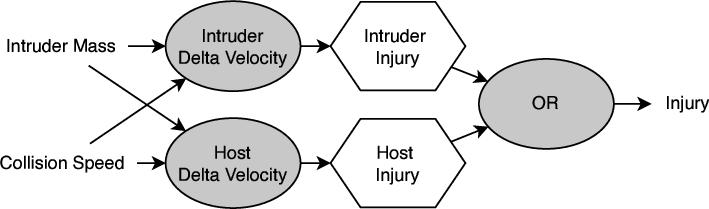}
\caption{Detail view for 
    \textit{Injury} in Fig.~\ref{fig:dag_partially_blocked_lane}} \label{fig:injury_risk_graph}
\end{figure}

To model the overlap collision with the intruder, it is necessary to combine two injury risk functions: $g_{I_x,\host}(\deltavhost)$ for the frontal collision of the host vehicle and $g_{I_x,\intr}(\deltavintr)$ for the rear-end collision of the intruder, as shown by the hexagons in Fig.~\ref{fig:injury_risk_graph}. Both assume that the injury probability is only conditional on the difference in vehicle speeds $\deltavhost$ and $\deltavintr$, respectively, induced by the collision.
The differences in speed, in turn, are influenced by the colliding vehicles' masses. Assuming a simplified model of a perfectly inelastic collision, the conservation of momentum yields
\begin{equation*}
    \Delta v_\host = v_\text{crash} \frac{m_\intr}{m_\host + m_\intr}, \quad
    \Delta v_\intr = v_\text{crash} \frac{m_\host}{m_\host + m_\intr}.
\end{equation*}
The risk acceptance criterion is defined according to the definition in Table~\ref{tab:il} such that events are considered whenever \textit{at least} one injury of a certain severity (or greater) has occurred. Therefore, we must apply the logical OR operation to the two injury probabilities, as illustrated in Fig.~\ref{fig:injury_risk_graph}. Due to the conditional independence of two injury nodes, we get the combined injury risk function 
\begin{align*}
g_{I_x}(\deltavhost, \deltavintr) &= P( I_{x,\host} \cup I_{x,\intr}  | \deltavhost, \deltavintr) \\
&= P( I_{x,\host} | \deltavhost) + P( I_{x,\intr} | \deltavintr) - P( I_{x,\host} | \deltavhost) \cdot P( I_{x,\intr} | \deltavintr) \\
&=g_{I_x,\host}(\deltavhost)+ g_{I_x,\intr}(\deltavintr) - g_{I_x,\host}(\deltavhost) \cdot g_{I_x,\intr}(\deltavintr).
\end{align*}
The individual Injury Risk models $g_{I_x,\host}$ and $ g_{I_x,\intr}$ can be referenced from \cite{nishimoto2017serious}. Alternatively, injury risk models can be derived based on detailed collision dynamics simulations used for the evaluation of passive safety systems. 
\end{exmp}

\subsection{Empirical Modeling of Safety Performance} 
\label{sec:empirical_performance_modeling}
\objective{to model for each \acrshort{spv} in the \acrshort{dag} the marginal and conditional distributions.}

With the current state of the art, one cannot rule out that a perception component might occasionally fail, for example, by misdetecting an object altogether. Furthermore, there is also the possibility that certain effects occur that cause an error to become so large that it is reasonable to conservatively consider the outlier as a failure. Since  failures generally have the most severe consequences, e.g., the vehicle does not respond at all and collides at full speed, the modeling of failures requires special attention. \\
At the same time, large amounts of data are required to estimate failure probabilities and rates for reliable components accurately. Since data is in general a limiting factor, the (epistemic) uncertainty of the parameter estimate needs to be quantified, cf.\ Ex.~\ref{ex:2oo3_aleatory_and_epistemic_uncertainty}. For this, we distinguish again the continuous and the discrete operating mode, this time not at the vehicle level as in Sec.~\ref{sec:risk_assessment}, but at the component level.

\subsubsection{Failures in Discrete Mode}\label{sec:failure_prob_discrete}\label{sec:nononformativeprior}
A Bernoulli process is widely used to model how many times a component fails in relation to how often the component is demanded to perform a task. The model assumes that the failure events occur with a constant probability $p_f$ per demand independently of each other. \\
One way to estimate the failure probability $p_f$ is to collect data of a certain number of (independent) demands (\quotes{trials}), denoted as $n$. The number of observed failures $n_f$ is then divided by this number,
\begin{equation}\label{eq:max_likelihood_mean}
    \hat p_f = \frac{n_f}{n},
\end{equation}
which constitutes the \acrshort{mle} of $p_f$.
This estimate should not be used carelessly in further calculations, as it does not account for epistemic uncertainty. Instead, a common approach in safety applications is to use a Bayesian estimate of the failure probability \cite{rausand2021system, fenton2018risk, betz2022bayesian} given by the formula 
 \begin{equation}\label{eq:bayes_estimate}
    \hat p_f = \frac{n_f + 1} {n + 2}.
\end{equation}
This estimate arises from modeling the failure probability as a random variable with a uniform prior distribution over the range [0, 1]. Given $n_f$ observed failures out of $n$ total trials, the posterior distribution of the failure probability $p_f$ follows a beta distribution with shape parameters $\alpha = n_f + 1$ and $\beta = n - n_f + 1$, cf. Fig.~\ref{fig:posterior}. The mean of this beta posterior distribution gives the Bayesian point estimate \eqref{eq:bayes_estimate}, which accounts for the epistemic uncertainty in the failure probability and provides in general more robust predictions than \eqref{eq:max_likelihood_mean}. Applying \eqref{eq:bayes_estimate} has the effect of progressively shifting the failure probability estimate more towards the observed data, starting from the generally conservative prior with a mean of $1/2$.

\begin{exmp}[Bayesian detection failure probability estimate]\label{ex:bayesian_failure_probability}
Since \acrshort{hs}\,1 encounters are rare in traffic, the scenario is reenacted 1000 times on a test track, each experiment based on a random sample from the \acrshorts{ifv} distribution learned in Ex.~\ref{ex:joint_ifvs} from fleet data. It be observed that the sensors channels fail to detect the intruder 0, 1, and 2 times, respectively. This leads with \eqref{eq:bayes_estimate} to the failure probability estimates
\begin{align*}
 \hat p_{1} &= (0+1)/(1000+2) \approx 1.0 \times 10^{-4}, \\
 \hat p_{2} &= (1+1)/(1000+2)  \approx 2.0 \times 10^{-4},\\
 \hat p_{3} &= (2+1)/(1000+2)  \approx 3.0 \times 10^{-4}.
\end{align*}
\end{exmp}

\subsubsection{Failures in Continuous Mode}
Another widely used model in risk assessment is the homogeneous\footnote{Software generally does not undergo an aging process. However, if the aging of sensors needs to be considered, an inhomogeneous Poisson process can be employed.} Poisson process, which can be used to describe how often a component fails in relation to the duration the component is performing a task. The model assumes that the failure events occur independently of each other at a constant rate $\lambda_f$ over time $T$. \\
Analogous to \eqref{eq:max_likelihood_mean}, \acrshort{mle} provides
\begin{equation}\label{eq:max_likelihood_mean_rate}
    \hat \lambda_f = \frac{n_f}{T},
\end{equation}
which does not account for the epistemic uncertainty. The Bayesian estimate
\begin{equation}\label{eq:bayes_mean_rate}
\hat \lambda_f = \frac{n_f + 1}{T}
\end{equation}
arises from modeling the failure rate $\lambda_f$ as a random variable with an improper uniform prior distribution over the range [0, $\infty$). Given $n_f$ observed failures over a total observation time $T$, the posterior distribution of $\lambda_f$ follows a gamma distribution with shape parameter $\alpha = n_f + 1$ and rate parameter $\beta = 1$. The mean of this gamma posterior distribution gives the Bayesian point estimate \eqref{eq:bayes_mean_rate}. The more failures are detected, the closer the estimate is to the MLE \eqref{eq:max_likelihood_mean_rate}.

\begin{exmp}[Bayesian lane-keeping failure rate estimate]
To assess the lane estimation performance, fail/pass criteria are defined and representative driving tests are conducted in real-world traffic. The evaluation of 2000\,h of data, reveals zero failure events, which leads with \eqref{eq:bayes_mean_rate} to the failure rate estimate
\begin{equation}\label{eq:bayes_mean_rate_example}
\hat \lambda_f = \frac{0 + 1}{2000 \,h} = 5.0 \times 10^{-4} /h.
\end{equation}
\end{exmp}

\begin{figure}[htb]
\centering
\subfigure[Explicit modeling of dependent failures]
{\includegraphics[width=0.35\textwidth]{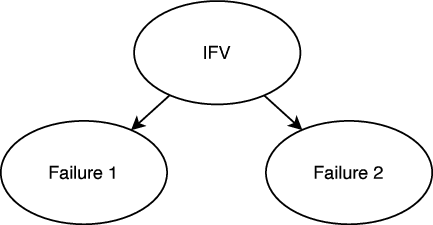}
\label{fig:dependent_failure_explicite}}
\hspace{.5cm}
\subfigure[Joint noise model for regression]{\includegraphics[width=0.53\textwidth]{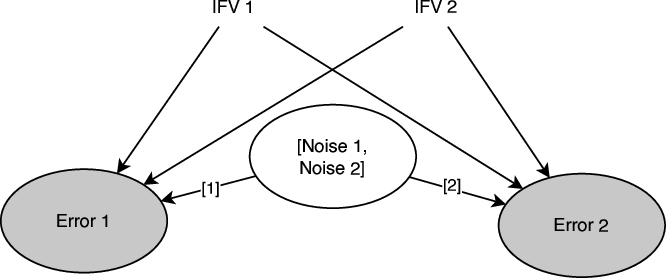}\label{fig:residual_copula}}
\caption{Special cases of dependent errors and failures}
\end{figure}

The formulas \eqref{eq:bayes_estimate} and \eqref{eq:bayes_mean_rate} can be utilized to estimate the marginal probabilities of failures based on data, which is sufficient if statistical independence between the failures can be assumed. This assumption can be supported by technical measures, such as the heterogeneously redundant sensor channels trained on independent data sets, as outlined by Ex.~\ref{ex:safety_concept} and Ex.~\ref{ex:independen_ml_training}. If a (discrete) common causes can be identified in the design, explicit modeling of the dependency of the failures can be employed, as shown in Fig.~\ref{fig:dependent_failure_explicite}. In this case, the dataset must be split according to the common causes and \eqref{eq:bayes_estimate} and \eqref{eq:bayes_mean_rate} can be used again to estimate the failure probabilities for the different common causes. \\
Moreover, the literature presents a variety of alternative dependent failure models (e.g.\ Beta-Factor Model, Binomial Failure Rate Model \cite{rausand2021system}); however, the challenge lies in estimating the dependency parameters from the rare failure events. 

\subsubsection{Errors}
Fortunately, state-of-the-art perception system components do not fail often, but there is in general an error present that is often best described by a continuous distribution. The control errors of an \acrshort{ads} exhibit a continuous behavior too. \\
The errors are often the result of the combination of numerous influences, which motivates the modeling with a normal distribution (central limit theorem) \cite{james2023introduction}. As long as these influences only have an effect on individual components, this noise can be modeled as a marginal distribution in the \acrshorts{bn}. However, if the noise is induced by certain influences that have an impact on multiple uncertainties, this dependency needs to be modeled, as discussed in detail in the previous Sec.~\ref{sec:dependent_performance_anaylsis}, and often requires the estimation of a conditional probability. A suitable methodology for this is regression analysis, which was shortly introduced in the context of Factor Screening in Sec.~\ref{sec:factor_screeining}. We can built upon these results, yet in this section, the noise term plays a more significant role, as it represents an uncertainty that can significantly contribute to the overall risk, especially in combination with other uncertainties. \\
There are a variety of regression methods, a simple and yet powerful one is linear regression. The underlying assumptions are \cite{james2023introduction, montgomery2021introduction}
\begin{enumerate}
    \item a linear relationship between the input and output variables,
    \item an additive, normally distributed noise term (homoscedasticity, normality), and 
    \item independence of the observations,
\end{enumerate}
which is given by the model \eqref{eq:interactions}. Additionally, the data used for learning should not be strongly correlated (multicollinearity), i.e., the data should not be located along a line.
The following example shows how the model structure can be selected, the parameters learned, and the assumptions verified.

\remark{Note that if one of the first two assumptions does not hold true for the original variables, it is often possible to apply transformations to the inputs and outputs, such as logarithmic, polynomial, or other functional transformations, to create a new set of variables for which the assumptions are (approximately) satisfied. By carefully selecting these transformations, we can effectively model complex relationships using linear regression techniques.}

\begin{figure}[htb]
\newcommand{\imgwidth}{0.32\textwidth}
\centering
\subfigure{\includegraphics[width=\imgwidth]{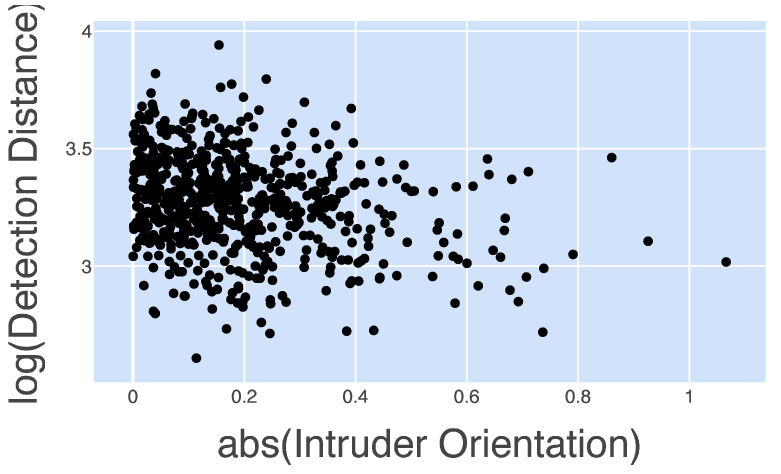}}
\subfigure{\includegraphics[width=\imgwidth]{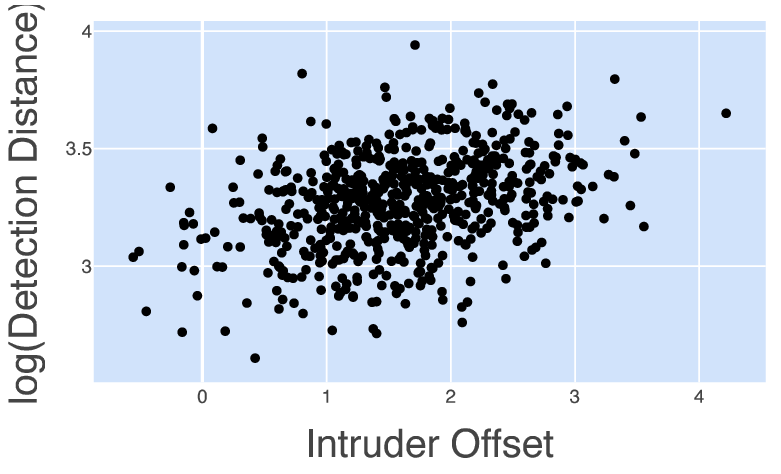}}
\subfigure{\includegraphics[width=\imgwidth]{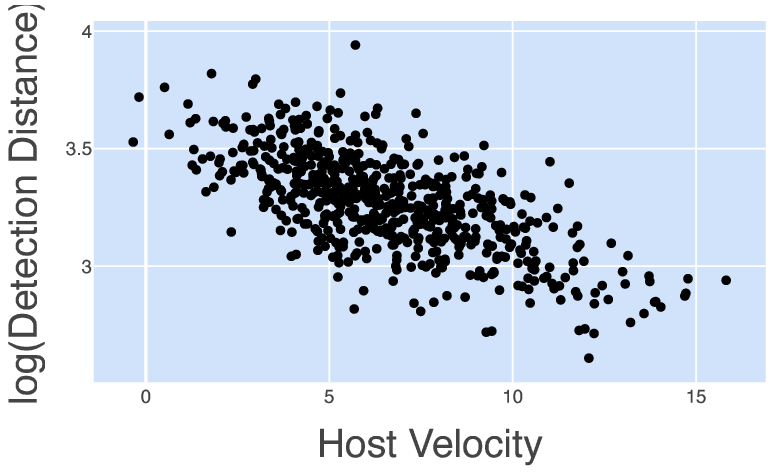}}
\caption{Dependent vs.\ independent variable plots}
\label{fig:dependent_vs_independent}
\end{figure}

\begin{figure}[htb]\label{fig:qq_residuals}
\newcommand{\imgwidth}{0.32\textwidth}
\centering
\subfigure{\includegraphics[width=\imgwidth]{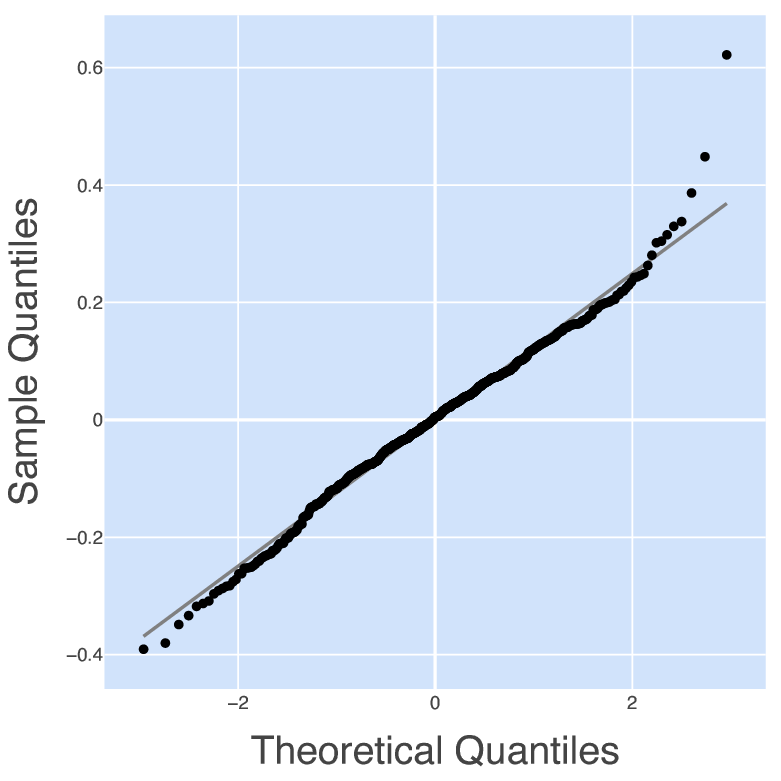}}
\subfigure{\includegraphics[width=\imgwidth]{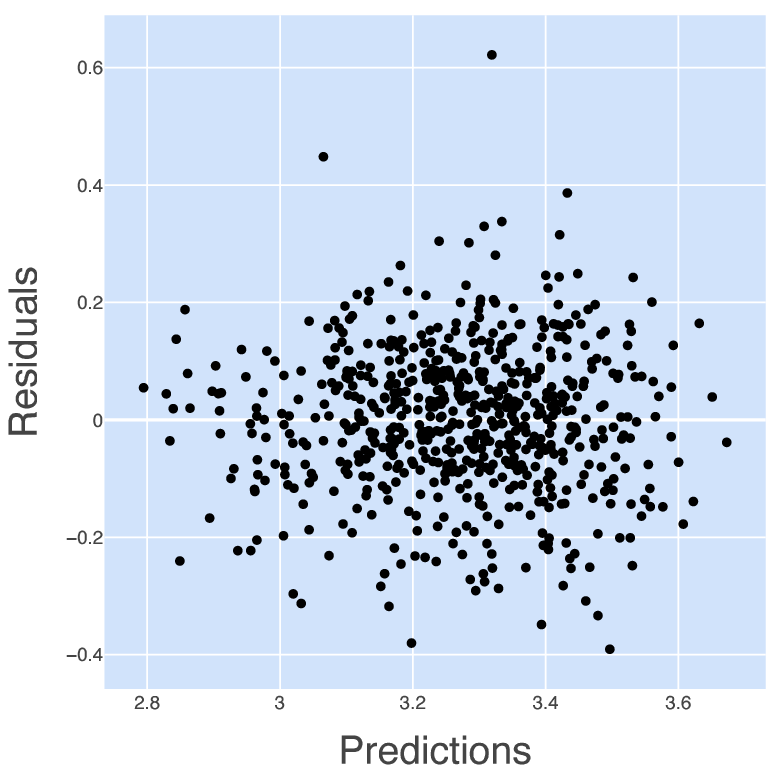}}
\subfigure{\includegraphics[width=\imgwidth]{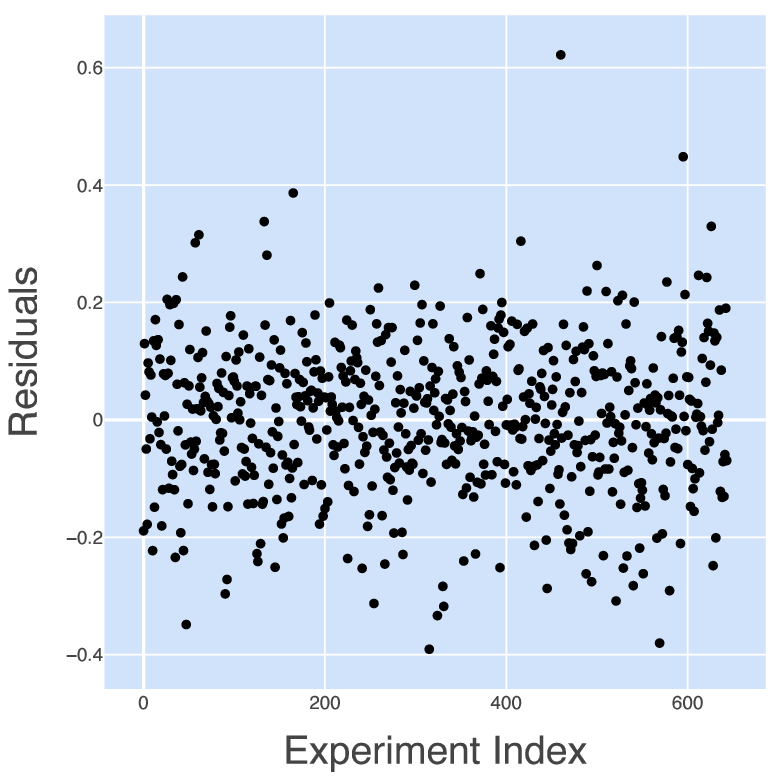}}
\caption{Q-Q plot of residuals (left), residuals against predictions (middle) and experiment index (right)}\label{fig:validation_plots_regression}
\end{figure}

\begin{exmp}[Linear Regression Model for Detection Distance]\label{ex:linear_regression}
Based on expert knowledge and the results from Factor Screening, the \textit{Detection Distance} of a Sensor (different from the one in Ex.~\ref{ex:factorial}) is modeled as
\begin{equation}\label{equ:log_regression_equation}
    \log(\textit{detection distance}) \equiv y = \beta_0 + \beta_1 x_1 + \beta_2 | x_2 | + \beta_3 x_3 + \epsilon, \quad \epsilon \sim \mathcal{N}(0, \sigma^2), 
\end{equation}
with the same independent variables as already defined in Tab.~\ref{tab:factors_and_ranges}. Notice that the log-transformation is often used in linear regression to address issues of nonlinearity, heteroscedasticity, and non-normality. 
The parameters $\hat\beta_i$ as well as $\hat\sigma$ are obtained through \acrshort{mle} based on the non-failing data samples from Ex.~\ref{ex:bayesian_failure_probability}.
Furthermore, the following checks are performed to verify the assumptions of the linear regression model \cite{james2023introduction, montgomery2021introduction}:

\begin{enumerate}
    \item Linearity is checked by plotting the log-transformed detection distance against each independent variable, see Fig.~\ref{fig:dependent_vs_independent}.
    \item The noise is checked for normality by a Quantile-Quantile (Q-Q) plot of the residuals (the error estimates) shown in Fig.~\ref{fig:validation_plots_regression} (left). The straight line indicates that the residuals are mostly normally distributed. Furthermore, the plot of the residuals against the fitted values in Fig.~\ref{fig:validation_plots_regression} (middle) shows a random scatter, indicating that the variance of the residuals is constant.
    \item Lastly, the independence of the observations is checked by plotting the residuals over the experiments in execution order, see Fig.~\ref{fig:validation_plots_regression} (right).
\end{enumerate}
As the data was generated using random samples from the weakly correlated influence factors from Ex.~\ref{ex:joint_ifvs}, multicollinearity can be excluded.
\end{exmp}

The coefficient of determination \cite{james2023introduction, montgomery2013design}, $R^2\!\approx\!0.48$, indicates that a substantial portion of the variance in the logarithmized \emph{Detection Distance} can be explained by the modeled influence factors. However, a non-negligible part of the variance remains as independent noise in the model and needs to be considered in the Stochastic Simulation in Sec.~\ref{sec:mcs}. For this, samples from the model are drawn by sampling the independent, normally distributed noise with $\mu = 0$ and $\hat \sigma^2$ and then combined with the systematic influences according to \eqref{equ:log_regression_equation} using $\hat\beta_i$. Taking the exponential of the log-distances provides the detection distance.

\remark{
If multiple \acrshorts{spv} are influenced by the same data sources, it can be observed that the residuals of the regression models are statistically dependent. For example, the residuals of the two regression models for position and velocity errors in the object tracking of the camera system are correlated. This can be explained by the fact that the camera determines the object velocity from the object position over time and that the \acrshorts{ifv} do not fully explain the variability in the \acrshort{spv}. In such cases, one can model the residuals using a copula \cite{berk2019safety}, see Fig.~\ref{fig:residual_copula}, based on the methods introduced in Sec.~\ref{sec:multivariate_factors}.
}

\begin{rmk}\label{rmk:overfitting}
It is essential that the validation is based on data that is separate and independent from the data used in the development of the \acrshort{ads}. This is a common requirement in safety validation, in the same way that the hold-out data set in \acrshort{ml} must not be used for training and hyperparameters optimization to prevent overfitting. To ensure that the development is not influenced, even indirectly, by the hold-out data, it is recommended to restrict the developer team's access to this data.
\end{rmk}

In practice, models encountered are often more complex than the one in the Ex.~\ref{ex:linear_regression} so that model selection strategies such as forward/backward-selection \cite{montgomery2021introduction, james2023introduction}, as well as methods based on cross-validation \cite{james2023introduction} are indispensable for this purpose. Furthermore, there is a variety of methods that can significantly extend the applicability of linear regression through adept transformation of input and output data (e.g., the logarithmic transformation in Ex.~\ref{ex:linear_regression}). Finally, if no transformation can effectively address non-normal residuals, an alternative is to model the noise with a different distribution, like the generalized normal distribution.

\subsection{Stochastic Simulation} \label{sec:mcs}
\objective{to estimate the risk modeled by the \acrshort{bn}.}
The chapter summarizes two important aspects for simulating the risk of a hazard scenario:
In Sec~\ref{sec:mil_sil}, the distinction between model-based and software-in-the-loop approaches is explained, both modeling the deterministic behavior of the system within the Bayesian network. This completes the modeling of the \acrshorts{hs}. Section~\ref{sec:mcs_sampling} then illustrates how the overall risk of \acrshorts{hs} can be estimated based on sampling.

\subsubsection{Model and Software in the Loop}\label{sec:mil_sil}
With the completion of the previous step, we have fully established the uncertainty nodes in Fig.~\ref{fig:probabilistic_modelling}, their dependencies, and the respective marginal and conditional probabilities. What remains is to define the gray nodes, namely the deterministic mappings to the \textit{System Reaction Variables} and the \textit{Collision Parameters}.
The \textit{System Reaction Variables} are determined by the Safety Concept. However, in early development stages, the \acrshort{sw} implementation for this is typically not yet available. Despite this, it is imperative to assess the viability of the concept through simulation early on. Therefore, it is recommended to develop a model that encapsulates the crucial system behaviors, an approach known as \acrfull{mil}. Creating this functional model also helps verify the completeness of the requirements set forth in the Safety Concept.
To anticipate the \textit{Collision Parameters}, we need a physical model for the host vehicle. Interactions with non-stationary objects additionally necessitate a kinematic model. Ideally, these models should be integrated with the functional system reaction model to formulate a comprehensive yet computationally efficient model that does not require advancing the simulation at a time rate (i.e., numerically solving an initial value problem). This can be demonstrated with the following example:

\begin{exmp}[Functional-physical model \acrshort{hs} Partially Blocked Lane] \label{ex:functional_pyhsical}
We now derive the deterministic mapping of the gray \textit{Collision Speed} node in Fig.~\ref{fig:dag_partially_blocked_lane} as detailed in Fig.~\ref{fig:dag_collision_speed}.\\

\begin{figure}[htb]
    \centering
    \includegraphics[width=1.0\textwidth]{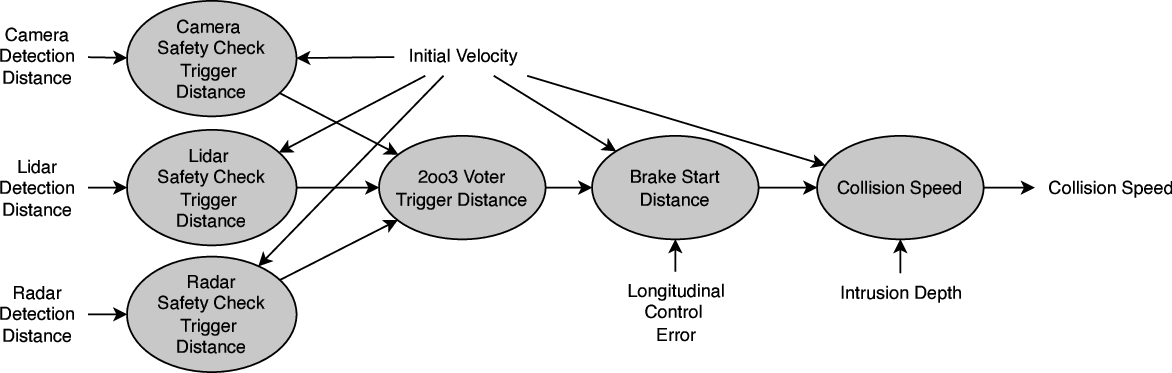}
    \caption{Detail modeling for 
    \textit{Collision Speed} in Fig.~\ref{fig:dag_partially_blocked_lane}} \label{fig:dag_collision_speed}
\end{figure}
With the assumption that the \acrshort{ads} drives with constant velocity until a uniform deceleration is initiated to standstill (or collision), we get the brake profile shown in Fig.~\ref{fig:impact_velocity}.

\begin{figure}[htb]
    \centering
    \includegraphics[width=.9\textwidth]{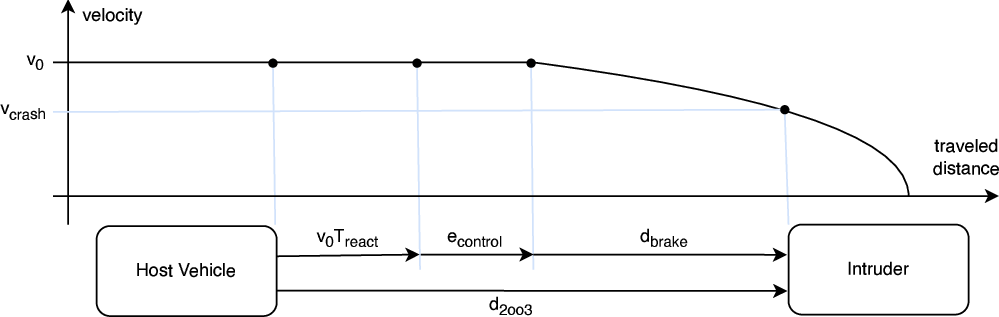}
    \caption{Brake profile and impact velocity} \label{fig:impact_velocity}
\end{figure}
Based on that we first compute the \textit{Safety Check Trigger Distance} for the three sensor channels. The safe distance, cf.\ Ex.~\ref{ex:safety_concept}, simplifies\footnote{Notice that \cite{althoff2016can} includes the control error in the system reaction delay and adds an additional term to account for the host's worst-case acceleration.} for a stationary target and a constant velocity $v_0$ (assumed to be known) to
\begin{align*}
    d_\text{safe} =  \frac{1}{2a}v_0^2 + v_0 T_\text{react} + \bar{e}_\text{control},
\end{align*}
where $a = 7.0\, m/s^2$ represents the specified constant brake deceleration,
$T_\text{react}\geq 0$ is the known reaction delay common to all channels, and $\bar{e}_\text{control}\geq 0$ is a safety margin defined in the Safety Concept to compensate the unknown control error. \\
When the safety distance is breached by the estimate target in a sensor channel $S_i$, the respective safety check triggers, which leads to
\begin{align*}
    d_{\text{trigger}, S_i} =  \min(d_\text{safe},  d_{\text{detection}, S_i}), \quad  i\in\{1, 2, 3\}.
\end{align*}
 \\
 The \textit{\acrshort{too3} Voter
Trigger Distance} is given by the middle of the three trigger distances, which can be expressed by the median, namely
\begin{align*}
    d_\text{2oo3} = \text{median}(d_{\text{trigger}, S_1}, d_{\text{trigger}, S_2}, d_{\text{trigger}, S_3}).
\end{align*}
Furthermore,  as depicted in Fig.~\ref{fig:impact_velocity}, the \textit{Brake Start Distance} is conservatively approximated by
\begin{align*}
    d_\text{brake} = d_\text{2oo3} - v_0 T_\text{react} - e_\text{control},
\end{align*}
taking into account the \textit{actual} control error $ e_\text{control}\geq 0$ as defined in Ex.~\ref{ex:spv}. \\
Based on the kinematics of uniformly accelerated motion, the \textit{Collision Speed} of the \acrshort{ads} with the object can be calculated \cite{althoff2016can} by
\begin{align} \label{eq:v_collision}
    v_\text{crash} = \sqrt{v_0^2 - 2ad_\text{brake}}.
\end{align}
The formula is only valid to a limited extent, so we have to consider two special cases:
A brake start distance $d_\text{brake} \leq 0 $ results in $v_\text{crash} = v_0$ as the \acrshort{ads} collides before decelerating. And a radicand in \eqref{eq:v_collision} smaller than zero leads to a safe standstill as the \acrshort{ads} and the target never intersect during the braking maneuver. \\
Finally, in this example, we make the simplifying assumption that the host vehicle remains centered in the lane and only collides with intruders that have an intrusion depth of at least 1.0\,m.
\end{exmp}

The previous step completes the \acrshorts{bn}, and we can proceed to assess the risk of the scenario.
\newcommand{\xc}{{\mathbf{x}_\text{crash}}}
\newcommand*\diff{\mathop{}\!\mathrm{d}}
\subsubsection{Sampling-based Risk Estimation}\label{sec:mcs_sampling}
\newcommand{\Xc}{{\mathbf{X}_\text{crash}}}
Ultimately, we are only interested in the probability of injury $P(I_x)$ of the three levels $I_x \in \{I_1, I_2, I_3\}$ given the \acrshort{hs}, that is, the hexagonal \textit{Injury} node in Fig.~\ref{fig:dag_partially_blocked_lane}. Because the injury risk model is a conditional probability dependent on collision parameters $\Xc$, namely $g_{I_x}(\xc) := P(I_x | \xc)$, 
we need to marginalize these parameters out, specifically:
\begin{align}\label{eq:riskintegral}
    P(I_x) &= \int_{\Xc} P(I_x | \xc) f_{\Xc}(\xc) \, \diff{\xc} =  \int_{\Xc} g_{I_x}(\xc) f_\Xc(\xc) \, \diff{\xc}.
\end{align}
Since we cannot directly compute $f_{\Xc}(\xc)$ we have to exploit the joint distribution defined by the \acrshort{bn} and marginalize again. This results in a multidimensional integral over all random variables. Even if we further exploit the structure of the \acrshorts{dag}, this becomes computationally demanding for networks with numerous variables. \\
Therefore, we resort to a \acrfull{mcs} for solving \eqref{eq:riskintegral}. \acrshort{mcs} a simple, powerful and especially intuitive approach. For that we generate $n$ artificial samples $\xc_{,i}$ with $i \in \{1 \ldots n\}$ 
from the joint distribution of the \acrshort{bn} and evaluate $g_{I_x}(\xc_{,i})$ for each sample. The integral in \eqref{eq:riskintegral} is then approximated by a sum, namely
\begin{align}\label{eq:mcs_risk}
    P(I_x) &\approx  \hat p_{I_x} :=  \frac{1}{n}\sum_{i=1}^n g_{I_x}(\xc_{,i}).
\end{align}

Due to the properties of a \acrshort{dag}, sampling from a \acrshort{bn} is simple. We first sample the marginal distributions, that is the nodes that do not have any dependencies, and then sequentially sample from the conditional distributions of descendant nodes given the values of their parents. This so-called Ancestral Sampling can be best demonstrated by a simple example:
\begin{exmp}[Ancestral Sampling for Ex.~\ref{ex:simple_joint_dist}]\label{ex:ancestral}
The joint distribution of the \acrshort{bn} shown in Fig.~\ref{fig:example_joint_dist} can be written as the product of marginal and conditional distributions, concretely
\begin{align*}
    f(x_1, x_2, x_3, x_4) = f(x_1) \, f(x_2)  \,  f(x_3|x_1, x_2)  \,  f(x_4|x_2, x_3).
\end{align*}
We first draw a random sample from each marginal distribution, for example $x_1 = 1.3$ and $x_2=-2.7$, and then sample from $f(x_3|x_1 = 1.3, x_2 = -2.7)$ resulting in, e.g., $x_3=9.2$. Lastly, we sample $f(x_4|x_2=-2.7, x_3=9.2)$ and get $x_4=1.5$, for example. This generates a single sample from the joint distribution defined by the \acrshort{bn} and needs to be repeated $n$ times for \acrshort{mcs}.\footnote{It can be faster to generate all $n$ samples of a random variable before moving on to the next variable.}
\end{exmp}
The treatment of deterministic nodes in the \acrshort{bn} differs in that no conditional probability needs to be sampled. Instead, the deterministic function is simply evaluated for the given input variables. Once all samples are available for the collision parameters, we can finally evaluate \eqref{eq:mcs_risk}. \\

On the one hand, the use of random sampling introduces additional sampling uncertainty. On the other hand, this approach now allows us to apply a wide array of statistical methods to evaluate the samples.\\
For instance, let us assume that we collect a sufficient number of collision samples during the simulation, even in low-probability-high-consequence areas such as system failure. 
Then we can conclude that $\hat{p}$, as the sum of many independent and identically distributed samples, see \eqref{eq:mcs_risk}, is normally distributed, according to the Central Limit Theorem. Based on the sample variance 
\begin{equation}\label{equ:standard_deviation}
s^2 = \frac{1}{n-1}\sum_{i=1}^{n} (g_{I_x}(\xc_{,i}) - \hat{p})^2 ,
\end{equation}
we can calculate the 95\% \acrfull{ci} to be
\begin{equation} \label{equ:ci}
  CI = \hat{p} \pm 1.96 \cdot \frac{s}{\sqrt{n}}.
\end{equation}
Due to the inherent variability of the method, this interval is specific to each \acrshort{mcs} run and contains the actual risk defined by the \acrshort{bn} in 95\% of the runs.

\begin{exmp}[Risk quantification in \acrshort{hs}\,1]\label{ex:mcs}
Applying Ancestral Sampling with $n=10^5$ samples to the \acrshort{bn} in Fig.~\ref{fig:dag_partially_blocked_lane} with the sub-networks of Fig.~\ref{fig:injury_risk_graph} and Fig.~\ref{fig:dag_collision_speed} and evaluating \eqref{eq:mcs_risk}, \eqref{equ:standard_deviation}, and \eqref{equ:ci} returns for I2+
\begin{equation*}
\hat p_{I_{2+}} = 5.60\times 10^{-5} \pm 1.08 \times 10^{-5}.
\end{equation*}
Since $p_{I_{2+}}$ gives us the probability of $I_{2+}$ given the scenario, we have to multiply the simulation result with the occurrence rate $\lambda_s = 2.0 \times 10^{-2} / h$ of the scenario, cf.\ Ex.~\ref{ex:hira_partially_blocked_lane}, to get the occurrence rate of $I_{2+}$, namely
\begin{equation*}
\hat \lambda_{I_{2+}} = \lambda_s \cdot \hat p_{I_{2+}} = 1.12 \times 10^{-6} \pm 2.16 \times 10^{-7} / h.
\end{equation*}
This result exceeds the total budget by the order of one magnitude, cf.\ Ex.~\ref{ex:rac},
making it imperative that further risk reduction measures are implemented, see Ex.~\ref{ex:iterative_system_modification} below. 
\end{exmp}

With an efficient implementation of the deterministic functional-physical nodes, processing millions of samples generally poses no issue. \\
Once the Safety Concept has been implemented, we can select a subset of the samples to compare the \acrshort{mil} with the actual vehicle behavior, a procedure known as back-to-back testing. 
Alternatively, we can perform a \acrfull{sil} simulation integrating the \acrshort{sw} implementation and a physical vehicle model within a virtual environment. For this the simulation's ground truth must be corrupted by the errors and failures obtained as samples from the \acrshorts{spv} of the \acrshort{bn} according to their definitions, similar to the traditional method of error injection. \\
In general, however, \acrshort{sil} demands significantly more computational resources than \acrshort{mil}, and even with high-performance computing clusters, processing millions of samples remains time-consuming.
When combined with \acrshort{sil}, it is therefore imperative to use more sophisticated sampling techniques suitable for rare-event simulation \cite{rubinstein2016simulation, bucklew2004introduction}  such as importance sampling or splitting methods. \\

In the case of discrete mode, as described in Sec.~\ref{sec:risk_assessment}, the occurrence of a \acrshort{hs} is considered as given, and the system's reaction to this scenario is simulated over a few seconds, cf.\ Ex.~\ref{ex:mcs}. For the continuous mode, this would usually translate into a continuous simulation over a computationally prohibitive long time interval. Discrete-event simulations \cite{rubinstein2016simulation} offer a solution for this problem, as they make large temporal leaps between certain discrete events, such as components failures, making them significantly more efficient.

\subsection{Sensitivity Analysis and Iterations}
\label{sec:sensitivity_analysis}
\label{sec:iterations}
\begin{obj} 
The main objectives of this section are 
\begin{itemize}
    \item to assign the risk to its sources and
    \item to identify ways to counteract them.
\end{itemize}
\end{obj}

The previously described steps 1-4 in Fig.~\ref{fig:validation_overview} are aimed at systematically deriving a combined statistical model of the system and the environment per \acrshort{hs}, which then culminate in the risk quantification in step 5. As experience has shown, cf.\ Ex.~\ref{ex:mcs}, the result will typically not meet the target budget designated in the \acrshort{hira} of Sec.~\ref{sec:hira} in the first attempt. Generally, this budget can only be shifted between scenarios within limits, requiring other measures to progressively reduce the estimated risk in a time and cost-efficient way.

\begin{figure} [h]
    \centering
    \small
    \includegraphics[width=0.9\textwidth]{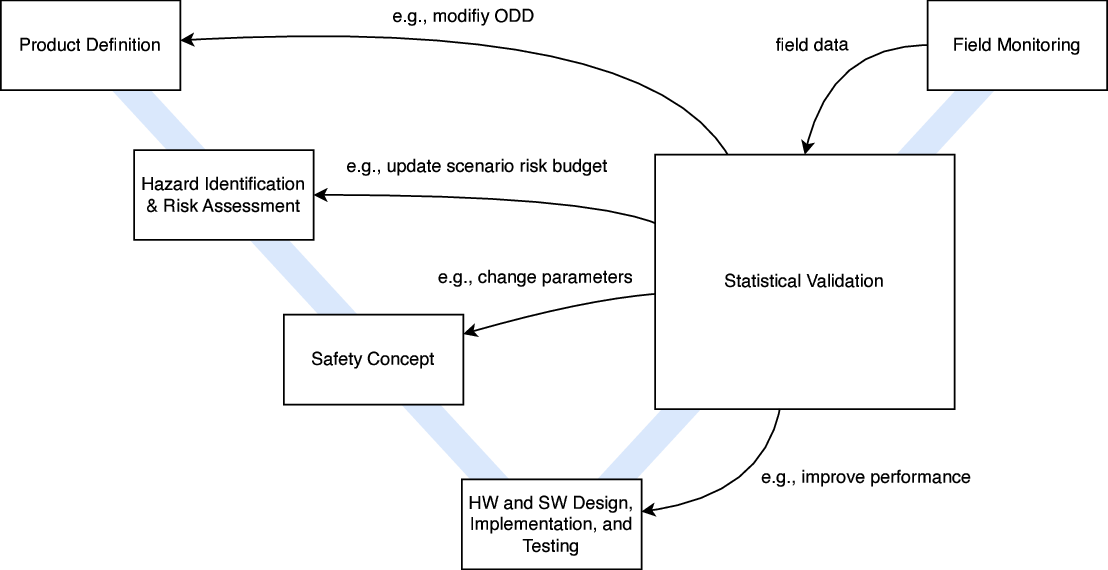}
    \caption{Iterations as part of the V-model}
    \label{fig:iteration}
\label{fig:iterations}
\end{figure}

\begin{figure} [htbp]
    \centering
    \includegraphics[width=0.9\textwidth]{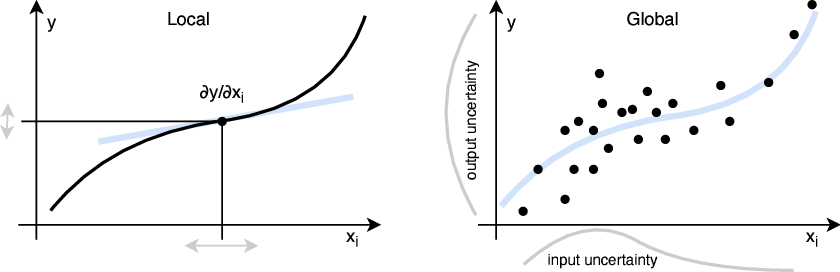}
    \caption{Two common approaches to \acrfull{sa}}
    \label{fig:local_and_global_sa}
\end{figure}

We distinguish between two types of measures that influence different parts of the development cycle. First, measures can be implemented that affect the system's design, which typically initiates a new development cycle, as illustrated in Fig.~\ref{fig:iterations}.
Second, conservative assumptions within the model can be moderated by incorporating additional data, leading to a less conservative overall result. This adjustment will introduce an iteration of the statistical validation steps shown in Fig.~\ref{fig:validation_overview} and discussed in this chapter. \\
However, identifying which measures will significantly reduce risk with manageable effort is a challenging task.
Valuable support for this process is provided by \acrfull{sa} \cite{saltelli2008global}, a crucial component of any risk assessment. \acrshort{sa} helps discern how variations in input variables affect the model's output. Two types of SA are distinguished, each serving different purposes in the design and validation of an \acrshort{ads}. These are local and global \acrshort{sa} as illustrated in Fig.~\ref{fig:local_and_global_sa}.

\subsubsection{Local Sensitivity Analysis}
A common approach in the engineering field is local \acrshort{sa}. This technique focuses on analyzing the gradient of the model output at a specific value of an input (possibly a vector), which is depicted on the left side of Fig.~\ref{fig:local_and_global_sa}. The gradient can be determined by evaluating the model few times with different inputs so that the derivative can be approximated through the difference quotient. Renowned for its efficiency, local sensitivity analysis is especially suited for optimizing parameters that are devoid of uncertainty. \\
In the risk optimization of the \acrshort{ads}, the model output usually represents the estimated risk given by the mean in \eqref{eq:mcs_risk}. The model input often include parameters related to safety mechanisms, like thresholds, see Ex.~\ref{ex:iterative_system_modification}. \\
To enhance the analysis for uncertain inputs, this method can be adapted by multiplying the gradient with the variance of input, leading to sigma normalized derivatives. This adaptation assumes the model behaves approximately linearly, a property that is not given for most \acrshorts{hs}.
\subsubsection{Global Sensitivity Analysis}
\begin{figure} [htbp]
    \centering
    \includegraphics[width=1.0\textwidth]{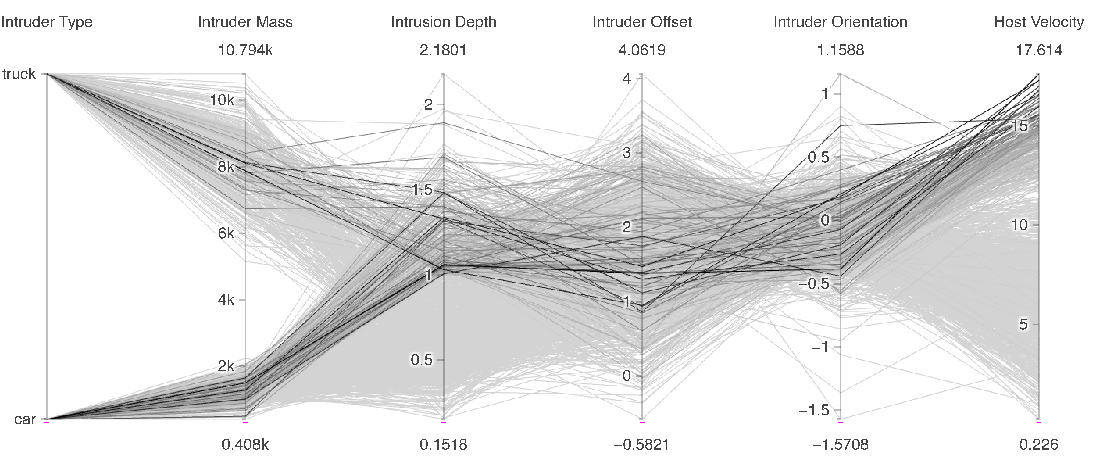} \\
    \includegraphics[width=1.0\textwidth]{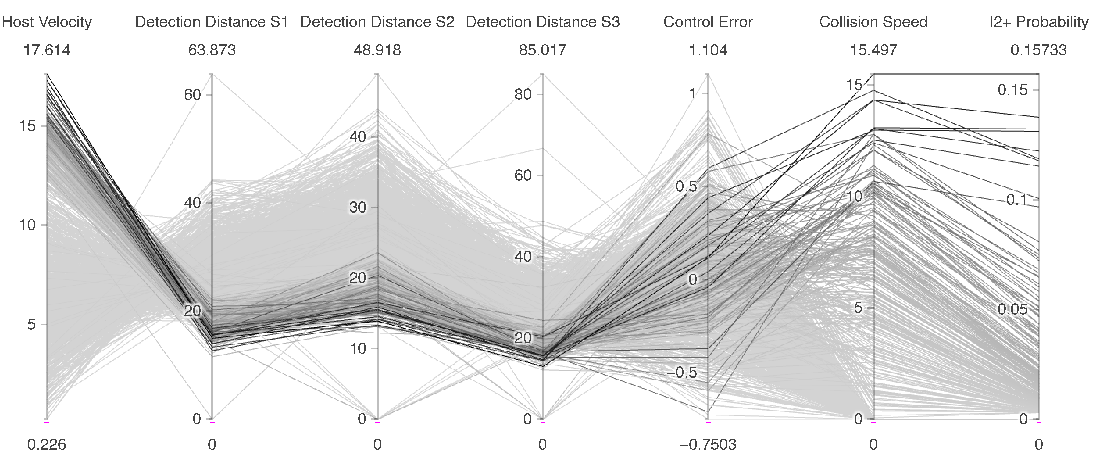}
    \caption{Parallel coordinates plot of \acrshort{mcs} samples for \acrshort{hs}\,1. Darker lines indicate higher injury probabilities.}
    \label{fig:parallel_coordinates}
\end{figure}

Global \acrshort{sa} is a powerful approach that provides a broader perspective by considering the entire distribution of uncertain input variables, see right side of Fig.~\ref{fig:local_and_global_sa}. This can be done visually with scatter plots, i.e.\ displaying the output values against the input values for all \acrshort{mcs} samples from Sec.~\ref{sec:mcs}. It can also be applied to visualize first-order interactions between input variables by using scatter plots that incorporate color coding of the risk. If there are more than two interacting variables, we can resort to so-called parallel coordinates plots as shown in Fig.~\ref{fig:parallel_coordinates}. \\

However, conducting visual \acrshorts{sa} for high-dimensional \acrshort{hs} can be time-consuming. We can overcome this limitation with \acrshorts{sa} algorithms that analyze how much of the variance in the output can be attributed to the different input variables, leading to a so-called variance-based \acrshort{sa} \cite{saltelli2008global}. These algorithms provide sensitivity coefficients that can also consider the interactions between input variables.

\begin{figure}[htb]
\centering
\subfigure{\includegraphics[width=80mm]{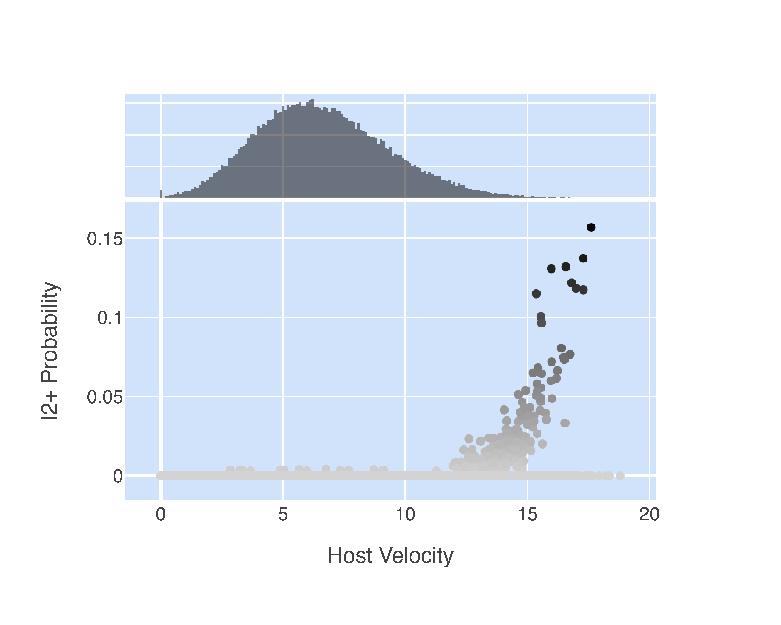}} 
\subfigure{\includegraphics[width=80mm]{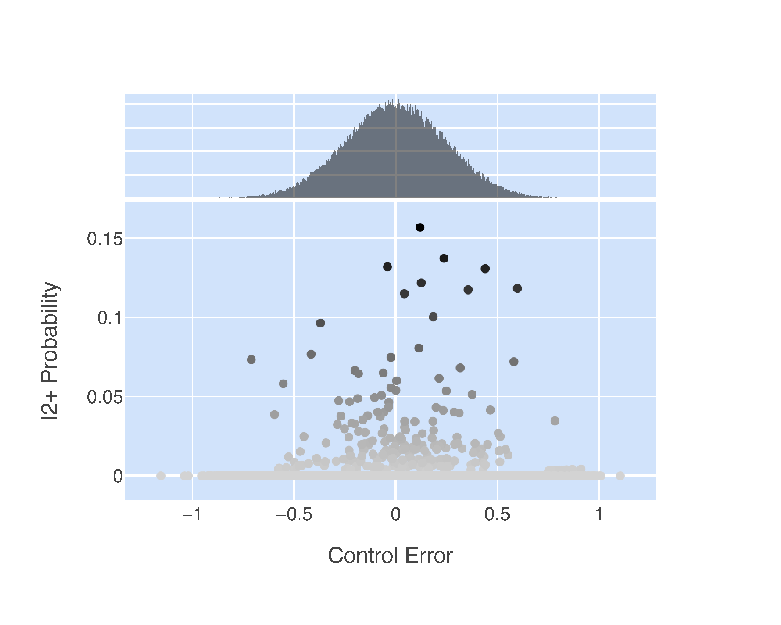}}
\caption{Scatter plot for visual \acrshort{sa}}\label{fig:sa_scatter}
\end{figure}

\begin{exmp}[Iterative system modification for \acrshort{hs}\,1]\label{ex:iterative_system_modification}
A visual \acrshort{sa} for \acrshort{hs}\,1 suggests that the overall risk is dominated by the initial \textit{Host Velocity}, which can be seen from  Fig.~\ref{fig:sa_scatter} (left). Fig.~\ref{fig:parallel_coordinates} shows that higher host velocity is associated with shorter detection distances for all three sensor channels, thereby triggering an emergency braking late. Additionally, at higher velocities, the braking distance also increases, resulting in more frequent and severe collisions. In contrast, Fig.~\ref{fig:sa_scatter} (right) suggests that improving the control accuracy is not expected to lead to significant safety improvements. \\
The \acrshort{rac} is exceeded by the considered scenario alone, cf.\ Ex.~\ref{ex:mcs}, so concrete measures must be initiated.
Two seemingly obvious measures for risk reduction need to be discarded: Debouncing in the sensor channels, which is primarily responsible for the delay, is indispensable to keep the risk of false braking low. A general reduction of the speed limit below which the \acrshorts{ads} is functional would have too drastic an impact on user-friendliness because of a disproportionately decreased availability of the function. \\
However, the analysis of the fleet data used in Example \ref{ex:joint_ifvs} reveals that the reason for the intruder is the congested traffic in the adjacent lane. For this reason, the Safety Concept is extended to include a safety mechanism that limits the differential speed to the neighboring lanes to an upper limit. Based on a local \acrshorts{sa}, the threshold for the differential speed limit is chosen so that the permissible overall budget is maintained without compromising the availability of the system.
\end{exmp}

For a more detailed understanding of local and global SA, we recommend referring to \cite{saltelli2008global}. This resource covers advanced topics, including challenges when input variables are correlated. \\


\remark{As the example above demonstrates, a Stochastic Simulation is not a black box whose results must be blindly trusted. By combining it with \acrshort{sa}, the main failure modes can be identified and validated.
}

\remark{
The V-model suggests a linear development process. Indeed, the framework is structured in such a way that the individual steps build upon each other. The final release documentation also follows this linear structure providing a traceable safety argumentation (safety case), cf.\ \cite{kelly2004goal, bishop2000methodology}. However, the path to this point is anything but linear, as explained in this chapter and visualized in Fig.~\ref{fig:iteration}. We advise initiating the validation loop early in the development process. Instead of relying on evidence, assumptions must be made at this stage based on previous projects. It is then often sufficient to combine an \acrshort{fta} with an \acrfull{eta} \cite{rausand2021system} that quantify the impact of  failures in a binary manner. Based on this, estimates for the required data volume can be made. The detailed error modeling follows as soon as the data becomes available, possibly leading to additional failure modes revealed by the in-depth Stochastic Simulation.
}

\remark{
While conservative assumptions may not always lead to conservative outcomes, they often simplify the modeling process significantly. By adopting a cautious approach in risk modeling, we can identify within the \acrshort{sa} which conservative assumptions contribute the most to the risk estimate. These assumptions can then be systematically replaced with more nuanced and less conservative alternatives, while retaining those conservative assumptions that do not significantly affect the overall risk.
}

\remark{
After the \acrshort{hira}, the high-risk scenarios are already known. These require in general the largest data amount and often the most complex Safety Concepts. It is therefore advisable to run the development loops first with these scenarios and then gradually address the remaining scenarios.
}

\remark{All sensitivities have to be weighted by the risk of the respective \acrshort{hs}. In other words, a 10\%  risk reduction in the most hazardous scenario can potentially be more beneficial than a 90\% risk reduction in a  harmless scenario. Moreover, there are interdependencies among the risks of the \acrshorts{hs}. A common occurrence is that risk reduction in a critical scenario, which necessitates a response from the host vehicle, might inadvertently increase the risk of false reaction during the entire drive. This trade-off requires an optimization over multiple scenarios.
}

\remark{
While it is certainly feasible to continuously integrate more safety features into the system, it is crucial to evaluate how sensitive the user experience is to these additions.
Although it is often the case that significant increases in safety may only slightly diminish customer satisfaction, this is not a universal rule. In instances where the system's availability and comfort may be significantly affected, alternative safety measures need to be considered. Ultimately, \acrshorts{ads} can only serve to enhance road safety if they are embraced and utilized by the users.
}

From experience, the iterative process of risk assessment and counteracting risk sources is marked by a steady decline in risk levels, although this progression is occasionally interrupted by the incorporation of new data. This reverse trend occurs as preliminary assumptions are substituted with more accurate data, leading to a reassessment of the risk. The process is deemed complete when all relevant evidence has been integrated into the models and the simulations confirm that the risk over all \acrshorts{hs} has been reduced to the \acrshort{rac}.

\section{Field Monitoring and Road Clearance}\label{sec:field_monitoring_and_road_clearance}
\begin{obj} The main objectives of this section are
\begin{itemize}
    \item to design and implement a system that observes the ADS customer fleet for deviations from expected safety behavior and
    \item  to establish precautionary measures that will be executed in the event of deviations.
\end{itemize}
\end{obj}

Despite the rigorous and systematic approach employed in the development, verification and validation of automated driving technology, it is crucial to acknowledge that it represents a new frontier, potentially introducing additional risks that have not been considered in the risk estimates. For example, there remains a residual probability that \acrshort{hs} may be overlooked during development. Furthermore, deviations in real-world conditions from the assumptions made during the development process may occur. Lastly, distributions of \acrshorts{hs} may shift over time, possibly through the introduction of the \acrshorts{ads} itself. \\
Consequently, it is essential to promptly detect any discrepancies in the field in order to respond to them before they could result in harm. This proactive monitoring is required by the \acrshort{alks} standard \cite{UNECE2021Reg157} and described in detail in ISO\,21448.\footnote{\isosotif, Clause~13}

\begin{exmp}[Analysis of customer fleet \acrshorts{mrm}]
Based on the findings of the risk assessment that deems, for example, the frequent execution of \acrshorts{mrm} as safety-relevant, a dedicated team is appointed to analyze all \acrshorts{mrm} performed across the fleet around the clock. In the event of unexpected incident clusters within specific geographic regions or road sections, criteria are established that dictate the immediate shutdown of operations in those areas. 
\end{exmp}

As for the field monitoring of the BMW Personal Pilot L3, all parameters indicate that it is operating within the expected norms and performing optimally mirroring the reliability and safety we have strived to achieve by the application of the presented framework.

\section{Summary and Concluding Remarks}\label{sec:summary_and_outlook}
This publication introduces \acrfull{sifad}, a safety framework for \acrshort{l3} and above
that has been validated through practical application. The contributions of this framework are twofold:

\begin{enumerate}
    \item \acrshort{sifad} focuses on the quantitative safety assessment. For each \acrfull{hs}, uncertainties in the redundantly designed system and environment are systematically identified. For their quantification, the framework leverages public statistics, expert knowledge and customer fleet data reducing the required amount of field and proving ground testing to a manageable level. Through Stochastic Simulation, the risk in each \acrshort{hs} is quantified and aggregated into the overall system risk, expressed in terms of the predicted frequency of different levels of injuries. The presented combination of probabilistic methods in the context of \acrshort{l3} automated driving is novel.
    
    \item  \acrshort{sifad} embeds the quantitative safety assessment methods into established qualitative methods and processes. The experience has shown that, right from the start of development, the \acrshorts{hs} have to be identified and the system must be specially designed and implemented to reliably handle them. However, the systematic identification of \acrshorts{hs} and the derivation, integration, and testing of safety requirements are deeply rooted in the ISO\,26262 approach. For this reason, efforts have been made to unify the novel quantitative methods with the established qualitative patterns and to present them in a neutral language. We are fully convinced that this unified approach leads to a safer ADS. 
\end{enumerate}

There remain two limitations regarding this unified approach, the first one of which is associated with the \acrshort{rac} of ISO\,26262: Even though its order of magnitude can be reconstructed\footnote{E.g., $10^{-8} / h$ for \acrshort{asil}~D hardware required by Part 5, Table 6 in combination with Part~3, Table 4 and Tables~B.1, B.2, B.3} and lies in the range of the Minimum Endogenous Mortality (MEM)\footnote{
    $1/20 \, \text{MEM} = 1/20 \cdot 2 \cdot 10^{-4} 1/a = 10^{-5} 1/a = 2.5 \cdot 10^{-8} / h$ for a yearly vehicle usage of $a=400h$} 
\cite{junietz2019macroscopic},  the \acrshort{rac} is not explicitly defined in ISO\,26262. Instead, the \acrshort{rac} is implicitly taken into account by the \acrshort{asil}-determining table.\footnote{Part~3, Clause 6, Table 4} Thus, unlike IEC\,61508, it is not possible to prescribe a different \acrshort{rac} for the derivation of integrity levels. If the \acrshort{prb} is used as the \acrshort{rac}, as is common for \acrshort{l3} and above, then the following inconsistencies may arise for new products:
\begin{enumerate}
    \item The (implicit) \acrshort{rac} of ISO\,26262 is a constant. However, variability in human performance may stem from different \acrshorts{odd}, which in turn can lead to a range of \acrshort{rac} outcomes for different products. Therefore, from the perspective of a \acrshort{prb}, the derived \acrshorts{asil} can be conservative (in \acrshorts{odd} with low human safety performance) or optimistic (in \acrshorts{odd} with high human safety performance). 
    
    \item The \acrshort{asil} increases in general with the annual usage time of the \acrshort{ads}. In contrast, the \acrshort{prb} is independent of usage time, as the system is measured against human driving performance during any activation. Hence, from the perspective of a \acrshort{prb}, the longer the usage, the more conservative are the derived \acrshorts{asil}.
    
    \item The (implicit) \acrshort{rac} of ISO\,26262 is the same for all hazards, regardless of the number of hazards. For the \acrshort{prb}, however, the sum of the risks over all hazards ultimately counts. Therefore, it may be sensible to assign individual risk budgets for each scenario, just like the human performance might vary over different scenarios. 
\end{enumerate}

A second limitation was already mentioned in Remark~\ref{rem:simplified_model} and \ref{rem:compute_asil}: For risk assessment of certain hazards, it is only possible to make a rough estimate of the risk with the E, C, and S factors. Ultimately, it is not the individual factors that matter for the required integrity level but their combined effect. The effect, however, can be determined more accurately with advanced statistical methods, such as \acrshort{mcs}, without the detour via the independent factors. \\

Both shortcomings could be resolved, e.g., as part of a revised or a new standard, with a mapping between \acrshorts{asil} and (average) failure probabilities ($p_s$) and rates ($\lambda_s$), cf.\ Table~\ref{table:sil_mapping_iec}, in combination with an independently prescribed \acrshort{rac}, both analogous to IEC\,61508.\footnote{cf.\ IEC\,61508-5, Annex B and D} This would ensure that qualitative and quantitative processes and methods can be consistently aligned throughout the development of future \acrshorts{ads}. \\

This brings us to the final point, which is particularly challenging for new \acrshort{ads} that require high levels of integrity: The implementation of a safety requirement through multiple architectural elements can -- given certain prerequisites -- lead to a reduction of the required integrity of the individual elements compared to the required integrity of the overarching safety requirement.\footnote{ISO\,26262-9:2018 Clause 5 (Requirements decomposition with respect to \acrshort{asil} tailoring)} For example, an \acrshort{asil} D requirement can be realized by two \acrshort{asil} B elements. Two important prerequisites for a so-called \quotes{requirement decomposition} are, that the elements are \quotes{sufficiently independent} and \quotes{shall comply with the initial safety requirement}. This way errors in the individual channels do not propagate due to redundancy. \\
Checking these prerequisites in the presence of uncertainties in the involved elements (\acrshorts{spv}) is non-trivial but could be done by a numerical analysis quantifying the interaction between software faults and performance limitations. This analysis, however, requires certain assumptions that would need to be agreed upon, e.g., in a revised version of the standard. \\

In conclusion, the paper aims to provide an overview and foster discussion on the application of the \acrshort{sifad} to new \acrshorts{ads}. Ultimately, the introduction of \acrshorts{ads} resulting in a positive risk balance is instrumental in contributing to an increase in road safety.


\clearpage
\printglossary[title=Acronyms, toctitle=List of Acronyms]
\printbibliography
\end{document}